\pdfoutput=1
\documentclass[sigconf]{acmart}
\usepackage{comment}
\usepackage{amsmath}

\usepackage{amssymb}
\usepackage{graphicx}
\usepackage{subfigure}
\usepackage{multirow}
\usepackage[misc]{ifsym}
\usepackage{tabu}
\usepackage{algorithm} 
\usepackage{algorithmic} 
\usepackage{bbding}

\newcommand{\sign}[1]{\mathrm{sgn}(}

\copyrightyear{2021} 
\acmYear{2021} 
\setcopyright{acmcopyright}\acmConference[CIKM '21]{Proceedings of the 30th ACM International Conference on Information and Knowledge Management}{November 1--5, 2021}{Virtual Event, QLD, Australia}
\acmBooktitle{Proceedings of the 30th ACM International Conference on Information and Knowledge Management (CIKM '21), November 1--5, 2021, Virtual Event, QLD, Australia}
\acmPrice{15.00}
\acmDOI{10.1145/3459637.3482247}
\acmISBN{978-1-4503-8446-9/21/11}

\begin{document}
	\fancyhead{}
	
	\title{Deep Self-Adaptive Hashing for Image Retrieval}
	
	\author{Qinghong Lin$^1$, Xiaojun Chen$^1${\Envelope}, Qin Zhang$^1$, Shangxuan Tian$^2$, Yudong Chen$^3$}
	\affiliation{%
		\institution{$^1$Shenzhen University, $^2$Tencent, $^3$The University of Queensland}
		\country{}
	}
	\email{linqinghong@email.szu.edu.cn, xjchen@szu.edu.cn, qinzhang2014@gmail.com}
	\email{tianshangxuan@u.nus.edu, ydchen7@foxmail.com}
	
	\begin{abstract}
		Hashing technology has been widely used in image retrieval due to its computational and storage efficiency. Recently, deep unsupervised hashing methods have attracted increasing attention due to the high cost of human annotations in the real world and the superiority of deep learning technology. However, most deep unsupervised hashing methods usually pre-compute a similarity matrix to model the pairwise relationship in the pre-trained feature space. 
		Then this similarity matrix would be used to guide hash learning, in which most of the data pairs are treated equivalently. The above process is confronted with the following defects:
		1) The pre-computed similarity matrix is inalterable and disconnected from the hash learning process, which cannot explore the underlying semantic information.
		2) The informative data pairs may be buried by the large number of less-informative data pairs. 
		To solve the aforementioned problems, we propose a \textbf{Deep Self-Adaptive Hashing~(DSAH)} model to adaptively capture the semantic information with two special designs: \textbf{Adaptive Neighbor Discovery~(AND)} and \textbf{Pairwise Information Content~(PIC)}. 
		Firstly, we adopt the AND to initially construct a neighborhood-based similarity matrix, and then refine this initial similarity matrix with a novel update strategy to further investigate the semantic structure behind the learned representation.
		Secondly, we measure the priorities of data pairs with PIC and assign adaptive weights to them, which is relies on the assumption that more dissimilar data pairs contain more discriminative information for hash learning. 
		Extensive experiments on several datasets demonstrate that the above two technologies facilitate the deep hashing model to achieve superior performance.
	\end{abstract}

	
	\maketitle
	
	\section{Introduction}
	
	With the rapid development of the Internet, the explosive growth of multimedia data poses huge challenges to accurate nearest-neighbor based searching methods. Instead, due to its high efficiency, the approximate nearest neighbor~(ANN)~\cite{indyk1998approximate} based search methods have attracted increasing attention. 
	Among them, hashing technologies contribute remarkably to their fast query speed and low storage overhead.
	
	To be specific, hashing techniques can be divided into supervised and unsupervised categories. Supervised hashing~\cite{ shen2015supervised, shen2015supervised, kang2016column, gui2017fast, jiang2017asymmetric, luo2018fast, chen2019deep, tu2021partial, tu2020deep} methods use the label information to train hashing models, which achieve fine performance. However, the annotations are highly time-consuming and also expensive to collect, which limits these methods in many practical applications. 
	Due to this scenario, unsupervised hashing methods~\cite{salakhutdinov2009semantic, kong2012isotropic, gong2012iterative, he2013k, wang2018deep, he2019k} have drawn a large amount of attention. Many unsupervised hashing methods have been proposed in the past decade, including Spectral Hashing (SH)~\cite{weiss2009spectral}, Hashing with Graphs (AGH) \cite{liu2011hashing}, Iterative Quantization (ITQ)~\cite{gong2012iterative}, Stochastic Generative Hashing (SGH)~\cite{dai2017stochastic} , etc.
	The previous unsupervised hashing methods have made progress in this area, however, they normally adopt shallow architectures and severely depend on hand-crafted features~(such as SIFT features~\cite{lowe1999object}), which degrade the learning performance.
	
	In recent years, numerous deep learning techniques have been introduced into unsupervised hashing methods~\cite{lin2016learning,li2017deeps, tu2018object} due to their powerful feature representation capability.
	With the lack of labels, most of the unsupervised deep hashing methods construct a pairwise similarity matrix with pre-trained deep features.
	For example, 
	~\citet{yang2018semantic} compute the similarity matrix based on the observation that the distribution of the cosine distance for point pairs can be estimated by two half Gaussian distributions. 
	Once the similarity matrix is built, in most of the existing methods, it will be fixed to guide the hash code learning process. However, such a similarity matrix may be unreliable because it is computed with the pre-trained deep features and ignores the semantic information in the downstream retrieval task. To tackle this issue, ~\citet{shen2018unsupervised} employ a straightway by reconstructing a similarity graph with the fine-tuning features through the Gaussian kernel.
	However, such specific construction schemes mainly focus on the local structure, which limits the model performance, and also, the reconstruction strategy suffers from high computational costs.
	
	In addition, most of the existing methods~\cite{xia2014supervised, cao2016deep, deng2019unsupervised} ignore the semantic importance of data pairs during their design of the similarity-preserving loss function. As such data pairs are treated equivalently in the learning process. 
	A few of works~\cite{yang2018semantic, zhang2020deep, zhangz2020deep, qin2020unsupervised} primitively divide the data pairs into different types like confident or unconfident, they still fail to provide a fine-grained measurement of the importance of different data pairs. 
	The pairs in the same class are still treated equivalently.
	However, the priority of different data pairs should be different. Intuitively, the dissimilar data pairs might contain more discriminative information, and be more informative for model training.
	Otherwise, the informative data pairs will be buried by large number of less-informative training pairs and decline hash learning.
	
\begin{figure*}[!t]
		\centering
		\includegraphics[width=0.9\linewidth]{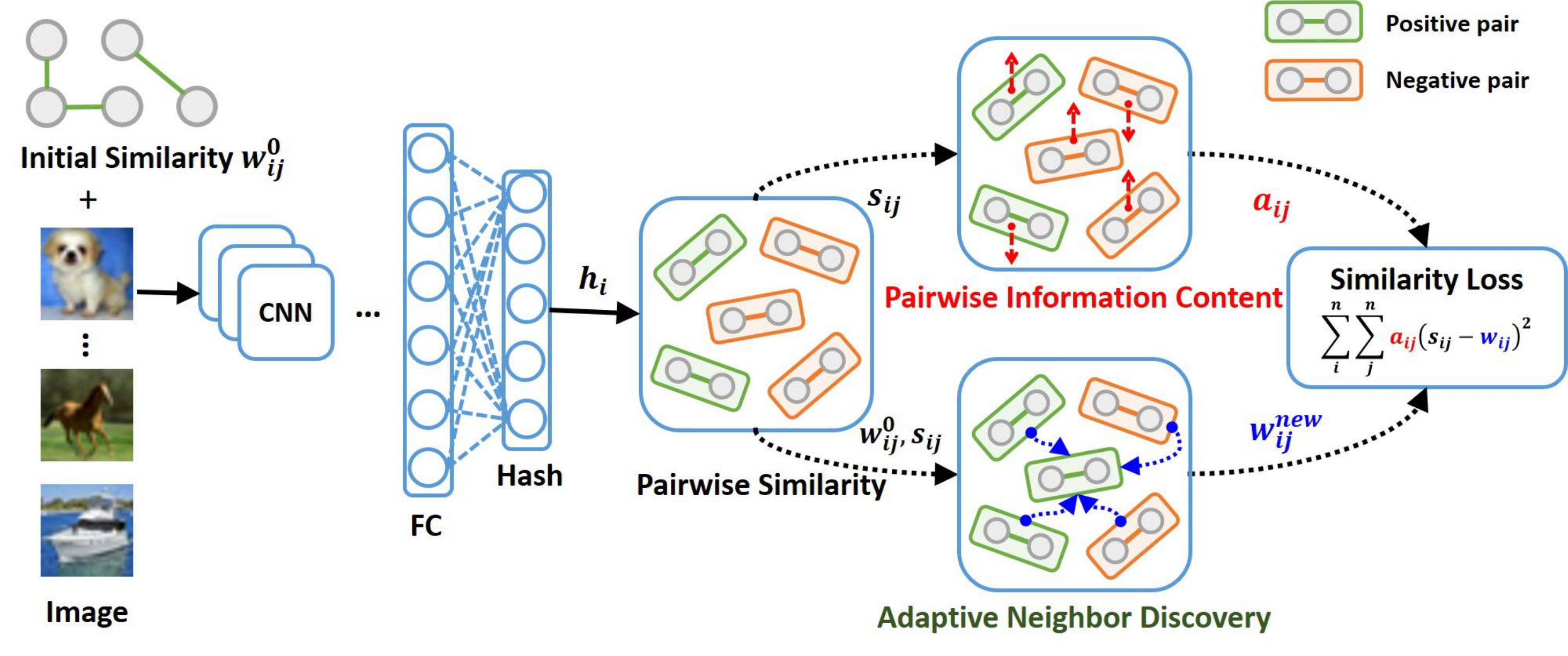}
		\caption{The pipeline of the proposed {Deep Self-Adaptive Hashing~(DSAH)}. 
			First, we initialize a similarity matrix based on the original features and extract the image features via VGG-19 to compute the pairwise similarity. Next, we provide two novel technologies: (i) We introduce {PIC} (Upper branch) to augment the priority of informative data pairs. (ii) We propose AND (Down branch) to refine the initialized similarity matrix and mine the implicit neighbors' relationships with learned representation. These two components assist DSAH to learn hash codes in a self-adaptive manner.}
		\label{pipeline}
	\end{figure*}

	To deal with the aforementioned issues, we propose a novel \textbf{Deep Self-Adaptive Hashing~(DSAH)} method to adaptively explore the semantic information under training, and provide two innovative components.
	The pipeline is shown in Figure~\ref{pipeline}. 
	In particular, to overcome the disadvantage of fixed pre-computed similarity, we adopt an \textbf{Adaptive Neighbor Discovery~(AND)} technique, which mines underlying neighbor relationships behind the fine-tuning features gradually during training, and update the initial similarity graph continuously.
	Further, we introduce an adjustable pairwise weight term \textbf{Pairwise Information Content~(PIC)} to distinguish the different importance of data pairs. In this way, the data pairs with more information would gain greater weights and contribute more to model learning.
	These two techniques enable DSAH to fully explore the semantic information behind the data pairs and learn better hash codes in a self-adaptive manner. 
	The main contributions of this paper can be summarized as follows:
	
	\begin{itemize}
		\item We propose a novel \textbf{DSAH} method to yield better hash codes in a self-adaptive manner with the full  exploration of semantic information behind the data pairs.
		\item We propose the \textbf{AND} technique to refine the pre-computed similarity matrix with the fine-tuned representations during the learning process, which can adaptively capture the implicit neighbor relationships and improve the hash learning performance.
		\item We propose the \textbf{PIC} to measure the importance of different data pairs and use it to weigh them during training, where more dissimilar data pairs will be assigned larger weights to augment their discriminative power. These weights are further adaptively updated with the pairwise similarity distribution when the training iterates.
		\item The extensive experiments on several benchmark datasets show that our DSAH is indeed effective and achieves superior performance, which achieves 6.2\%, 8.7\%, 2.7\% improvement over the best baseline on CIFRA-10, FLICKR25K, and NUS-WIDE datasets, respectively.
	\end{itemize}
	
	\section{Notation and Problem Definition}
	Let us introduce some notations for this paper. 
	We use boldface uppercase letters $\mathbf{A}$ to represent a matrix, in which $\mathbf{a}_i$ represents the $i$-th row of $\mathbf{A}$, $a_{ij}$ represents the element of $\mathbf{A}$ which is in the $i$-th row and the $j$-th column. 
	$\|\cdot\|$ denotes the L2 norm and $\|\cdot\|_F$ represents the Frobenius norm. 
	$\mathbf{1}(\cdot)$ is the indicator function.
	$\tanh(\cdot)$ is the hyperbolic tangent function.
	$\sign(\cdot)$ is the sign function, which outputs $+1$ for positive numbers, or $-1$ otherwise.
	\begin{equation}
		\sign(x)=\left\{
		\begin{aligned}
			1,\quad x \geq 0 \\
			-1,\quad x< 0 \\
		\end{aligned}
		\right.
		\label{sign}
	\end{equation}
	
	Given a dataset contain $n$ samples $\mathbf{X}=\{\mathbf{x}_1,\mathbf{x}_2, \cdots, \mathbf{x}_n\}\in\mathbb{R}^{d\times n}$ without human annotations, where $d$ is the dimension of samples. The goal of hashing is to learn a function ${\mathcal{H}}: \mathbf{x}_i\rightarrow \mathbf{b}_i$ that maps the dataset $\mathbf{X}$ into a set of compact binary hash codes $\mathbf{B}=\{\mathbf{b}_1, \mathbf{b}_2, \cdots, \mathbf{b}_n\}\in\{+1,-1\}^{l\times n}$ where $l$ is the length of codes.
	
	\section{Related Work}
	In this section, we briefly review the traditional shallow hashing methods and recent deep hashing methods. These methods can be divided into supervised and unsupervised categories, and we mainly focus on unsupervised hashing methods.
	
	\subsection{Shallow Hashing}	
	Generally, traditional shallow hashing methods can be classified into two categories: data-independent hashing methods and data-dependent methods. Data-independent hashing methods \cite{gionis1999similarity, andoni2006near, kulis2009kernelized} typically use random projection to generate binary code.
	Locally Sensitive Hashing~(LSH)~\cite{andoni2006near} is a representative data-independent hashing method, which enables similar data to share similar hash codes in Hamming space. However, data-independent hashing methods require longer binary codes to achieve higher accuracy, which also results in higher storage costs. The data-dependent hashing methods \cite{weiss2009spectral, liu2011hashing, gong2012iterative, dai2017stochastic} have also received increasing attention in recent years, which aim to learn compact and high-quality binary codes from specific data. 
	One of the most classical data-dependent hashing methods is Spectral Hashing (SH) \cite{weiss2009spectral}, which was developed to learn binary codes by preserving local manifold structures. The objective function of SH can be written as:
	\begin{equation}
		\begin{gathered}
			\min_\mathbf{B }\sum_{ij}w_{ij}\|\mathbf{b}_i - \mathbf{b}_j\|^2 \\
			s.t. \mathbf{b}_i\in\{-1,+1\}^l,\quad \sum_i \mathbf{b}_i=0,\quad \frac{1}{n}\sum_i \mathbf{b}_i\mathbf{b}_i^T=I
		\end{gathered}
		\label{SH}
	\end{equation}
	where $w_{ij}=\exp(-\|\mathbf{x}_i-\mathbf{x}_j\|^2/\varepsilon)$ is the similarity weight between $x_i$ and $x_j$.
	$\varepsilon$ is the balance parameter. The optimization of Eq. \ref{SH} is equivalent to balanced graph partitioning and is an NP-hard problem. SH solves this problem by introducing spectral relaxation, which removes the discrete constraint. The objective function of SH is to binarize the input so that the Hamming distance approximates the Euclidean distance. Generally, the shallow hashing methods usually encode the input images with hand-crafted features~(e.g, $\|\mathbf{x}_i-\mathbf{x}_j\|^2$  of $w_{ij}$), which is insufficient to characterize the semantic information of data and thus degrade the performance.
	
	\subsection{Deep Hashing}	
	Recently, deep hashing methods~\cite{li2015feature, erin2015deep,li2017deep, qiu2017deep1, cao2017hashnet, liu2019cross, yang2019distillhash} show promising performance compared with the shallow hashing methods. ~\citet{xia2014supervised} is one of the earliest works that introduces the deep neural networks into hashing function learning, where the hash codes are learned by minimizing the following loss function
	\begin{equation}
		\mathcal{L}_0 = \sum_{i=1}^n \sum_{j=1}^n\left( s_{ij}-w_{ij}\right) ^2\\
		\label{CNNH}
	\end{equation}
	where $s_{ij}=\frac{1}{l}\mathbf{z}_i^T\mathbf{z}_j$ is the inner product between deep features. $\mathbf{z}_i\in [-1,+1]^l$ is the relaxation of hash codes $\mathbf{b}_i$ since it is difficult to directly optimize the discrete variables. 
	Suggested by ~\citet{cao2016deep}, cosine similarity might be a better choices to mitigate the diversity of code lengths and improve the performance, which defined as:
	\begin{equation}
		s_{ij} = cos(\mathbf{z}_i, \mathbf{z}_j) = \frac{\mathbf{z}_i^T\mathbf{z}_j}{\| \mathbf{z}_i \| \| \mathbf{z}_j \|}\\
		\label{sij}
	\end{equation}
	where $w_{ij}$ of Eq.\ref{CNNH} is the binary pairwise similarity, which is equal to $+1$ if $\mathbf{x}_i$ and $\mathbf{x}_j$ are semantically similar, or $-1$ otherwise.
	
	In supervised cases, $w_{ij}$ is constructed according to labels. But in the unsupervised settings, the labels are unavailable. To tackle this problem, 
	~\citet{yang2018semantic} empirically study the deep feature statistics and estimate two half Gaussian distributions to calculate the similarity. ~\citet{song2018binary} employs the $k$ Nearest Neighborhoods~($k$-NN) to compute a binary similarity with pre-trained features.
	While ~\citet{zhang2020deep} construct a hybrid-similarity matrix in advance.
	~\citet{tu2020mls3rduh} utilizes the local manifold structure to construct the similarity graph. 
	However, in these methods, the semantic similarity matrix is pre-computed on the original feature space and is separated from the hash learning process.
	Intuitively, a good similarity representation is beneficial to hashing function learning, and also, hashing function might provide meaningful feedback to similarity representation. Therefore we hope that the similarity matrix and hashing function can be learned simultaneously, which inspires us to design an adaptive similarity updating method rather than reconstructing a similarity structure based on the updated features~\cite{shen2018unsupervised}. 
	Moreover, the existing similarity-preservation loss function (e.g, Eq.\ref{CNNH}) usually neglects the priorities between different data pairs in the learning process. 
	Though several works~\cite{cao2017hashnet, yang2018semantic, zhang2020deep, qin2020unsupervised} divide the data pairs into different groups based on their pairwise similarity, these partitions are also pre-defined and fail to capture the fine-grained priority between data pairs since the pairs within the same group are still treated fairly. Thus the informative data pairs may be buried in a large number of uninformative training pairs.
	
	\section{Methodology}
	In this section, we develop our Deep Self-Adaptive Hashing model. The overall architecture of DSAH is illustrated in Figure~\ref{pipeline}, which contains three modules: Feature Extraction, Adaptive Neighbor Discovery and Pairwise Information Content. We will demonstrate each module in the following section in detail.
	
	\subsection{Feature Extraction}
	We apply the VGG-19~\cite{simonyan2014very} for hash function learning and denote it as $\mathcal{F}(\cdot, \Theta)$ with network parameters $\Theta$.  VGG-19 contains five convolutional layers and three fully-connected layers, to make the network suitable for hash learning, we replace the last layers with a fully-connected layer with 1000 hidden units and followed by a hash layer, in which the number of neurons is equal to hash codes length $l$. To resolve the ill-posed gradient of $\sign(\cdot)$, we adopt the $\tanh(\cdot)$ as the activation function of the hash layer and then we get the approximation of hash code $\mathbf{b}_i$ as follows:
	\begin{equation}
		\mathbf{z}_i=\tanh(\mathcal{F}(\mathbf{x}_i;\Theta))\in [-1,+1]^l
		\label{zij}
	\end{equation}
	Once we finish the training process, we can obtain the discrete hash code $\mathbf{b}_i$ as follows:
	\begin{equation}
		\mathbf{b}_i=\sign(\mathcal{F}(\mathbf{x}_i;\Theta))\in \{-1,+1\}^l
		\label{bij}
	\end{equation}
	
	\subsection{Adaptive Neighbor Discovery}
	In this subsection, we propose the AND, which consists of two steps: similarity matrix initialization and updating.
	\subsubsection{Initialization}
	\label{Initialization}
	Recent works~\cite{song2018binary, yang2018semantic,zhang2020deep, tu2020mls3rduh} have shown that the rich semantic structure can be captured via an elaborately designed similarity matrix. 
	In our AND, we construct an initial similarity matrix at the beginning based on $k$-NN.
	
	We first extract deep features from the relu-7 layer of a pre-trained VGG-19 model and then select $k_1$ images with the highest cosine similarity as the neighbors of each image. Then we construct a \textit{Low-order} similarity matrix $\mathbf{W}_L$ as follows:
	\begin{equation}
		(\mathbf{W}_L)_{ij}=
		\left\{
		\begin{aligned}
			+1,&\quad \text{if $\mathbf{x}_j$ is $k_1$-NN of $\mathbf{x}_i$}
			,\\
			-1,&\quad \text{otherwise}.
		\end{aligned}
		\right.
		\label{wL}
	\end{equation}
	
	Notably, we consider that if the neighbors of two images are highly similar, then these two images should also be very similar. Thus we calculate the similarity of two images neighbors using the expression  $\frac{1}{1 + \| (\mathbf{W}_L)_i - (\mathbf{W}_L)_j \|}$. Then we select the top $k_2$ samples to construct a \textit{High-order} similarity matrix $\mathbf{W}_H$ as:
	\begin{equation}
		(\mathbf{W}_H)_{ij}=
		\left\{
		\begin{aligned}
			+1,&\quad \text{if $\mathbf{x}_j$ is $k_2$-NN of $\mathbf{x}_i$}
			,\\
			-1,&\quad \text{otherwise}.
		\end{aligned}
		\right.
		\label{wH}
	\end{equation}
	
	In order to take full advantage of these two similarity matrices, we define our initial similarity matrix $\mathbf{W}^0$ by combining $\mathbf{W}_L$ and $\mathbf{W}_H$ together, which is based on the assumption that two similar images should not only be similar in feature space but also share similar neighbors.
	\begin{equation}
		(\mathbf{W}^0)_{ij}=
		\left\{
		\begin{aligned}
			+1,&\quad \text{if $(\mathbf{W}_L)_{ij}$ = +1 and $(\mathbf{W}_H)_{ij}$ = +1}\\
			-1,&\quad \text{otherwise}.
		\end{aligned}
		\right.
		\label{w0}
	\end{equation}
	
	However, a drawback of $\mathbf{W}^0$ is that
	it is built with the pre-trained features, which might contain noisy data. Ideally, we hope that the similarity maintains consistency with the fine-tuning features, while the  reconstruction is time-consuming. Hence, we next propose an efficient updating strategy to refine the $\mathbf{W}^0$.
	
	\subsubsection{Updating}
	AND provides a progressive mechanism for similarity updating. 
	Given a similarity matrix $\mathbf{W}^r$ in the $r$-th round~(or $\mathbf{W}^0$ in beginning), we first use it to update the model parameters $\Theta$. (e.g, Optimizing the $\mathcal{L}_0$ in Eq.~\ref{CNNH}).
	Next, we measure the pairwise cosine similarity $\mathbf{S}^r=\{s^r_{ij}\}\in \mathbb{R}^{n\times n}$ over the training data pairs through Eq.\ref{sij} and estimate a threshold $m^r$ by
	\begin{equation}
		m^r =  \mu^r + \gamma \cdot \sigma^r
		\label{mr}
	\end{equation}
	where the $\mu^r$ and $\sigma^r$ denote the mean and standard derivation of cosine similarity $s^r_{ij}$ for those data pairs with $w^r_{ij}=+1$, which is defined as:
	\begin{equation}
		\left\{
		\begin{aligned}
			\mu^r&=\frac{\sum_{i=1}^n \sum_{j=1}^n s^r_{ij}\cdot \mathbf{1}(w_{ij}^{r}=+1) }{n_+ }\\
			\sigma^r&=\left( \frac{\sum_{i=1}^n \sum_{j=1}^n \left( s^r_{ij} -\mu^r \right)^2\cdot \mathbf{1}(w_{ij}^{r}=+1) }{n_+ } \right) ^{\frac{1}{2}}\\
		\end{aligned}
		\right.
	\end{equation}
	where $n_+$ is equal to $\sum_{i=1}^n\sum_{j=1}^n \mathbf{1}(w_{ij}^{r}=1)$, which counts the number of neighbors in $\mathbf{W}^r$.
	$\gamma$ is a hyper-parameter that control the threshold.
	Finally, we renew the $\mathbf{W}^{r}$ as follows:
	\begin{equation}
		w^{r+1}_{ij}=
		\left\{
		\begin{aligned}
			+1,&\quad \text{if } w^r_{ij}=-1 \text{ and } s^r_{ij}\geq m^r,  \\
			w^r_{ij},&\quad \text{otherwise}.
		\end{aligned}
		\right.
		\label{update}
	\end{equation}

	\textbf{Analysis}. 
	The AND is motivated by the prediction interval, aiming to design a dynamic criterion~Eq.\ref{mr} based on the global distribution of learned features. The tuition behind Eq.\ref{update} is that these dissimilar data pairs with similarity $s_{ij}$ higher than average similarity $\mu$ of similar pair sets could probably be treated as candidate similar data pairs in next epoch, where $\gamma$ controls the tolerance, a larger $\gamma$ is more serious while a smaller $\gamma$ is looser.
	The reason why we focus on similar pairs instead of dissimilar pairs is that the neighborhoods-based $\mathbf{W}^0$ contains only a few similar pairs~($n_+\ll n^2$). There is still a lot of information (e.g, neighbors) worth mining in a large number of dissimilar pairs, as the model capacity increases, this information would be better distinguished.
	Therefore, AND is able to adaptively adjust the similarity $\mathbf{W}^0$ according to the learned representation, in a progressive mechanism.
	
	\subsection{Pairwise Information Content}
	In PIC, we discuss the priority of different data pairs for model training.
	Though Eq.~\ref{CNNH} provides a scheme to learn hash codes from similarity matrix $\mathbf{W}$, it neglects the importance of different data pairs. All the data pairs with their similarities are treated fairly when calculating loss, so that some informative data pairs may be buried in a large number of samples. To tackle this issue, we propose the PIC, which assigns an adaptive weight for each data pair. If a data pair is more important, it will receive a larger weight and contribute more to hashing function learning. So, we define the following loss function:
	\begin{equation}
		\mathcal{L}_1=\sum_{i=1}^n\sum_{j=1}^n a_{ij}\left( s_{ij}-w_{ij}\right) ^2
		\label{l1}
	\end{equation}
	where $a_{ij}$ is the weight for the data pair $(i,j)$ which represents the importance of this data pair. 
	
	According to the information theory~\cite{shannon1948mathematical}, given an event $\Phi$ with probability $p(\Phi)$, its \textit{information content} is defined as the negative log-likelihood:
	\begin{equation}
		I(\Phi) = -\log p(\Phi)
		\label{ic}
	\end{equation}
	If an event $\Phi$ has a probability 1 of occurring, then its information content is $-\log(1)=0$ and yields no information. While an event with probability 0, its information content is $+\infty$. Inspired by Eq.~\ref{ic}, we design the following definition:
	
	\begin{definition}
		Given a data pair $(i,j)$, we define an event $\Phi_{ij}$ with probability $p_{ij}$. The $\Phi_{ij}$ indicates that the image $j$ is the top-1 retrieval result of the given query image $i$, which is an optimum matching of image retrieval.
		Based on this, we define the \textbf{pairwise information content} $a_{ij}$ as:
		\begin{equation}
			\label{def_a}
			a_{ij} = I(\Phi_{ij}) = -\log (p_{ij})
		\end{equation}
	\end{definition}

	Typically, if an image $j$ is the top-1 retrieval result of the query image $i$, then $i$ and $j$ should be the most similar. Therefore probability $p_{ij}$ can be expressed as a measure of relative similarity between image $i$ and image $j$, and we define $p_{ij}$ as
	\begin{equation}
		p_{ij}=\frac{\exp (s_{ij} /\tau)}{\sum_{g=1}^{n}\sum_{k=1}^n \exp (s_{gk}/\tau)}
		\label{pij}
	\end{equation}
	where $\tau$ is a temperature parameter and $s_{ij}$ is the pairwise cosine similarity in Eq.~\ref{sij}.
	
\begin{figure}[!t]
		\centering
		\includegraphics[width=0.9\linewidth]{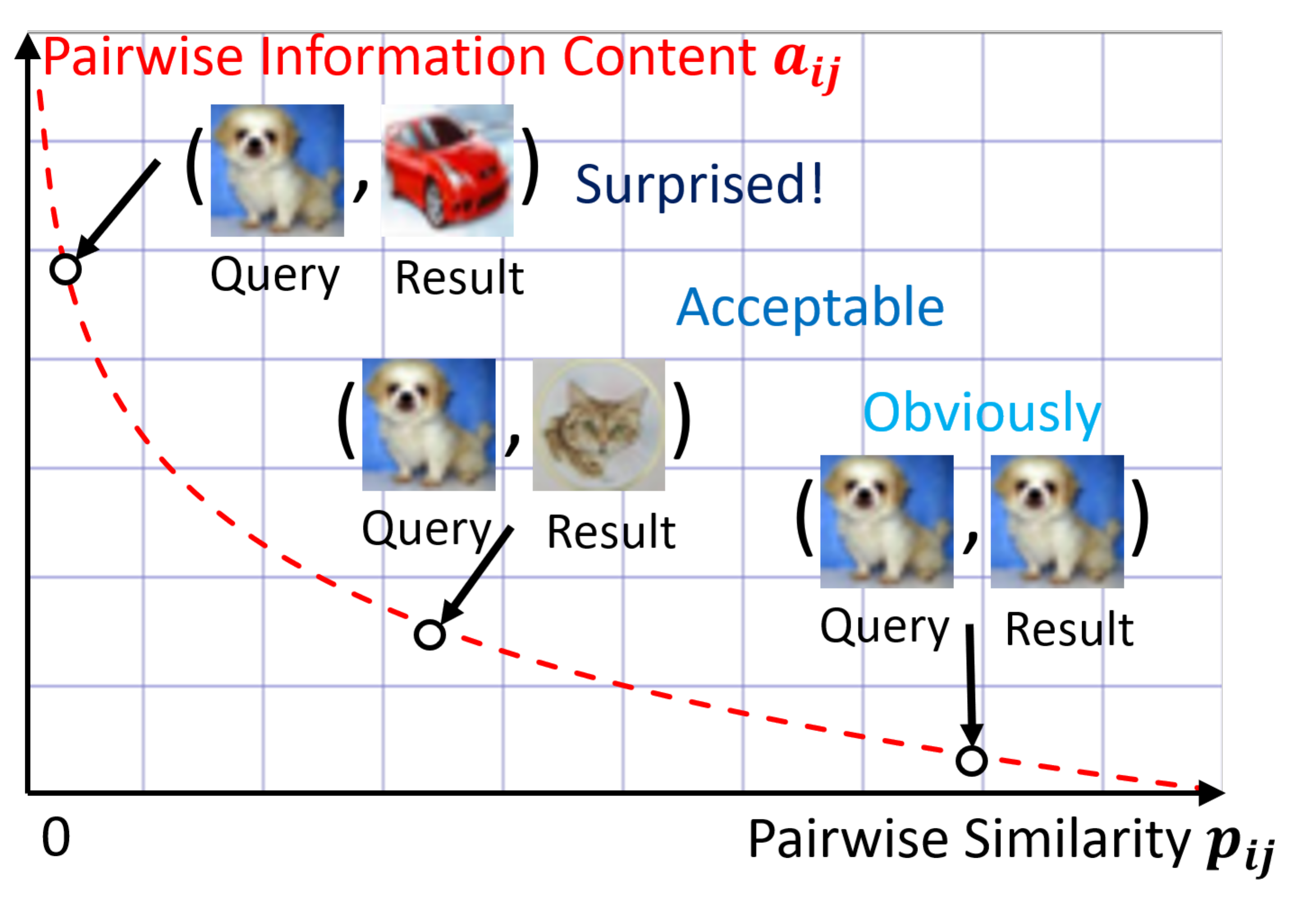}
		\caption{The relationship between PIC weight $a_{ij}$ and relative pairwise similarity $p_{ij}$. 
			If a data pair is highly dissimilar, their $a_{ij}$ will be higher, otherwise, it will be smaller.}
		\label{pic_fig}
		\vspace{-2.0em}
	\end{figure}

	\textbf{Analysis}. Given a data pair $(i,j)$, if $p_{ij}$ is higher, then the information content $I(\Phi_{ij})$ that the image $j$ can be retrieved through the image $i$ is lower. (A special case is to use an image to retrieve itself, then $I(\Phi_{ij})$ should be lowest or close to 0.) On the contrary, retrieving an image via a highly dissimilar query would bring a lot of information.
	Figure.\ref{pic_fig} shows an example, given a query image of \textit{Dog}, if we retrieve a \textit{Car}, we would be more surprised than retrieving a \textit{Cat}. So that we have $I(\Phi_{Dog, Car}) > I(\Phi_{Dog, Cat})$ while $sim({Dog}, {Car}) < sim({Dog}, {Cat})$, where $sim$ denotes the pairwise similarity.
	The tuition behind PIC can work is that a data pair contains highly dissimilar images would provide richer distinctive information, so it should be given a larger weight contributing to the hashing learning.
	The PIC is based on the pairwise similarity distribution $p_{ij}$, it is adjustable and adaptive.
	Moreover, our PIC could also be expressed as a kind of pairwise attention mechanism.
	Particularly, when $a_{ij}$ is equal to constant 1, the importance of data pairs will be the same and the Eq.~\ref{l1} will degenerate to Eq.~\ref{CNNH}.
	
	\subsection{Objective Function}

	In addition, to guarantee the quality of learned hash codes, we introduce a loss $\mathcal{L}_{2}=\|\mathbf{Z}-\mathbf{B}\|_F^2$ to minimize the quantization error between the variables $\mathbf{Z}$ and the binary codes $\mathbf{B}$. 
	Therefore, the total loss function can be formulated as
	\begin{equation}
		\begin{aligned}
			\mathcal{L} &= \mathcal{L}_1 + \lambda \mathcal{L}_2\\
			&=\sum_{i=1}^n\sum_{j=1}^n a_{ij}\left( s_{ij}-w_{ij}\right) ^2 + \lambda \|\mathbf{Z}-\mathbf{B}\|^2_F\\
			&=\sum_{i=1}^n\sum_{j=1}^n -\log(p_{ij})\left( \frac{\mathbf{z}_i^T\mathbf{z}_j}{\|\mathbf{z}_i \| \|\mathbf{z}_j \| }-w_{ij}\right) ^2 + \lambda \|\mathbf{Z}-\mathbf{B}\|^2_F
			\label{l_final}
		\end{aligned}
	\end{equation}
	
	\subsection{Optimization}
	Our DSAH consists of two main parts and since the AND is not end-to-end, we separate the optimization process into two alternate steps. First, we fix the similarity matrix $\mathbf{W}$ and optimize the network parameter $\Theta$ by back propagation (BP) with a mini-batch sampling. 
	\begin{equation}
		\Theta \leftarrow \Theta-\eta\nabla_\Theta(\mathcal{L}) 
		\label{bq}
	\end{equation}
	where $\eta$ is the learning rate and $\nabla_\Theta$ represents a derivative of $\Theta$.
	
	Second, we fix the $\Theta$ and update the similarity matrix $\mathbf{W}$ according to Eq.~\ref{update}. 
	Once the model training is completed, we can obtain the discrete hash code $\mathbf{B}\in \{-1,+1\}^{n\times l}$ through Eq.~\ref{bij}.
	The detailed algorithm of our proposed DSAH is described in Algorithm \ref{alg1}.
	
	\begin{algorithm}[!t]
		\caption{\textbf{Deep Self-Adaptive Hashing~(DSAH)}} 
		\label{alg1} 
		\begin{algorithmic}[1]
			\REQUIRE Image set $\mathbf{X}$, hash code length $l$, temperature $\tau$, weight coefficient $\lambda$, factor $\gamma$, rounds $R$, epochs per round $T$, learning rate $\eta$
			\STATE Initialize the network parameters ${\mathcal{F}}(\cdot, \Theta)$.
			\STATE Initialize the semantic similarity matrices $\mathbf{W}^0\in \{+1,-1\}^{n\times n}$ by Eq.~\ref{w0}.
			\FOR{$r=1$ to $R$}
			\STATE \textit{// Hash Function Learning with PIC}
			\FOR{$t=1$ to $T$}
			\STATE Update the $\mathbf{Z}$ via Eq.~\ref{zij}.
			\STATE Update the $\mathbf{S}$ via Eq.~\ref{sij}.
			\STATE Update the $\mathbf{P}$ via Eq.~\ref{pij}.
			\STATE Update the $\Theta$ via Eq.~\ref{bq}.
			\ENDFOR
			\STATE \textit{// Similarity Updating by AND}
			\STATE Update the $\mathbf{W}^{r}$ via Eq.~\ref{update}.
			\ENDFOR
			\STATE Obtain the $\mathbf{B}$ via Eq.~\ref{bij}.
			\ENSURE Hash codes set $\mathbf{B}$
		\end{algorithmic} 
	\end{algorithm}
	
	\begin{table*}[!t]
		\small
		\centering
		\begin{tabu}{l|l|llll|llll|llll}
			\hline
			&           & \multicolumn{4}{c|}{\textbf{CIFAR-10}} & \multicolumn{4}{c|}{\textbf{FLICKR25K}} & \multicolumn{4}{c}{\textbf{NUS-WIDE}} \\ \hline
			Method  & Reference & 16 bits    & 32 bits    & 64 bits  & 128 bits  & 16 bits     & 32 bits     & 64 bits & 128 bits   & 16 bits     & 32 bits    & 64 bits & 128 bits   \\ \hline 
			LSH+VGG~\cite{andoni2006near} & STOC-02     &   0.177    &   0.192    &  0.261  &  0.304  &    0.596  &  0.619   &  0.650  & 0.666  &  0.385   & 0.455  & 0.446  & 0.567  \\
			SH+VGG~\cite{weiss2009spectral}  & NeurIPS-09    &  0.254  &  0.248     &  0.229  & 0.293  &    0.661  &  0.608   &  0.606  & 0.614 &  0.508   & 0.449  & 0.441  & 0.505  \\
			ITQ+VGG~\cite{gong2012iterative} & PAMI-13    &  0.269  &  0.295     &  0.316  &  0.350 &    0.709  &  0.696   &  0.684   & 0.720 &  0.519   & 0.576  & 0.598 & 0.651   \\
			AGH+VGG~\cite{liu2011hashing} & ICML-11    &    0.397   &  0.428     &  0.441 &  0.435  &    0.744  &  0.735   &  0.771  & 0.703 &  0.563   & 0.698  & 0.725 &  0.722  \\
			SP+VGG~\cite{Xia_2015_CVPR}  & CVPR-15    &    0.280    &  0.343     &  0.365 &  0.406  &    0.726  &  0.705   &  0.713 & 0.703  &  0.581   & 0.603  & 0.673  & 0.681  \\
			SGH+VGG~\cite{dai2017stochastic} & ICML-17    &  0.286  &  0.320     &  0.347   & 0.395   &  0.608   &  0.657   &  0.693  &  0.689 & 0.463  & 0.588  & 0.638   & 0.670  \\ \hline
			GH~\cite{su2018greedy}      & NeurIPS-18    &   0.355      &  0.424     &  0.419  & 0.416  &    0.702  &  0.732   &  0.753  & 0.760 &  0.599   & 0.657  & 0.695  & 0.712  \\
			SSDH~\cite{yang2018semantic}    & IJCAI-18   & 0.241    &  0.239     &  0.256   & 0.246 &    0.710  &  0.696   &  0.737 & 0.760  &  0.542   & 0.629  & 0.635 &  0.606  \\
			BGAN~\cite{song2018binary}    & AAAI-18    &   0.535    &  0.575     &  0.587  & 0.591  &    0.766  &  0.770   &  0.795  & 0.802 &  0.719   & 0.745  & 0.761  & 0.759  \\
			MLS$^3$RDUH~\cite{tu2020mls3rduh}     & IJCAI-20    &   0.562   &    0.588   &   0.595  & 0.582   &   0.797  & 0.809  &  0.809   & 0.804 &   0.730 & 0.754 & 0.764 & 0.769  \\
			TBH~\cite{shen2020auto}     & CVPR-20    &   0.432      &  0.459     &  0.455 &  0.474  &    0.779  &  0.794   &  0.797  & 0.799  &  0.678   & 0.717  & 0.729   & 0.739  \\ \hline
			\textbf{DSAH}    & \textbf{Proposed}  &  \textbf{0.596}   &   \textbf{0.617}  &  \textbf{0.622} & \textbf{0.635} &   \textbf{0.805}    &   \textbf{0.816}     &  \textbf{0.831}  &  \textbf{0.836} &    \textbf{0.747}    &   \textbf{0.769}    & \textbf{0.787}  &  \textbf{0.793}	\\\hline 
		\end{tabu}
		\caption{MAP@5000 results on CIFAR10, FLICKR25K and NUS-WIDE. The best result is shown in boldface.}
		\label{sota}
		\centering
	\end{table*}
	
	\section{Experiments}
	In this section, we conduct experiments on several public benchmark datasets and evaluate our DSAH method. Particularly, we design the experiments to study the following questions:
	
	\begin{itemize}
		\item \textbf{Q1}: Compared with the state-of-the-art unsupervised hashing algorithms, does our DSAH method outperform them?
		\item \textbf{Q2}: How do the components: PIC and AND, affect the performance of DSAH separately? How do they work?
		\item \textbf{Q3}: Is DSAH sensitive to hyper-parameters?
		\item \textbf{Q4}: How about the efficiency of DSAH?
		\item \textbf{Q5}: What is the qualitative result of DSAH?
	\end{itemize}
	
	\subsection{Datasets}
	Following most hashing-based retrieval methods~\cite{song2018binary, deng2019unsupervised, qin2020unsupervised}, we adopt the following three widely used public benchmark datasets to evaluate the model performance: CIFAR-10, FLICKR25K and NUS-WIDE. The basic information is listed in Table.\ref{data}, while the setting details be introduced as follows:
	\begin{table}[H]
		\footnotesize
		\begin{tabular}{|l|l|l|l|}
			\hline
			\textbf{Dataset}     & \textbf{CIFAR-10} & \textbf{FLICKR25K} & \textbf{NUS-WIDE} \\ \hline
			Multi-Label & $\times$  & $\checkmark$ & $\checkmark$       \\ \hline
			\# Images   & 60,000   & 25,000    & 269,648  \\ \hline
			\# Classes  & 10       & 24        & 21       \\ \hline
			\# Train Set & 10,000  & 10,000 & 10,500       \\ \hline
			\# Query Set & 1,000  & 1,000 & 2,100       \\ \hline
			\# Retrieval Set & 59,000  & 24,000 & 267,548      \\ \hline
		\end{tabular}
		\caption{Characteristics of evaluation datasets.}
		\label{data}
		\vspace{-2.0em}
	\end{table}
	
	\textbf{CIFAR-10}
	~\cite{krizhevsky2009learning}:~
	Followed the setting of \cite{song2018binary}, we randomly selected 100 images for each class as the query set, 1,000 in total. Then we used the remaining images as the retrieval set, among them, we randomly selected 1,000 images per class as the training set.
	
	\textbf{FLICKR25K}
	~\cite{huiskes2008mir}:~
	We randomly selected 1000 images as query set and the remaining images were left for retrieval set. In the retrieval set, we randomly chose 10,000 images as the training set.
	
	\textbf{NUS-WIDE}
	~\cite{chua2009nus}:~
	According to the setting in~\cite{zhu2016deep}, 
	we selected 21 most frequent classes from the dataset and each class contains at least 5,000 related images. We randomly selected 2100 images as the query set and the remaining images were used as a retrieval set. We also randomly selected 10,500 images for training.
	
	For the latter two multi-label datasets, if the retrieved image shares at least one label with the query image, it is considered to be associated with the query image.	
	
	\subsection{Experiment Setup}
	\textbf{Metrics}: 
	Similar to~\cite{wang2017survey, shen2018unsupervised, yang2018semantic}, we employed four widely used evaluation metrics to evaluate the retrieval performance, including Mean Average Precision~(\textbf{MAP}), Precision of the top N retrieved images~(\textbf{Precision@N}), Precision curve~(\textbf{Precision Curve}), and Precision-Recall curves~(\textbf{PR Curve}). 
	For a fair comparison, all the methods used the same training and query sets. 
	
	\textbf{Baseline methods}:
	We compared our method with eleven unsupervised hashing methods, including six shallow hashing methods: \textbf{LSH}~(\citet{andoni2006near}), \textbf{SH}~(\citet{weiss2009spectral}), \textbf{ITQ}~(\citet{gong2012iterative}), \textbf{AGH}~(\citet{liu2011hashing}), \textbf{SP}~(\citet{Xia_2015_CVPR}), \textbf{SGH}~(\citet{dai2017stochastic}) and five deep hashing methods: \textbf{GH}~(\citet{su2018greedy}), \textbf{SSDH}~(\citet{yang2018semantic}), \textbf{BGAN}~(\citet{song2018binary}), \textbf{MLS$^3$RDUH}~(\citet{tu2020mls3rduh}) and \textbf{TBH}~(\citet{shen2020auto}). The parameters and architectures of the compared methods were according to the setting provided by the original papers. For a fair comparison, all shallow hashing methods used 4096-dimensional features generated by the relu7 layer of VGG19~\cite{simonyan2014very} pre-trained on ImageNet, 
	as same as the deep features used in the five deep hashing methods during their similarity structures construction.
	
	\textbf{Implementation details}:
	Our DSAH is implemented based on the Tensorflow framework, while all the experiments are conducted on a workstation with an Intel  l5-8500 CPU, and an Nvidia GTX2080 GPU.
	In the initial similarity construction~(Sec~\ref{Initialization}), the $k_1$ and $k_2$ are set to the same value 500. 
	When training the network, each training image was resized to $224\times 224$ as input. 
	We set $\lambda$ equal to 10 and adopt the adam optimization with learning rate $\eta$ equal to 1e-4, and the batch size was set to 50. 
	The $\tau$ is set to 1, while the $\gamma$ is set to \{1, 0, 1\} for CIFAR-10, FLICKR25K and NUS-WIDE datasets respectively. And the $R$ and $T$ are set to 3 and 10 respectively. 
	
	\begin{figure*}[!t]
		\centering
		\subfigure[PR Curve of CIFAR-10@64 bits]{
			\includegraphics[width=0.22\textwidth]{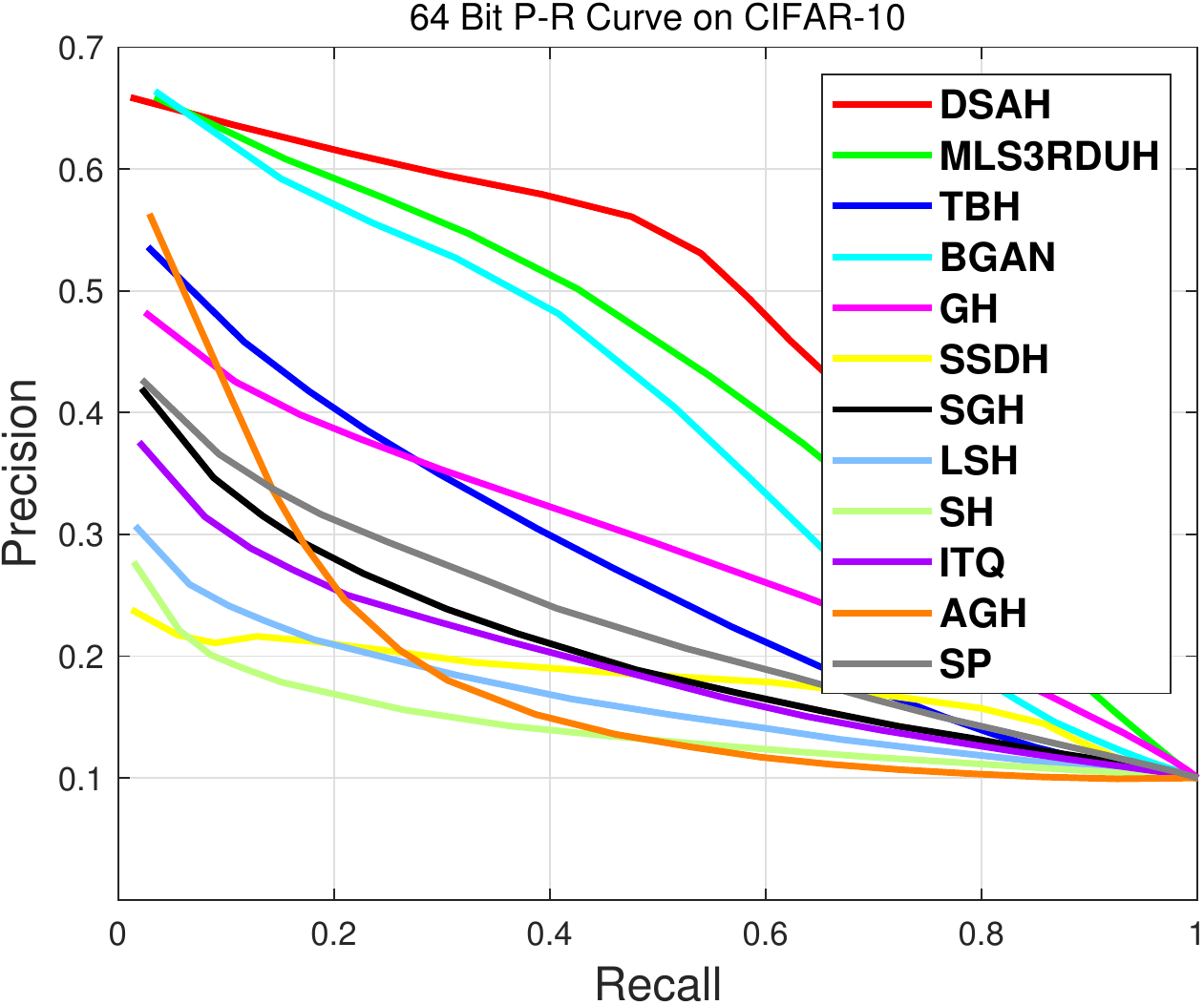}
		}
		\subfigure[Precision Curve of CIFAR-10@64 bits]{
			\includegraphics[width=0.22\textwidth]{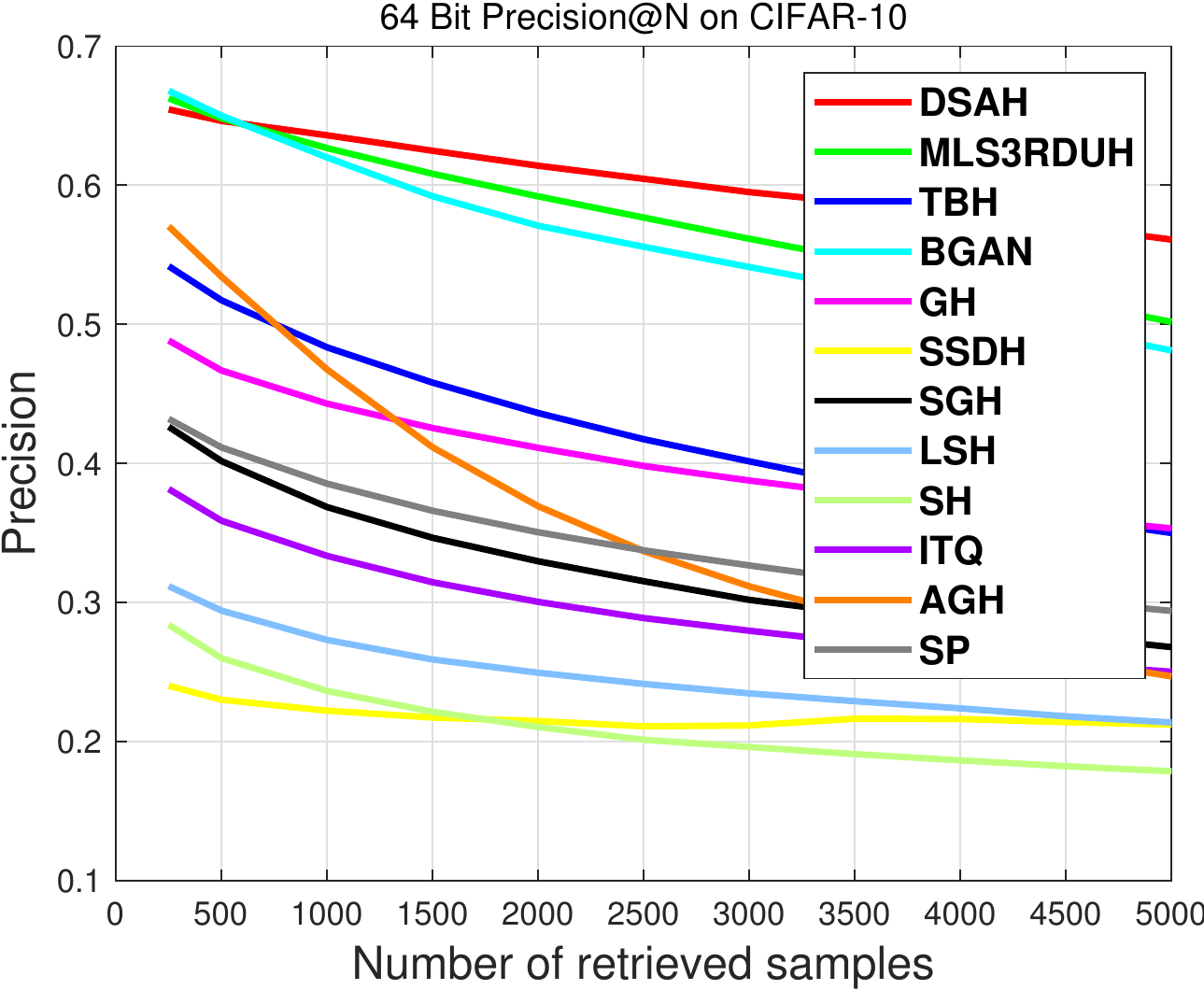}
		}
		\subfigure[PR Curve of CIFAR-10@128 bits]{
			\includegraphics[width=0.22\textwidth]{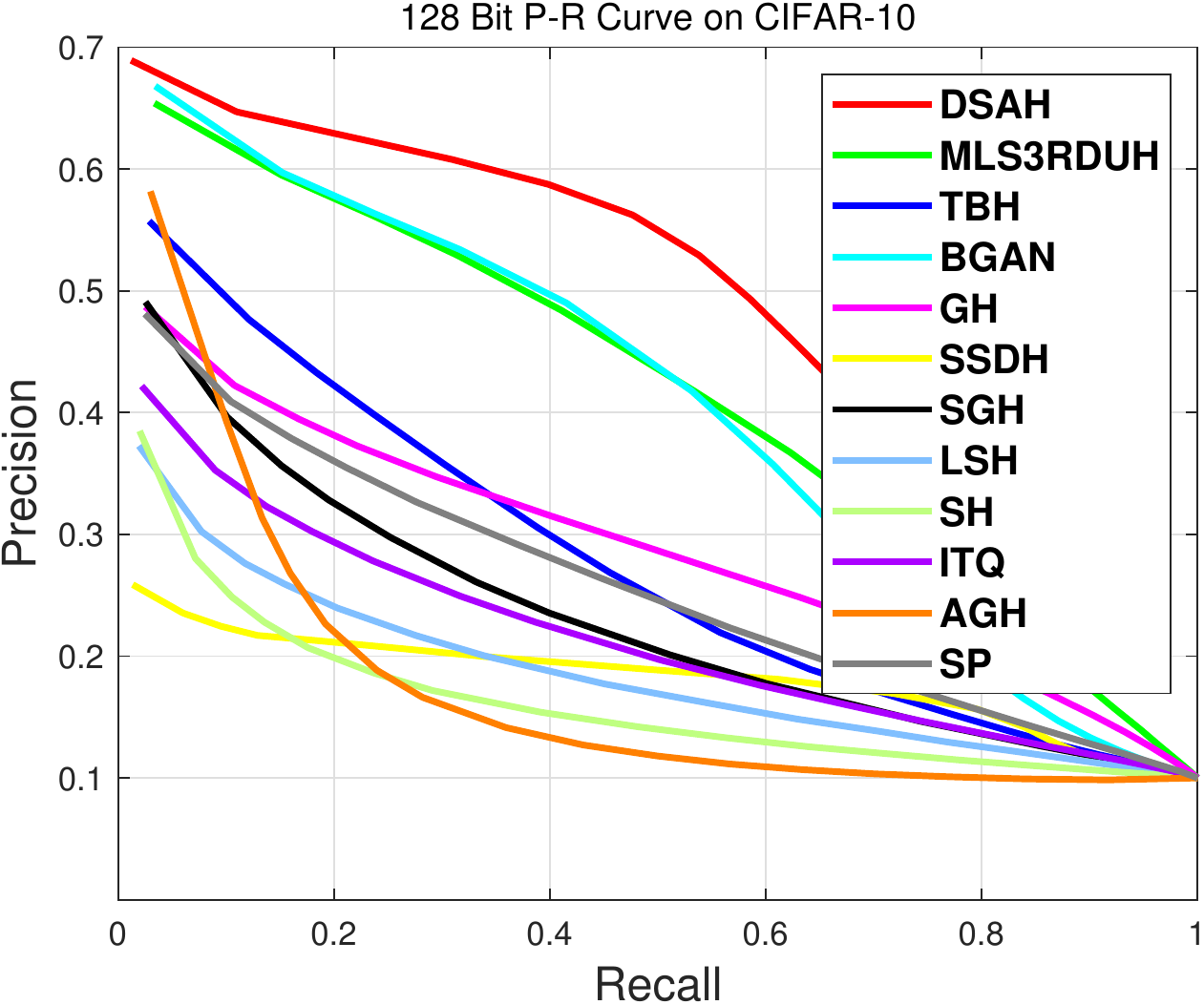}
		}
		\subfigure[Precision of CIFAR-10@128 bits]{
			\includegraphics[width=0.22\textwidth]{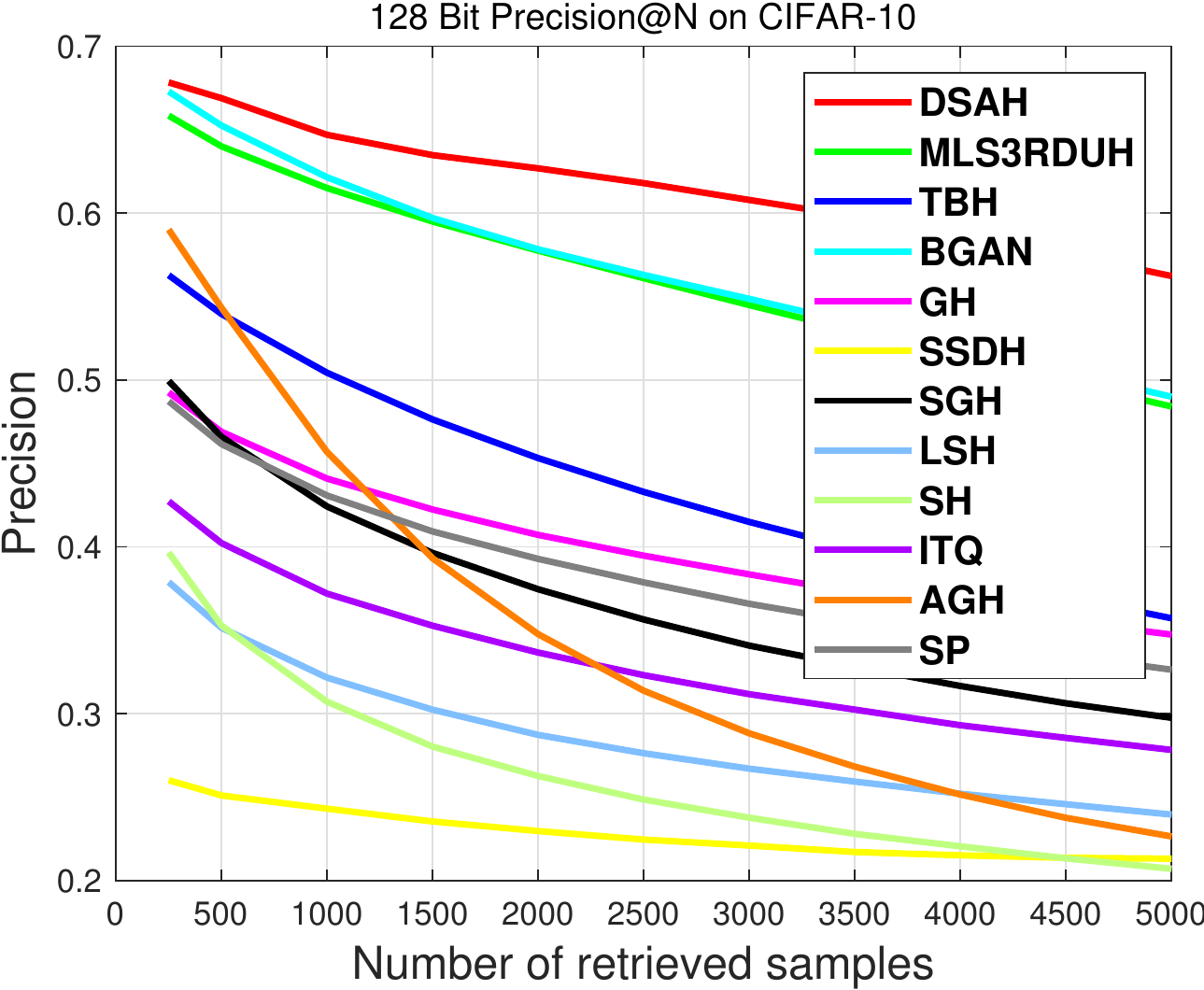}
		}
		\subfigure[PR Curve of FLICKR25K@64 bits]{
			\includegraphics[width=0.22\textwidth]{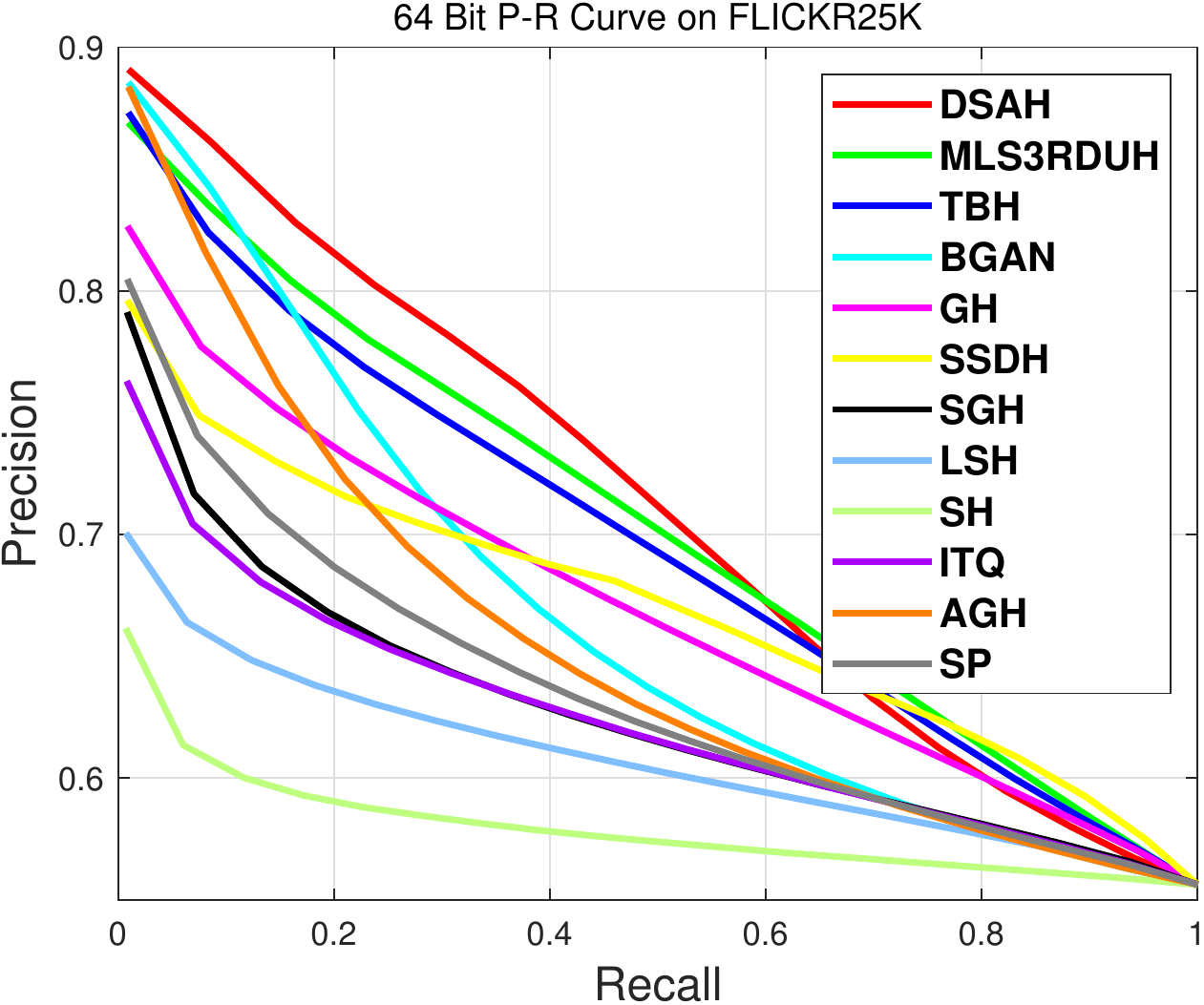}
		}
		\subfigure[Precision of FLICKR25K@64 bits]{
			\includegraphics[width=0.22\textwidth]{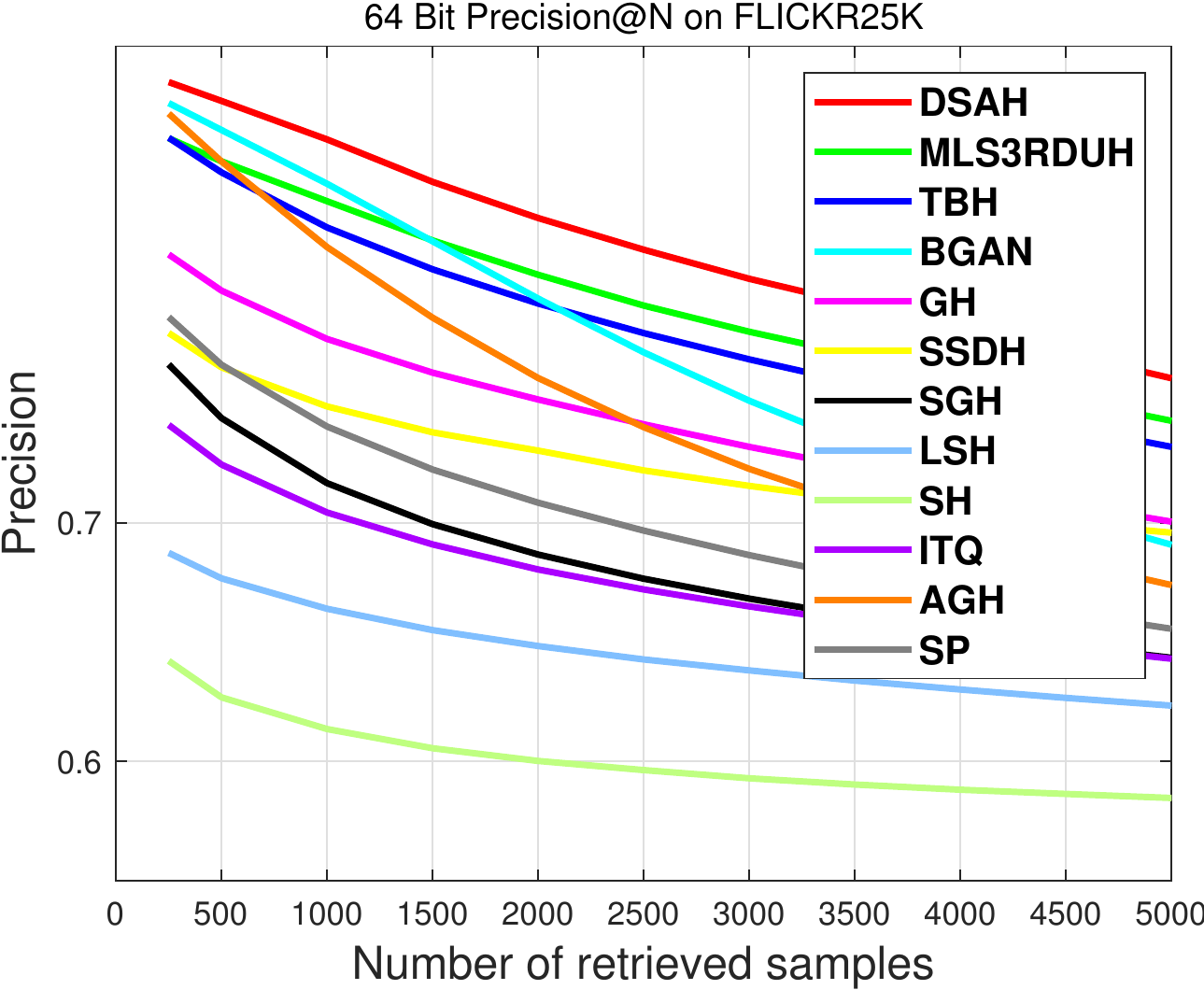}
		}
		\subfigure[PR Curve of FLICKR25K@128 bits]{
			\includegraphics[width=0.22\textwidth]{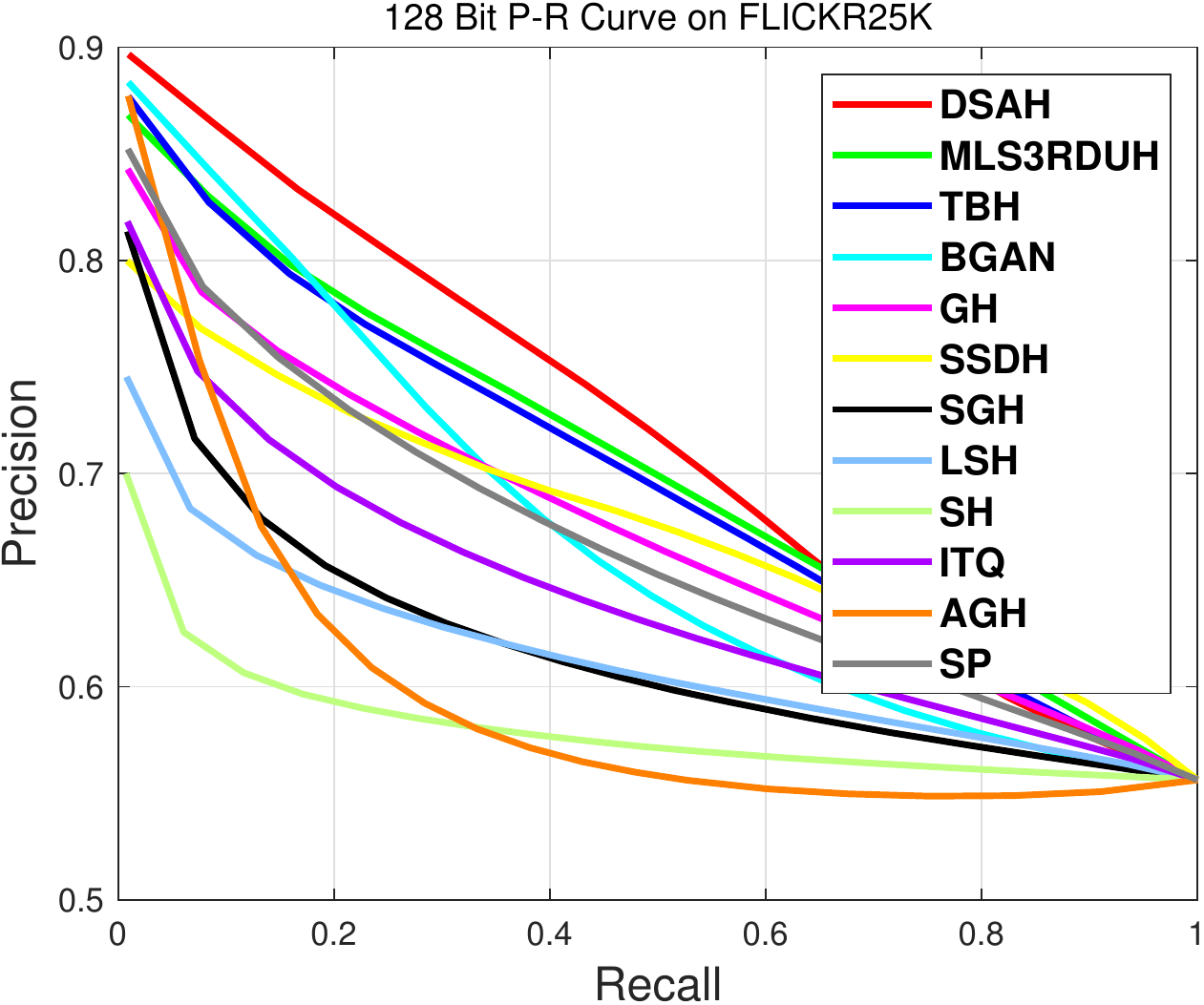}
		}
		\subfigure[Precision of FLICKR25K@128 bits]{
			\includegraphics[width=0.22\textwidth]{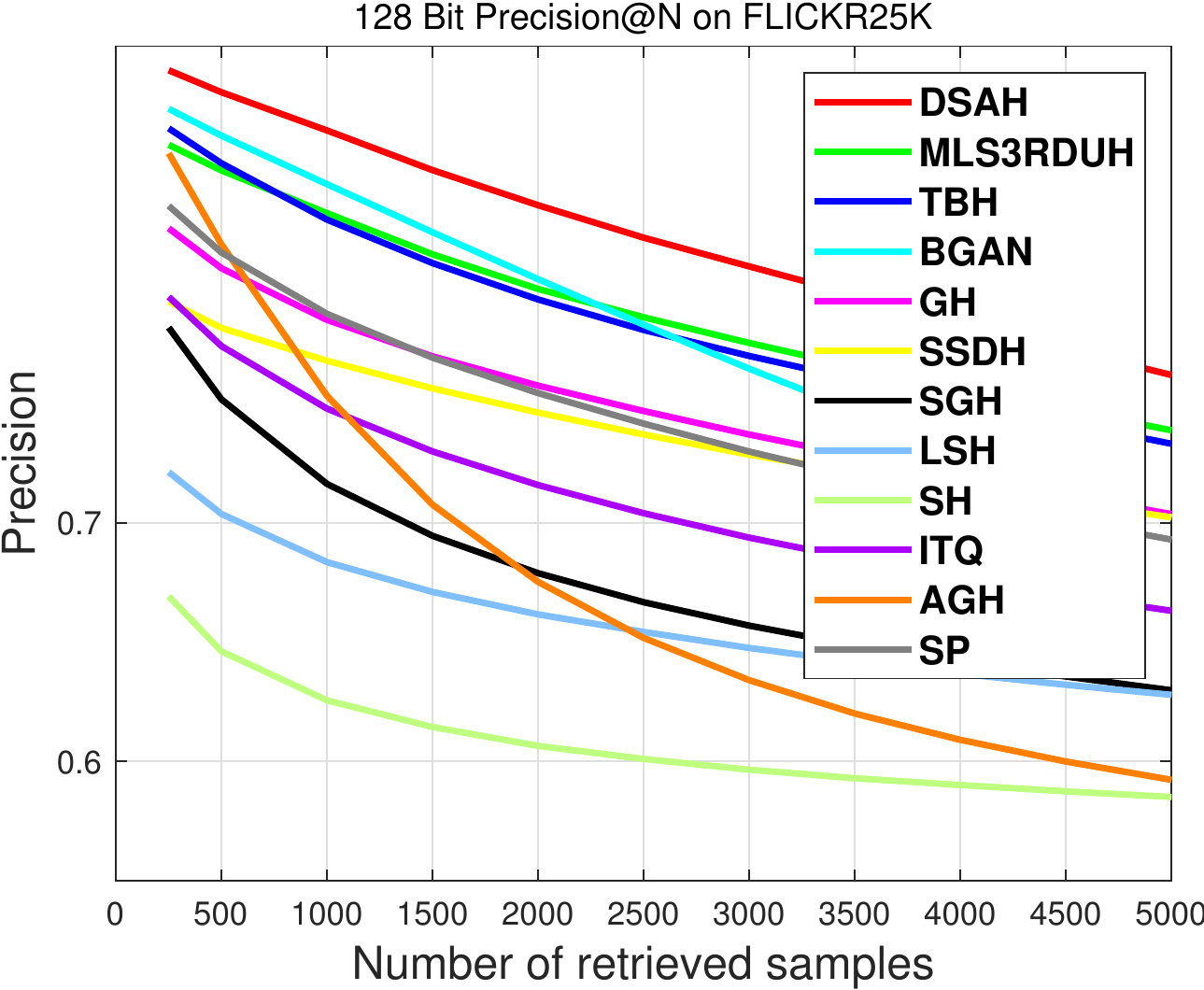}
		}
		\subfigure[PR Curve of NUS-WIDE@64 bits]{
			\includegraphics[width=0.22\textwidth]{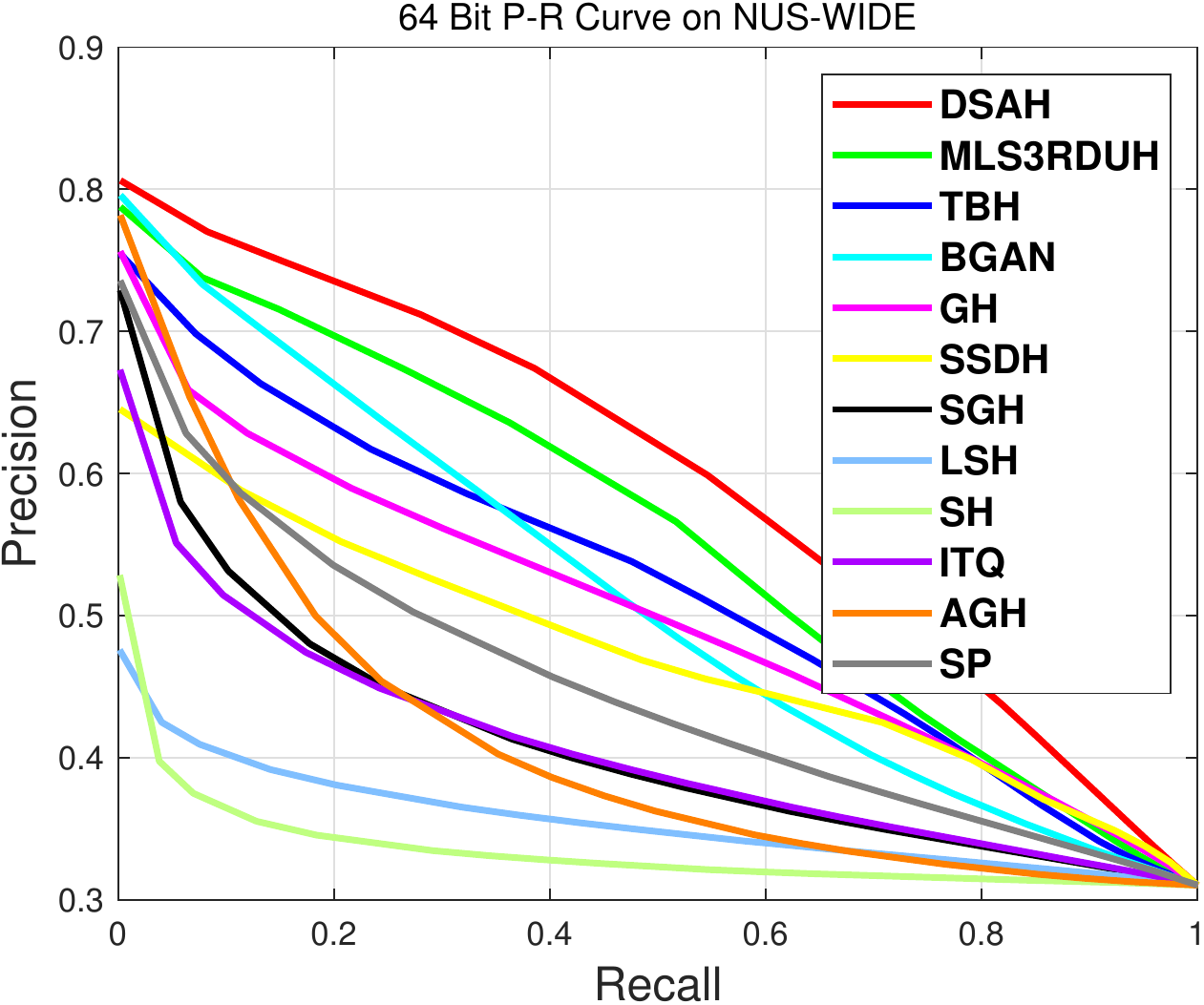}
		}
		\subfigure[Precision of NUS-WIDE@64 bits]{
			\includegraphics[width=0.22\textwidth]{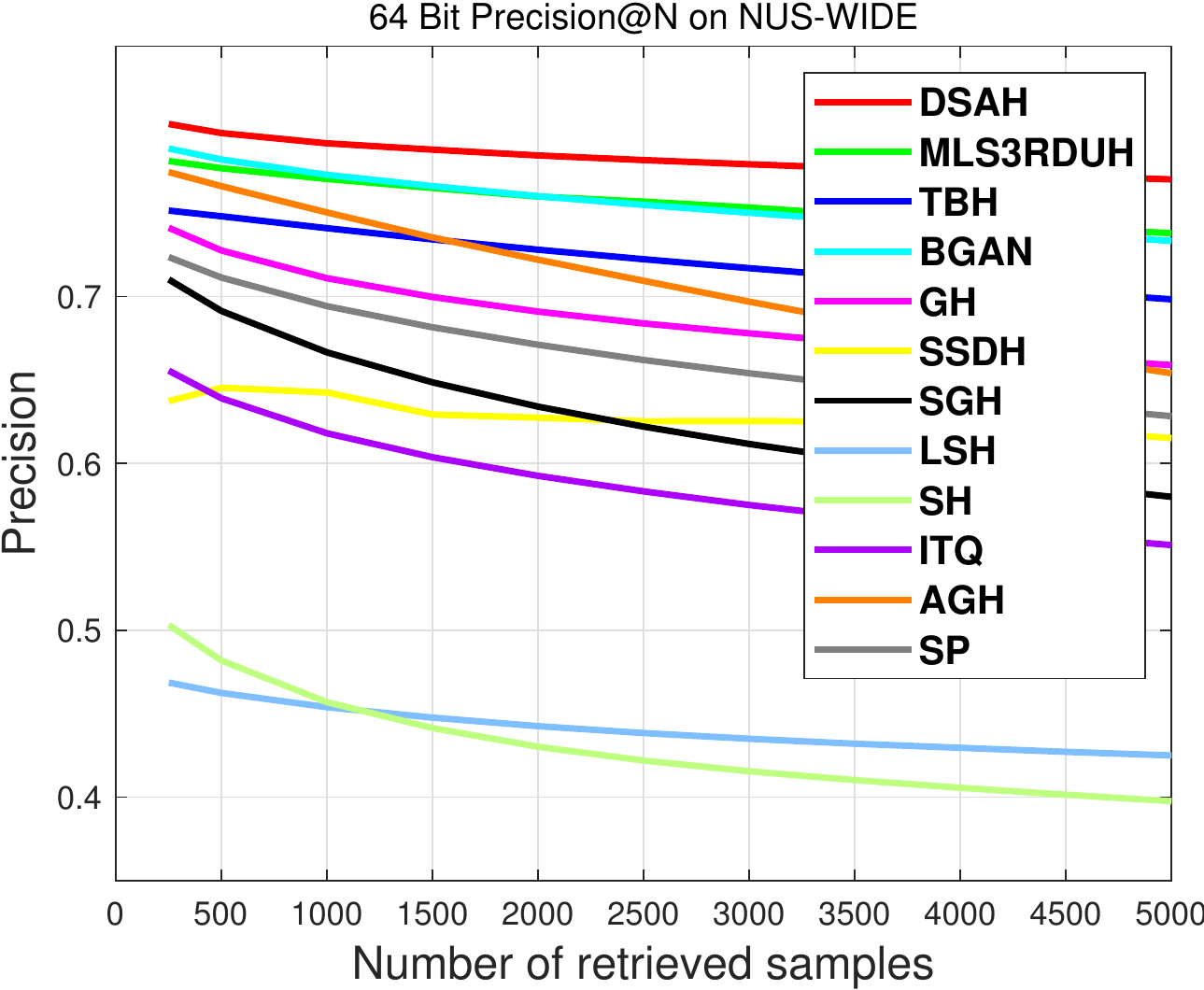}
		}
		\subfigure[PR Curve of NUS-WIDE@128 bits]{
			\includegraphics[width=0.22\textwidth]{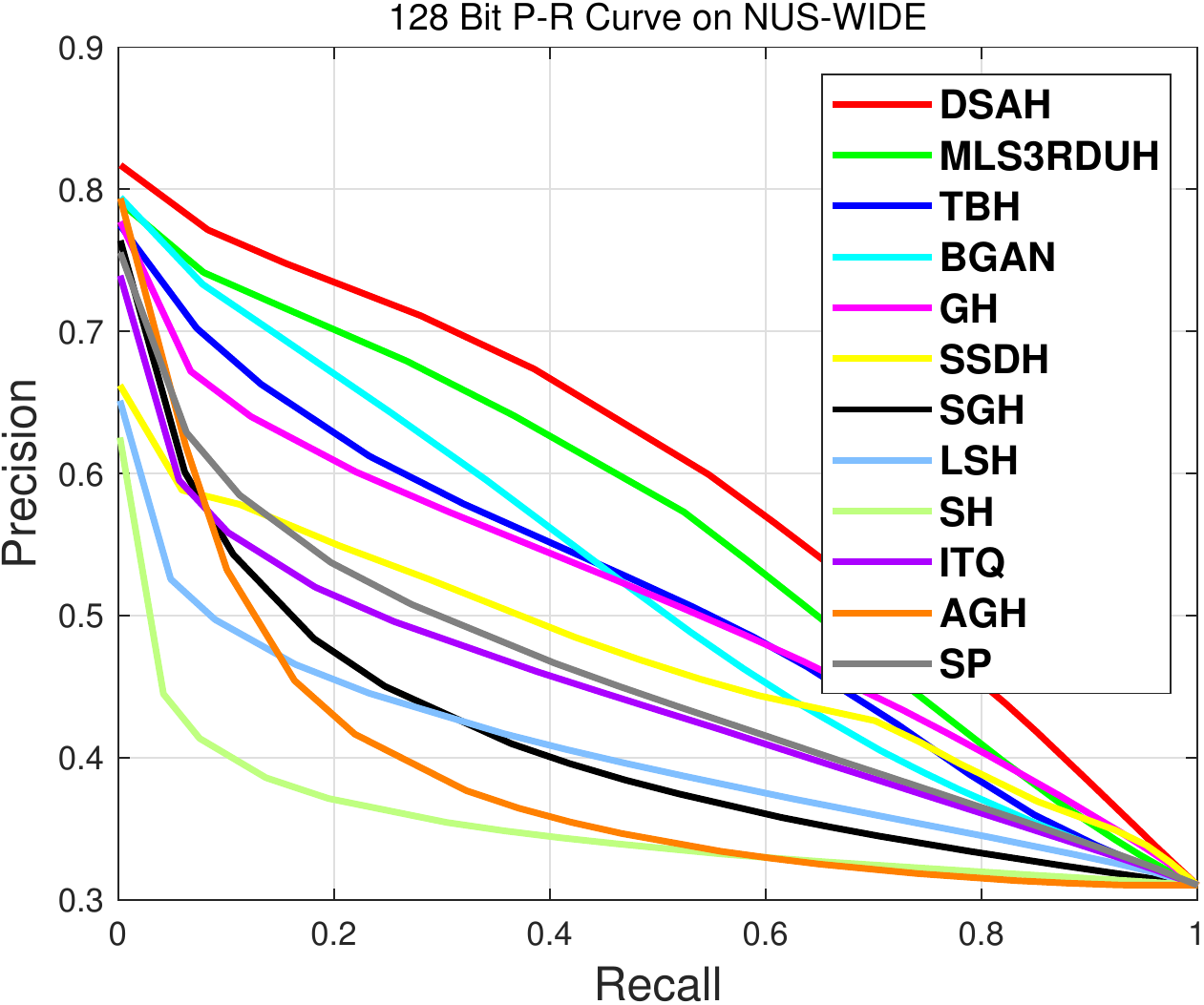}
		}
		\subfigure[Precision of NUS-WIDE@128 bits]{
			\includegraphics[width=0.22\textwidth]{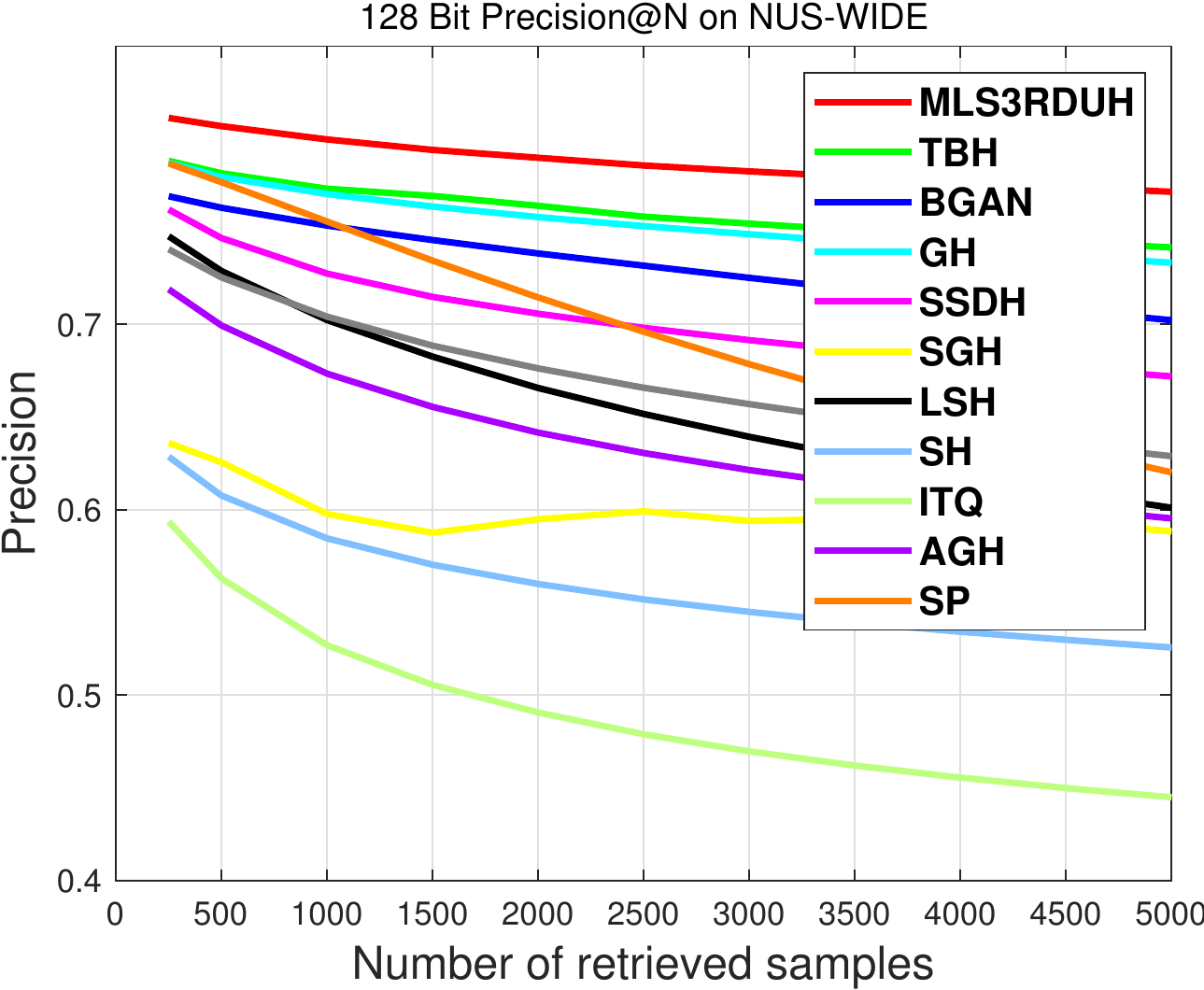}
		}
		\caption{Precision-Recall (PR) curves and Precision$@$N curves on the three datasets for 64 and 128 bits length.}
		\label{curve}
	\end{figure*}
	
	\subsection{Comparison Results and Discussions~(Q1)}
	\subsubsection{MAP \& Precision}
	The performance of our DSAH and baseline methods in terms of MAP@5000 are shown in Table.~\ref{sota}. 
	We can observe that our proposed method significantly outperforms other methods on the three datasets. On CIFAR-10, DSAH obtains an increase of 6.1\% (16 bits), 4.9\% (32 bits), 4.5\% (64 bits),  9.1\% (128 bits) comparing with the best competitor MLS$^3$RDUH respectively.
	On FLICKR-25K and NUS-WIDE, DSAH achieves average 8.7\% and 2.7\% improvement compared to the MLS$^3$RDUH. 
	
	In the practical scenarios, we pay more attention to the top retrieval responses. Thus, we further evaluate the performance on the precision  of the top-100 for each method. The results are demonstrated in Table.~\ref{pre100}, where we display the result of the shortest (16 bits) and longest (128 bits) binary codes. On FLICKR25K, DSAH achieves 1.5\% (16 bits) and 2.3\% (128 bits) increases comparing with TBH. On NUS-WIDE, DSAH is improved by 1.7\% (16 bits) and 2.9\% (128 bits) compared to BGAN. 
	Therefore, both the Table.\ref{sota} and Table.\ref{pre100} demonstrate that DSAH achieves better performance compared to previous unsupervised hashing methods.
	
	\begin{table}[!h]
		\footnotesize
		\begin{tabular}{l|ll|ll|ll}
			\hline
			\textbf{Dataset} & \multicolumn{2}{c|}{\textbf{CIFAR-10}} & \multicolumn{2}{c|}{\textbf{FLICKR25K}} & \multicolumn{2}{c}{\textbf{NUS-WIDE}} \\ \hline
			Methods & 16 bits      & 128 bits      & 16 bits       & 128 bits       & 16 bits       & 128 bits      \\ \hline
			LSH+VGG~\cite{andoni2006near}&   0.175   &  0.411  &  0.608   &  0.745  &  0.393   &   0.651    \\
			SH+VGG~\cite{weiss2009spectral}&  0.291  &  0.446  &  0.715   &  0.700  &  0.518   &   0.625    \\
			ITQ+VGG~\cite{gong2012iterative}& 0.283  &  0.458  &  0.760   &  0.818  &  0.545   &   0.739    \\
			AGH+VGG~\cite{liu2011hashing}&   0.444   &  0.617  &  0.794   &  0.877  &  0.572   &   0.794    \\
			SP+VGG~\cite{Xia_2015_CVPR}&   0.284   &   0.514   &  0.788   &  0.852  &  0.601   &   0.756    \\ 
			SGH+VGG~\cite{dai2017stochastic}&  0.285    & 0.536  &   0.622   & 0.814 & 0.457  & 0.764		\\ \hline
			GH~\cite{su2018greedy} &    0.410    &   0.519     &    0.773   &  0.843 & 0.651  & 0.777 \\
			SSDH~\cite{yang2018semantic} &   0.216      &   0.269   &  0.753  & 0.800 & 0.595 & 0.662 \\
			BGAN~\cite{song2018binary}&       0.591    &  0.691    &   0.839  & 0.884 & 0.749 & 0.794 \\
			MLS$^3$RDUH~\cite{tu2020mls3rduh}&   \textbf{0.618}  & 0.677  &  0.851  & 0.868 & 0.748 & 0.791 \\
			TBH~\cite{shen2020auto}&    0.499    &  0.586   &  0.849   &  0.877	&   0.702 &  0.775 \\ \hline
			\textbf{DSAH}&      0.616        &  \textbf{0.691}            &        \textbf{0.862}       &  \textbf{0.897}		&     \textbf{0.762} & \textbf{0.817}          \\ \hline
		\end{tabular}
		\caption{Precision@100 results on CIFAR10, FLICKR25K and NUS-WIDE. The best result is shown in boldface.}
		\label{pre100}
	\end{table}
	
	\subsubsection{Precision Curve \& PR Curve}
	To further illustrate the effectiveness of DSAH, we display the Precision curve and PR curve of 64 and 128 bits in Figure.~\ref{curve}. 
	The PR Curve Figure.\ref{curve}~(a)(c)(e)(g)(i)(k) clearly displays the precision at different recall values, which is a good representation of overall performance. In general, a larger area under the PR curve indicates better performance. It can be seen that our PR curve covers more areas in most cases, which means that when the precision is equal, the proposed method will recall more related images; when the number of recall related images is equal, the proposed method has higher precision. Thus, DSAH yields a stable and superior performance.
	Similar to the PR Curve setting, we display the Precision Curves in Figure.\ref{curve}~(b)(d)(f)(h)(j)(l). It can be seen that the precisions of our methods is relatively higher than the precisions of other methods in most cases. which indicates that given a fixed number of retrieval samples, our method can obtain higher precision outperform other methods.
	
	\subsection{Ablation Study~(Q2)}
	Since our method consists of two major components: PIC and AND, we further verify their effectiveness.
	
	\subsubsection{Effect of PIC}
	In order to validate the efficiency of PIC, we design the following variants with different  $a_{ij}$ in Eq.~\ref{l1}: 
	\begin{itemize}
		\item \textbf{PIC$^0$}: $a_{ij}$ is equal to constant 1, and it would degenerate to Eq.~\ref{CNNH}.  This could be treated as a baseline.
		\item \textbf{PIC}: $a_{ij}$ is equal to $-\log(p_{ij})$. This is our proposed PIC.
		\item \textbf{PIC$^-$}: $a_{ij}$ is equal to $-\log(1-p_{ij})$. This could be regarded as an opposite version of our PIC.
	\end{itemize}
	
	\begin{figure}[H]
		\centering
		\subfigure[CIFAR-10]{
			\includegraphics[width=0.145\textwidth]{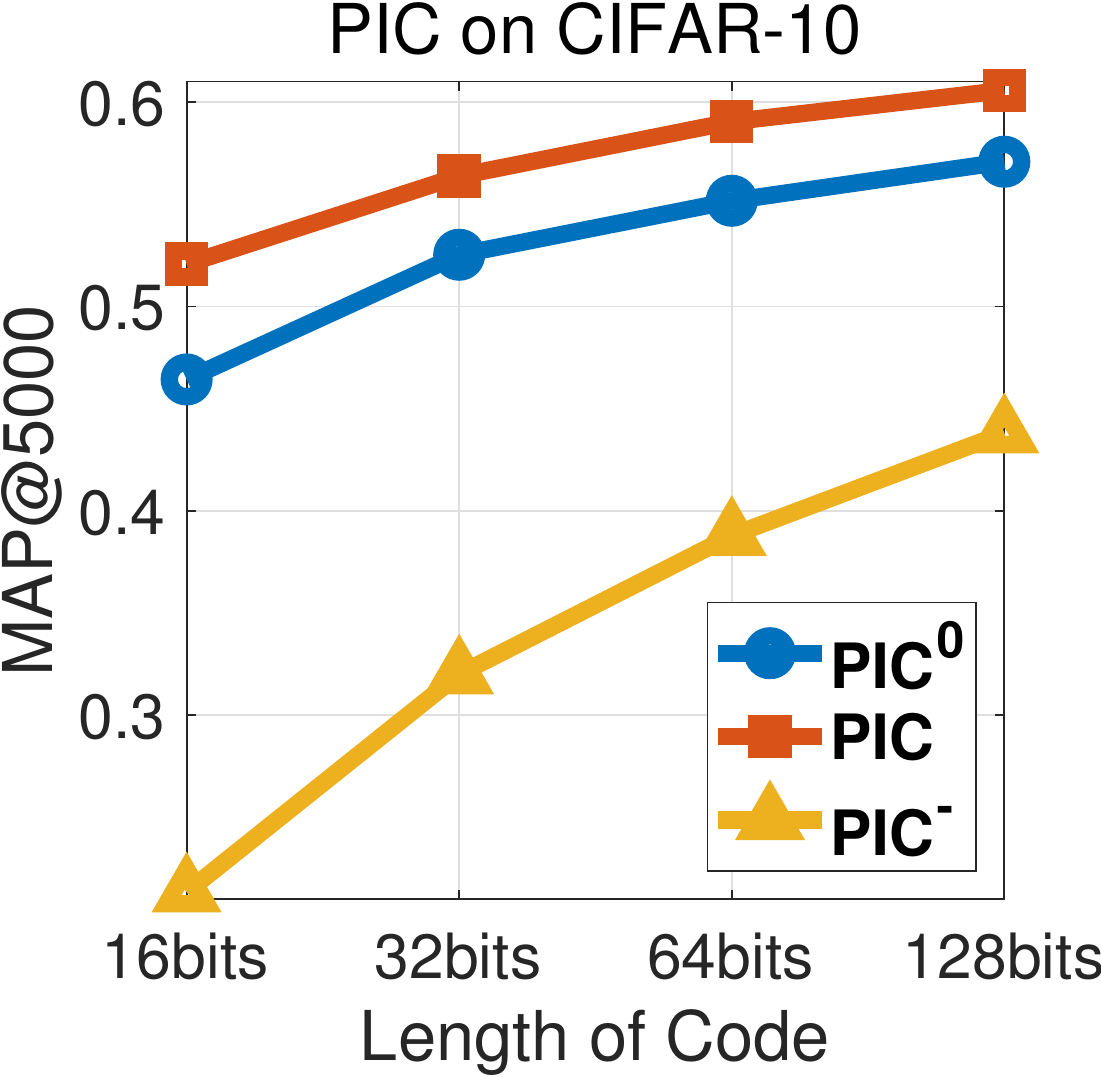}
		}
		\subfigure[FLICKR25K]{
			\includegraphics[width=0.145\textwidth]{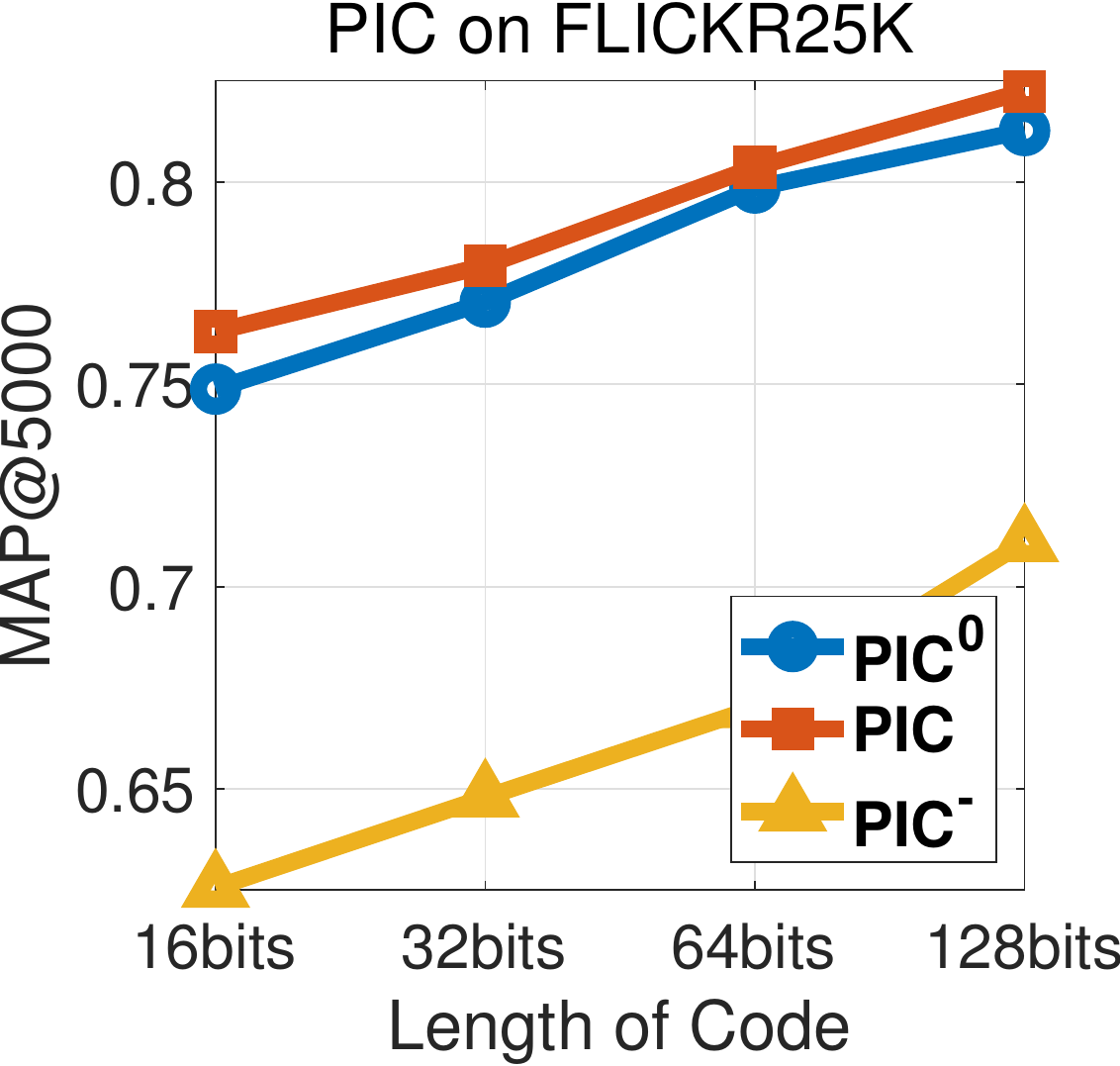}
		}
		\subfigure[NUS-WIDE]{
			\includegraphics[width=0.145\textwidth]{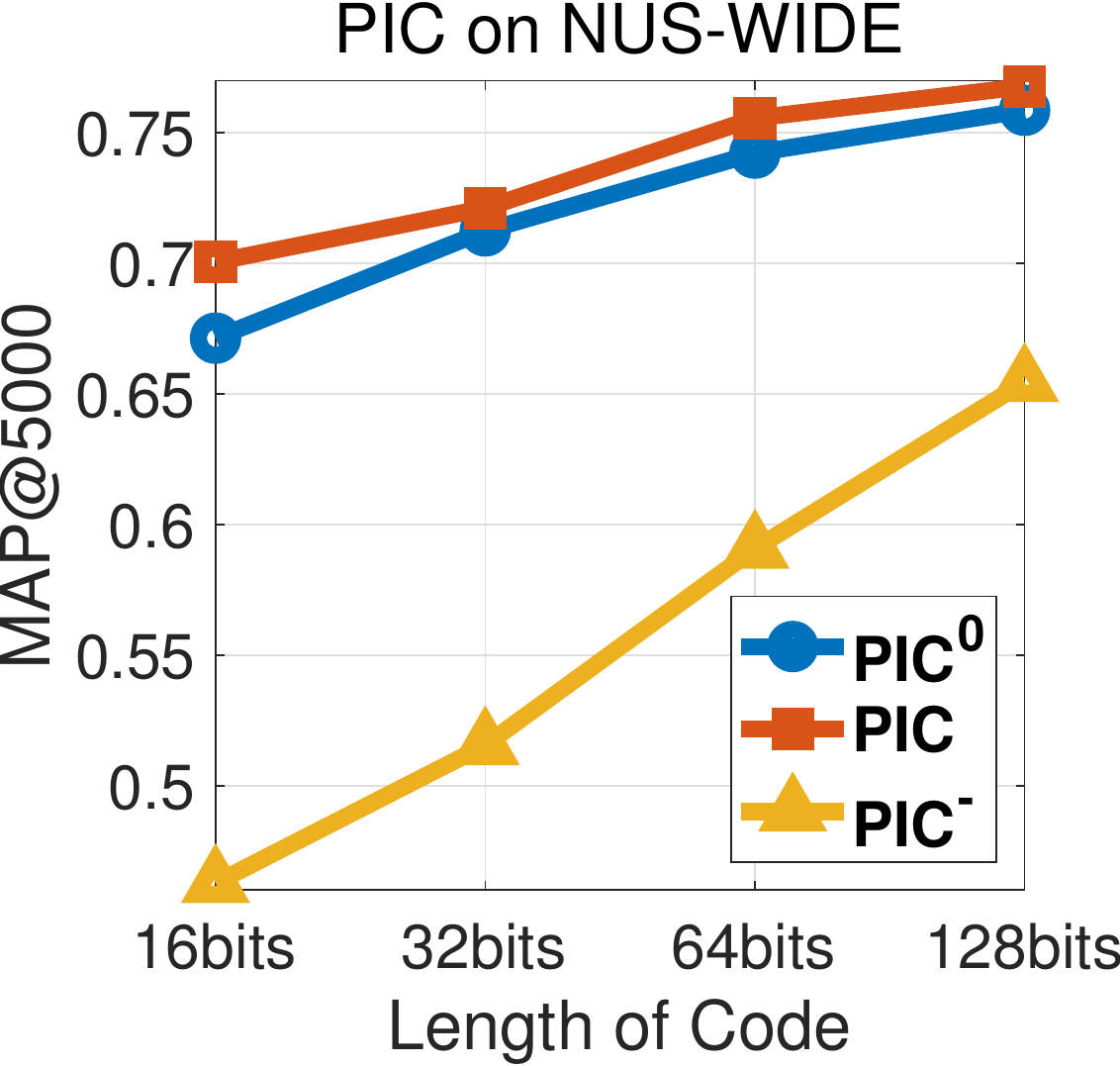}
		}
		\caption{Effect of PIC on three datasets with MAP@5000, the code lengths varying from 16 to 128.}
		\label{Q1_pic_a}
	\end{figure} 
	
	We report the experimental results on the three datasets in Figure \ref{Q1_pic_a}, which are evaluated with MAP@5000.
	As shown in Figure.~\ref{Q1_pic_a}~(a)-(c), we could observe that: Firstly, compared with baseline PIC$^0$, introducing PIC weights is able to obtain average 8.2\%, 1.2\%, 2.1\% MAP improvements on CIFAR-10, FLICKR25K and NUS-WIDE datasets, respectively.
	These results indicate that PIC indeed improves the discriminative power of our model. 
	Secondly, the performance degradation of the third PIC$^-$ shows that assigning large weights to similar data pairs will degenerate the performance, 
	which indicates that dissimilar data pairs should receive more attention rather than similar data pairs.
	These results validate the effectiveness of our proposed weighting method PIC.
	
\begin{figure}[!h]
		\centering
		\subfigure[Heatmap of $\mathbf{S}$]{
			\includegraphics[width=0.2\textwidth]{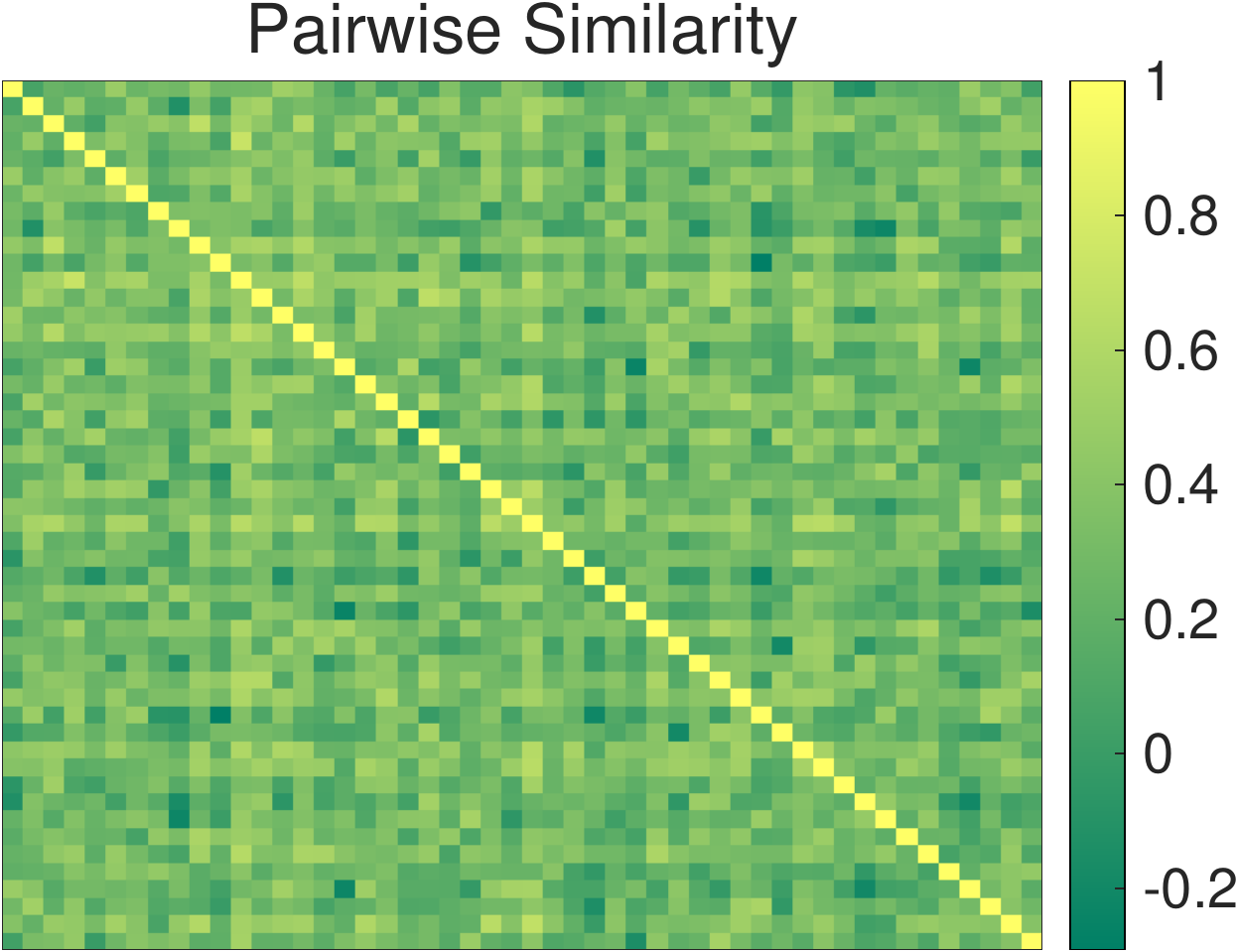}
		}
		\subfigure[Heatmap of $\mathbf{A}$]{
			\includegraphics[width=0.2\textwidth]{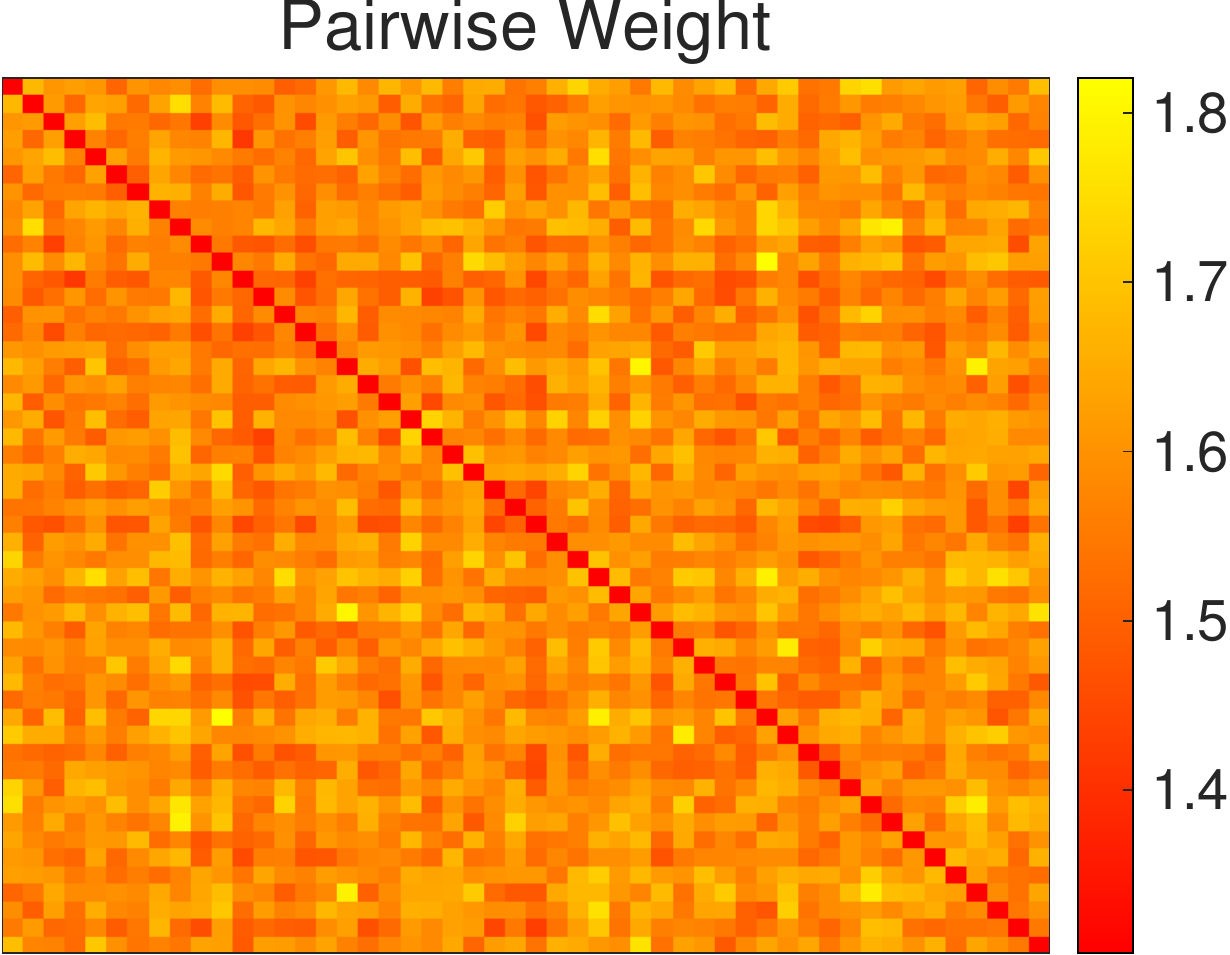}
		}
		\caption{The heatmap of (a)~pairwise similarity $s_{ij}$ and (b)~corresponding PIC weight $a_{ij}$ within a batch size.}
		\label{Q1_pic_b}
	\end{figure}

	\textbf{How PIC works.}
	To further understand PIC, we display the heatmap of pairwise similarity $s_{ij}$ and their PIC weights $a_{ij}$ within a batch size in Figure.~\ref{Q1_pic_b}. It is shown that those data pairs with higher similarity, especially in the diagonal~(self-similar), would receive a lower weight after PIC assignment and dissimilar pairs will reach higher weights.
	
	\subsubsection{Effect of AND}
	
	To reveal the effectiveness of AND, we conduct an experiment to compare the following variants: 
	\begin{itemize}
		\item \textbf{AND$^{0}$}: Training via Eq.~\ref{CNNH} with $\mathbf{W}^0$, which is baseline.
		\item \textbf{AND}: Training via Eq.~\ref{CNNH} with updated $\mathbf{W}^r$ via AND.
		\item \textbf{AND+PIC}: Training via Eq.~\ref{l1} with updated $\mathbf{W}^r$ via AND.
	\end{itemize}
	
\begin{figure}[!h]
		\centering
		\subfigure[CIFAR-10]{
			\includegraphics[width=0.145\textwidth]{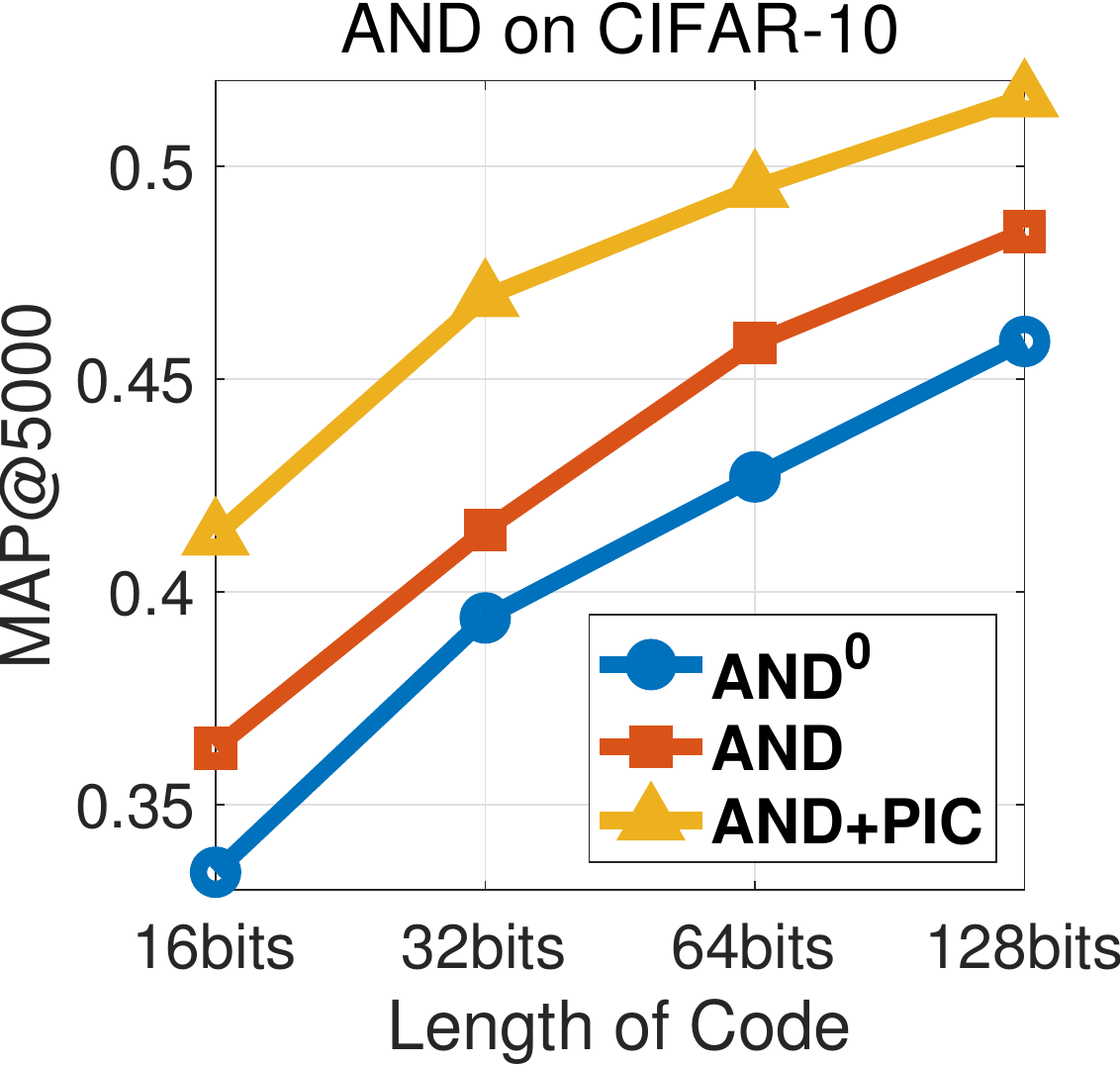}
		}
		\subfigure[FLICKR25K]{
			\includegraphics[width=0.145\textwidth]{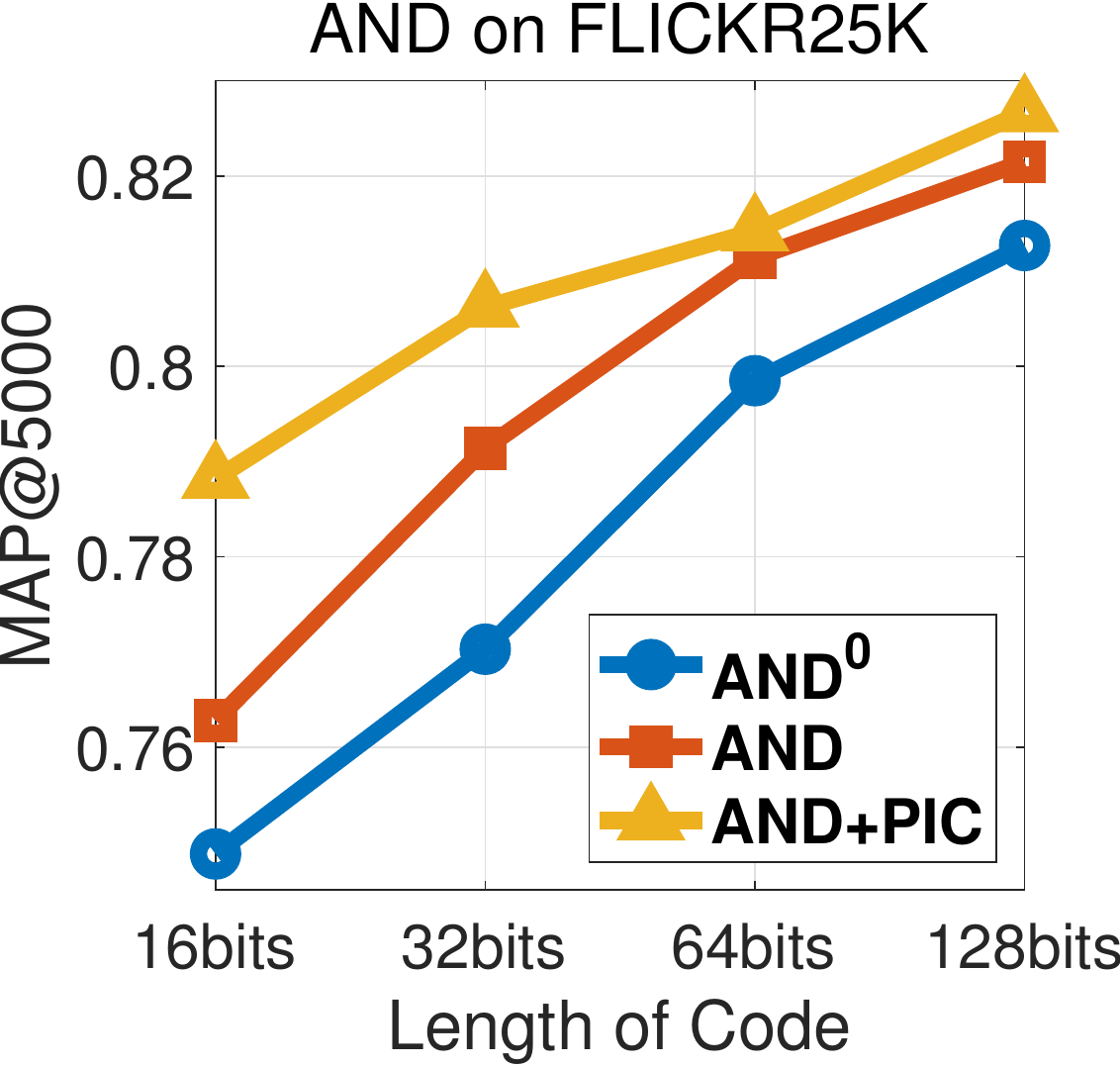}
		}
		\subfigure[NUS-WIDE]{
			\includegraphics[width=0.145\textwidth]{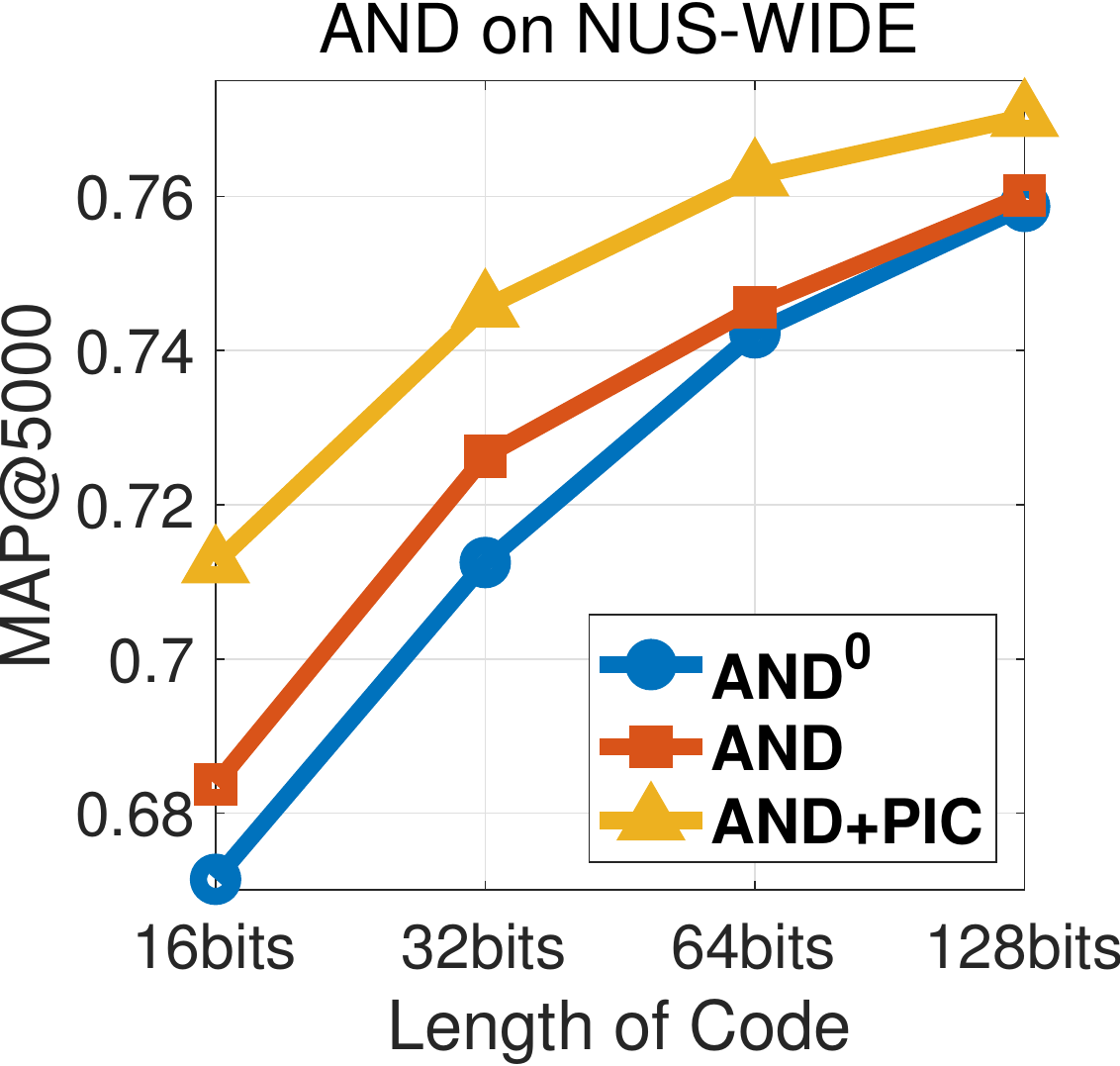}
		}
		\caption{Effect of AND on three datasets with MAP@5000, the code length varying from 16 to 128.}
		\label{Q1_and_a}
	\end{figure}

	We report the results evaluated with MAP@5000 on the three datasets in Figure.~\ref{Q1_and_a}.
	By employing the AND, the baseline model can be generally improved, achieving 1.7\%, 1.8\%, 1.0\% average improvements on three datasets, respectively. 
	Furthermore, the performance can be greatly improved by introducing both PIC and AND, resulting in 8.9\%, 3.4\% and 3.8\% average improvements over baseline in three datasets. The above experiments reveal the effectiveness of PIC and AND of DSAH.
	
\begin{figure}[H]
		\centering
		\subfigure[$\left( n^+, m\right) $ w.r.t $R$]{
			\includegraphics[width=0.22\textwidth]{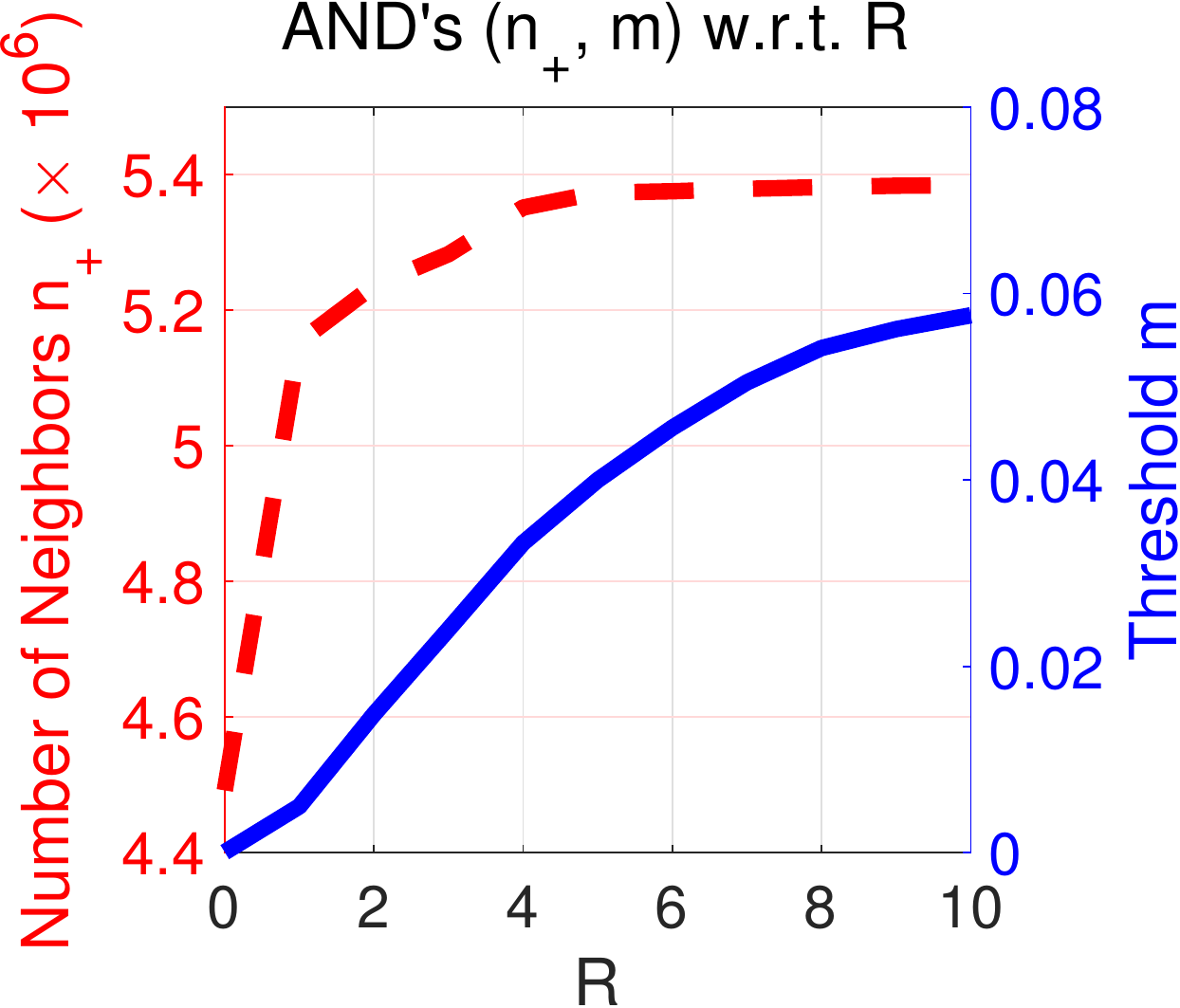}
		}
		\subfigure[$\left( \text{MAP@5000}, F_w\right) $ w.r.t $R$]{
			\includegraphics[width=0.22\textwidth]{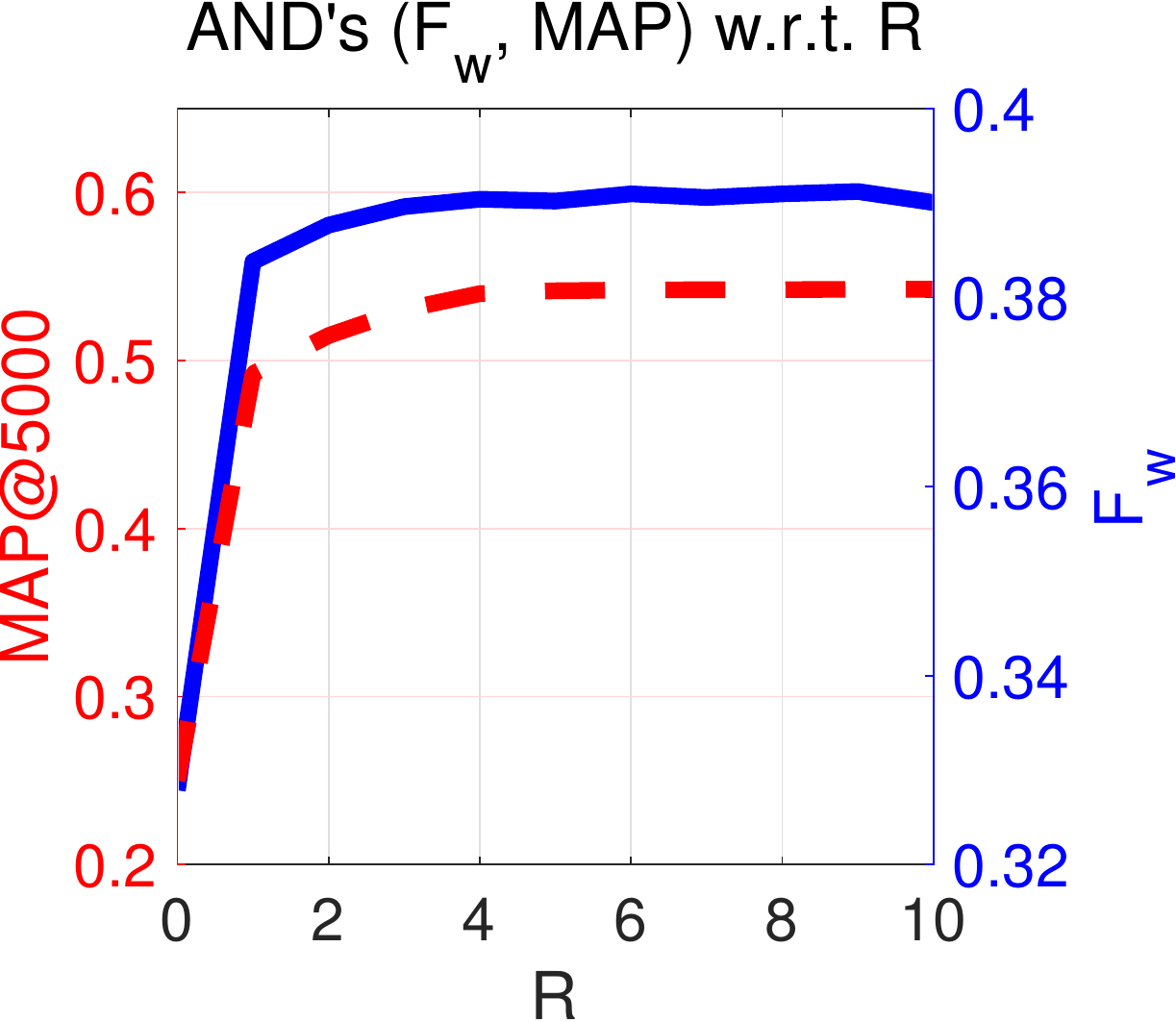}
		}
		\caption{The Mechanism of AND. We monitor the AND's $\left( n_+, m, MAP, F_w\right) $ w.r.t round $R$.}
		\label{Q1_and_b}
	\end{figure}

	\textbf{How AND works.}
	To better understand the mechanism of AND, we enlarge the round $R$ to 10 and record the changing of threshold $m$ and the number of neighbors pairs ${n_+}$. 
	Also, to monitor the quality of the updated similarity matrix $\mathbf{W}^r$, we compute an indicator matrix $\mathbf{G}\in \{-1, +1\}^{n\times n}$, which element $g_{ij}$ is equal to $+1$ iff data pairs $(i,j)$ belong to same class, otherwise equal to $-1$. 
	Next, we treat AND as a binary classification problem and $\mathbf{G}$ as ground-truth, and then design a metric ${F}_{w}$ to measure the quality of updated matrix $\mathbf{W}^r$, which is inspired by F-score:

	$$
	{F}_{w}=\frac{2}{\alpha^{-1}+\beta^{-1}}
	$$
	where $\alpha$ is equal to $\frac{\sum_{i=1}^n\sum_{j=1}^n\mathbf{1}(w^t_{ij}=g_{ij})}{\sum_{i=1}^n\sum_{j=1}^n\mathbf{1}(w^t_{ij}=+1)}$, denotes the precision of updated similarity matrix $\mathbf{W}^r$. And $\beta$ denotes the recall of $\mathbf{W}^r$, defined as $\frac{\sum_{i=1}^n\sum_{j=1}^n\mathbf{1}(w^t_{ij}=g_{ij})}{\sum_{i=1}^n\sum_{j=1}^n\mathbf{1}(g_{ij}=+1)}$. 
	In Figure.~\ref{Q1_and_b}. 
	We could see that in each round, additional data pairs are added to the neighbors set, and the number of similar data pairs converges eventually. This is mainly controlled by the increase of the adjustable threshold $m$. Notably, the climbing $F_w$ confirms the effect of AND, which indeed refines the quality of the similarity matrix and leads to the model's improvement, especially in the first updating.
	
	\subsection{Parameter Sensitivity~(Q3)}
	\subsubsection{Study of $\lambda$}
\begin{figure}[H]
		\centering
		\subfigure[Study of the $\lambda$.]{
			\includegraphics[width=0.2\textwidth]{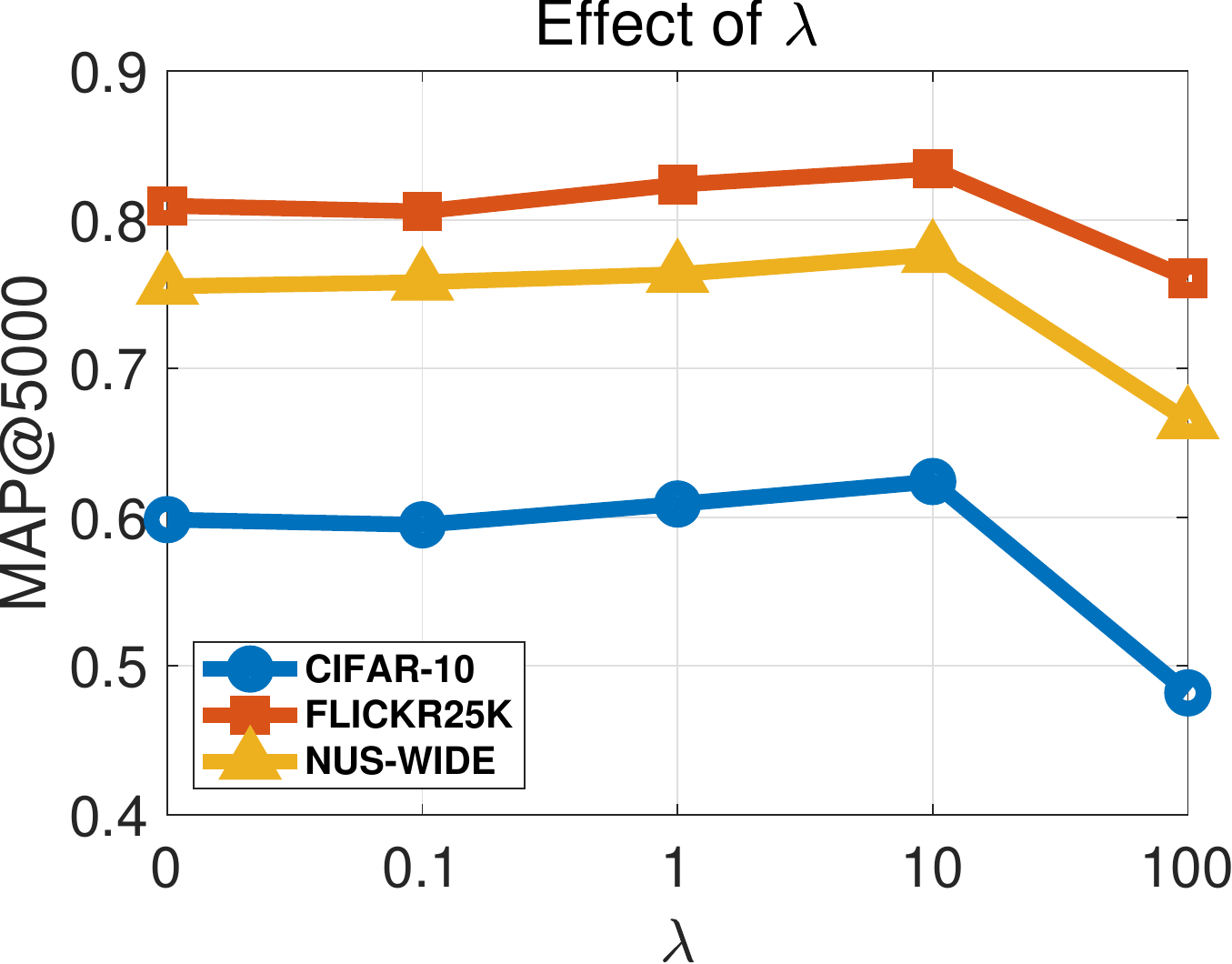}
		}
		\subfigure[Study of the $\tau$ of PIC.]{
			\includegraphics[width=0.2\textwidth]{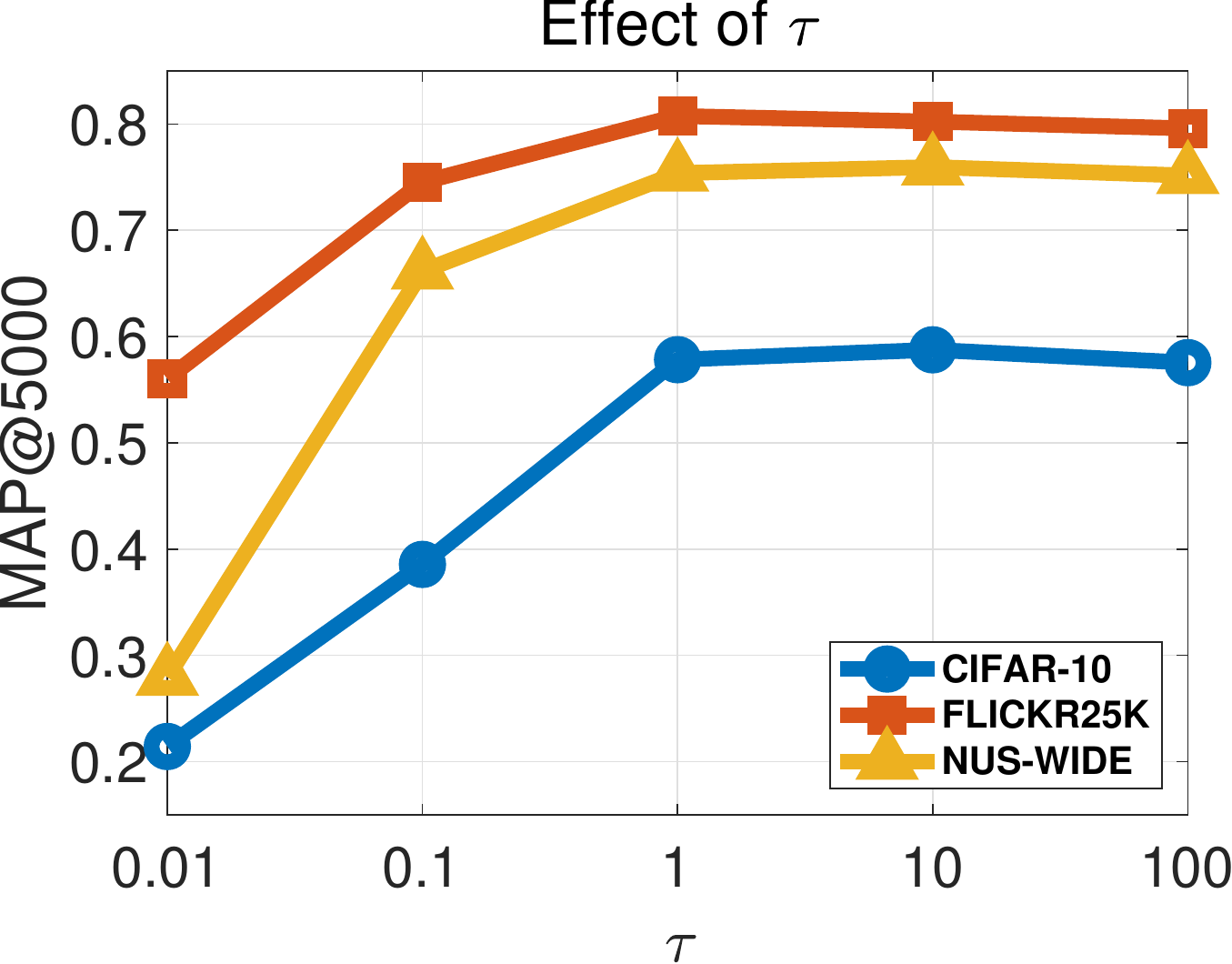}
		}
		\caption{MAP@5000 under (a)~different $\lambda$ and (b)~different $\tau$ of PIC, the code length is 64.}
		\label{PA}
	\end{figure}

	In Figure.~\ref{PA}~(a), we study the influence of the quantization effect with different $\lambda$ over three datasets, where the code length is 64. 
	From this figure, the performance will degrade when $\lambda$ is larger than 10.
	And the recommended value for $\lambda$ is 10, which would bring 4.3\%, 3.1\% and 2.8\% MAP improvements in CIFAR-10, FLICKR25K and NUS-WIDE, respectively.
	
	\subsubsection{Study of PIC}
	In Figure.~\ref{PA}~(b), we investigate the influence of hyper-parameter $\tau$ in PIC. It shows that the proposed method is sensitive and deteriorates when $\tau\in \{0.01, 0.1\}$. When $\tau$ is larger than 1, the model performance tends to stabilize and slightly decline, and the recommended value for $\tau$ is 1.0.
	
	\subsubsection{Study of AND}
\begin{figure}[H]
		\centering
		\subfigure[CIFAR-10: 64 bits]{
			\includegraphics[width=0.22\textwidth]{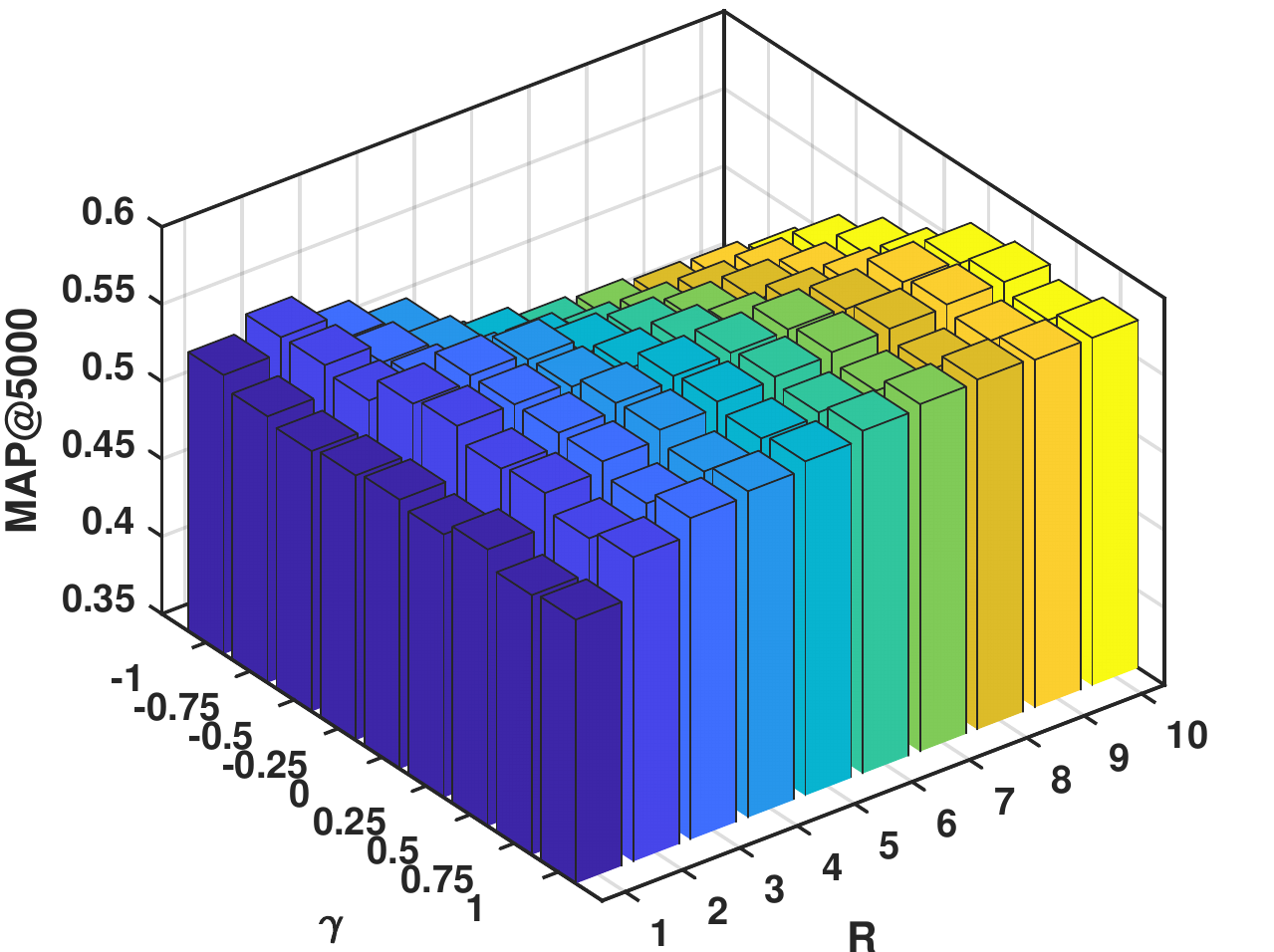}
		}
		\subfigure[FLICKR25K: 64 bits]{
			\includegraphics[width=0.22\textwidth]{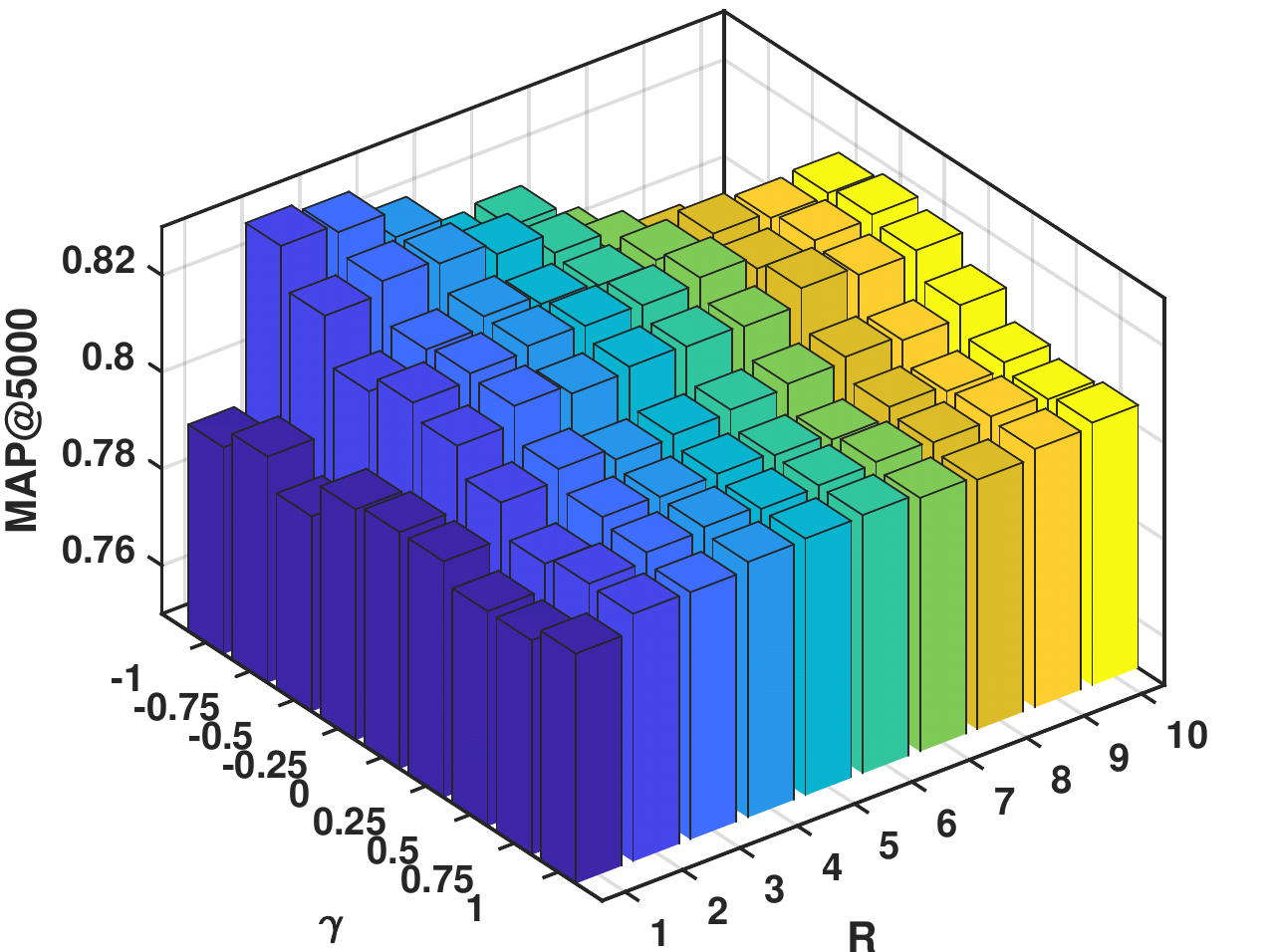}
		}
		\caption{Parameter Sensitivity of $(\gamma,R)$ in AND. The code length is set to 64.}
		\label{PA_and}
	\end{figure}

	In Figue.~\ref{PA_and}, we evaluate the effectiveness of different hyper-parameters $(\gamma, R)$ in AND on (a)~CIFAR-10 and (b)~FLICKR25K datasets, where $\gamma$ from -1.0 to 1.0 with a step of 0.25 and the maximum $R$ is set to 10, the code length is 64.
	The diagram in Figure.~\ref{PA_and}(a) indicates that MAP will increase with $R$ increases and then converges around 3 on CIFAR-10 dataset, and a large $\gamma$ ($\geq 0$) might be a good choice.
	On FLICKR25K, it is noteworthy that when $\gamma$ is smaller than $-0.5$, a significant increment would happen in the second round. This is due to the increase of similar pairs, but later proved that this threshold was too loose, leading to model degradation.
	
	\subsection{Efficiency Analysis~(Q4)}
	In Table.\ref{time}, we conduct an experiment to compare the training time of different deep methods on CIFAR-10 dataset.
	The results indicate that DSAH is efficient and only takes nearly one-third time of the fastest baseline SSDH; Also, DSAH achieves signiﬁcantly performance with only a fifth of MLS$^3$RDUH’s computation time.
	\begin{table}[!h]
		\small
		\begin{tabular}{l|c|c}
			\hline
			\textbf{Methods} & \textbf{MAP@5000}& \textbf{Training Time~(h)} \\ \hline
			SSDH~\cite{yang2018semantic} &   0.256  & 3.0   \\
			BGAN~\cite{song2018binary}  & 0.587&    5.0 \\
			MLS$^3$RDUH~\cite{tu2020mls3rduh} & 0.595 &    4.8 \\
			DSAH & 0.622   &  1.2  \\ \hline
		\end{tabular}
		\caption{Training time comparison among the unsupervised deep hashing methods.}
		\label{time}
	\end{table}
	
	\subsection{Qualitative Result~(Q5)}
	\subsubsection{t-SNE visualization}
\begin{figure}[!h]
		\setlength{\abovecaptionskip}{0.cm}
		\setlength{\belowcaptionskip}{-0.cm}
		\centering
		\subfigure[BGAN~\cite{song2018binary}]{
			\includegraphics[width=0.145\textwidth]{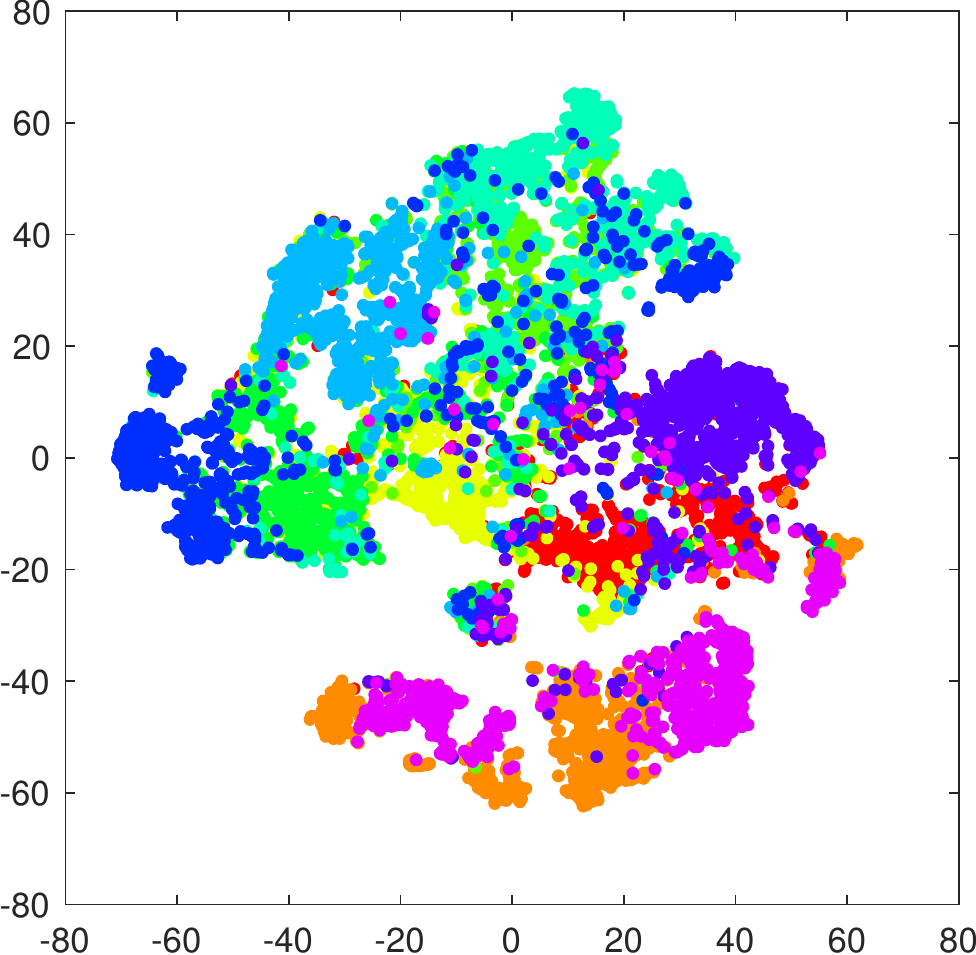}
		}
		\subfigure[MLS$^3$RDUH~\cite{tu2020mls3rduh}]{
			\includegraphics[width=0.145\textwidth]{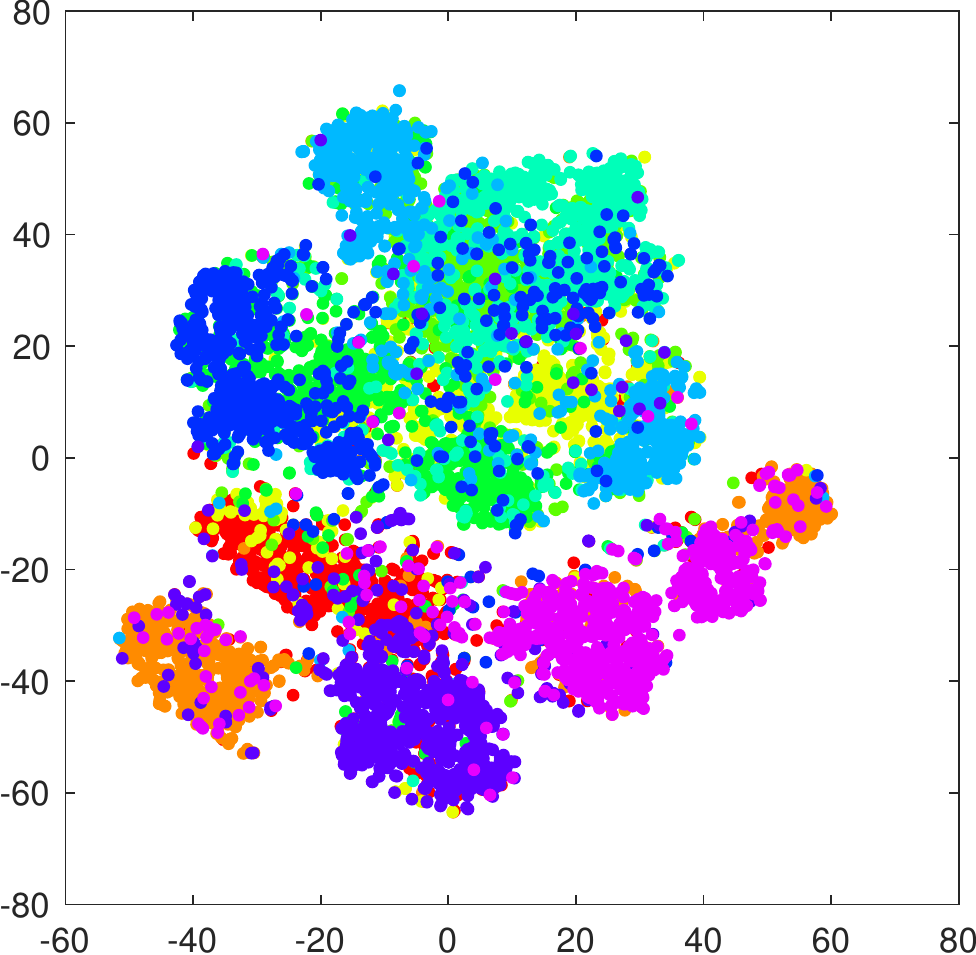}
		}
		\subfigure[DSAH]{
			\includegraphics[width=0.145\textwidth]{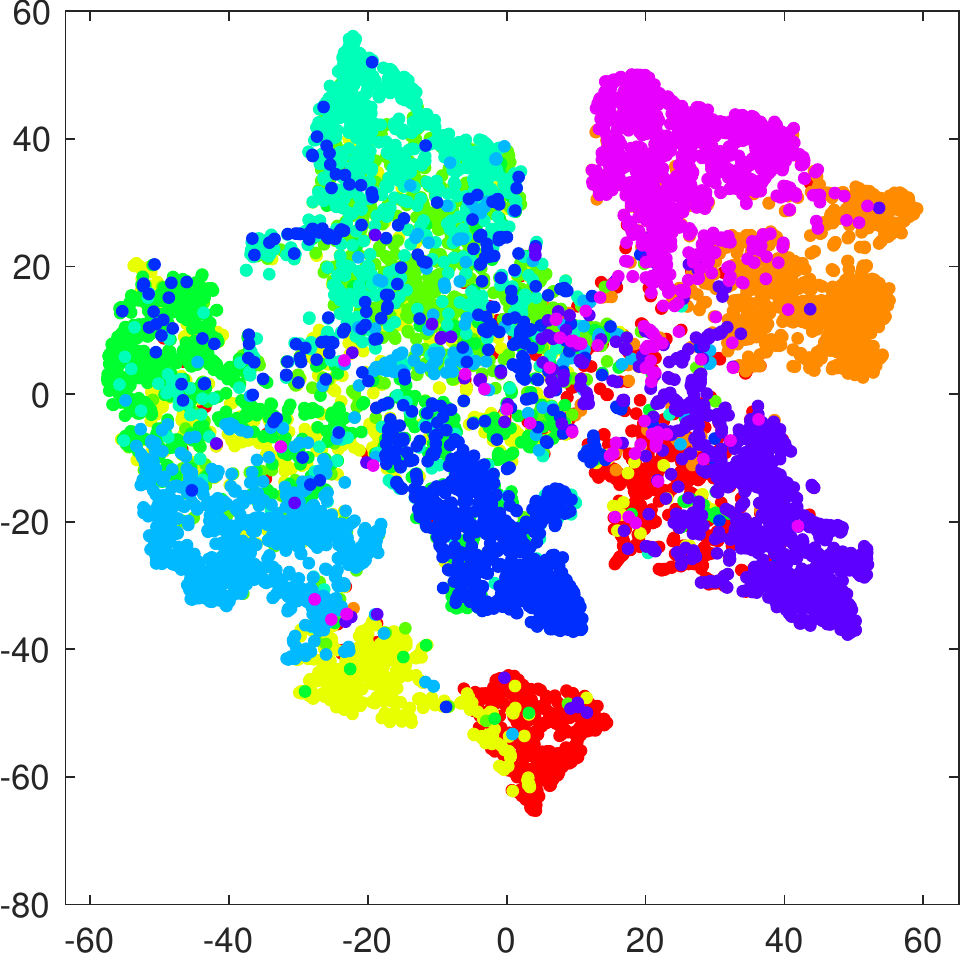}
		}
		\caption{t-SNE visualization on CIFAR-10 dataset. (64 bits)}
		\label{tsne}
		\vspace{-2.0em}
	\end{figure}

	To better understand the manifold structure of learned hashing code,
	We compare the t-SNE visualization~\cite{maaten2008visualizing} of BGAN, MLS3RDUH, and DSAH in Figure.~\ref{tsne}, in which the data points within the same colors belong to the same class.  It could be visually found that DSAH shows a clearer structure, in which we can find some meaningful clusters. Those scatter points within the same class~ (color) thus own smaller hamming distances with each other.
	
	\subsubsection{Retrieval Result}
\begin{figure}[h]
		\setlength{\abovecaptionskip}{0.cm}
		\setlength{\belowcaptionskip}{-0.cm}
		\centering
		\includegraphics[width=0.9\linewidth]{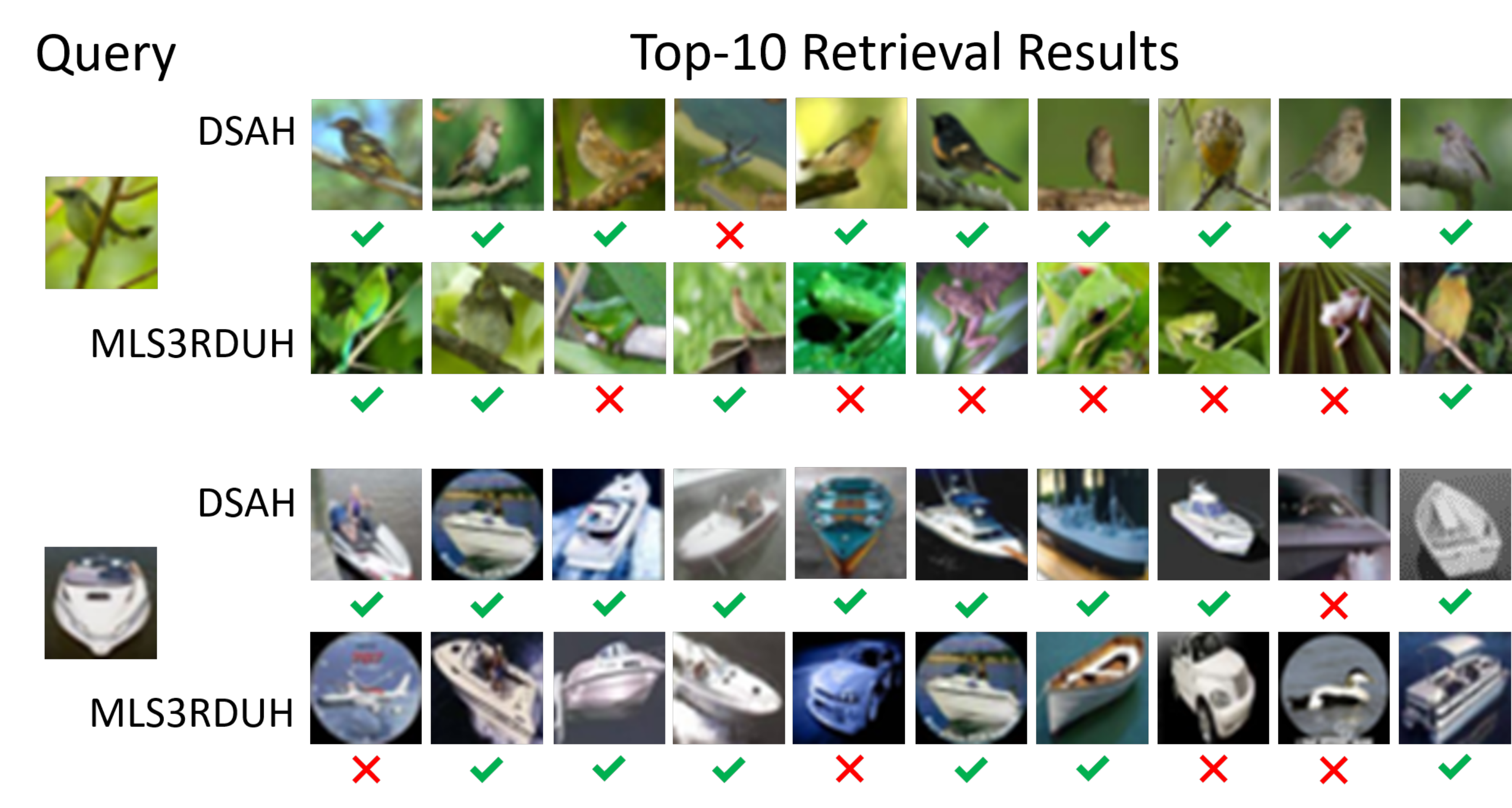}
		\caption{Top-10 retrieved results of DSAH and MLS$^3$RDUH on CIFAR-10 dataset. The green $\checkmark$ means that the retrieved samples belong to the same classes of the query images while the red $\times$ represents the returned in different classes.}
		\label{vis}
	\vspace{-2.0em}
	\end{figure}

	In Figure~\ref{vis}, we display some retrieval examples that return the top-10 samples based on the Hamming ranking distance on the CIFAR-10 dataset (64-bit). Comparing to the best baseline, our DSAH has fewer fault images. Specially, we found that MLS$^3$RDUH confusing birds in a green scene and flags, while the high-quality prediction of DSAH demonstrates that DSAH could well-distinguish the semantic information of images rather than the low-level feature.
	
	\section{Conclusions}
	In this paper, we propose a novel deep unsupervised hashing method DSAH to yield binary codes with fully exploring semantic information behind the data pairs. Particularly, DSAH provides two innovative designs: AND and PIC. To overcome the limitation of fixed semantic similarity, we adopt AND technique to refine the pre-computed similarity matrix with the learned representation and adaptively capture the implicit semantic structure behind the data distribution. Further, we employ PIC to distinguish the different importance of data pairs and assign an adjustable weight to each data pair, which fully explores the discriminative information of training data pairs. By combining PIC and AND in one framework, DSAH learns better hash codes in a self-adaptive manner. The extensive experiments on three benchmarks datasets demonstrate the effectiveness of our techniques, and DSAH can achieve competitive performance.
	
	\section{Acknowledgements}
	This work is jointly supported by the  2021 Tencent Rhino-Bird Research Elite Training Program, and
	the Major Project of the New Generation of Artificial Intelligence (No. 2018AAA0102900), NSFC under Grant No. 61773268, Natural Science Foundation of SZU (Grant No. 000346).

	\bibliographystyle{ACM-Reference-Format}
	\bibliography{sample-base}


\begin{thebibliography}{53}


\ifx \showCODEN    \undefined \def \showCODEN     #1{\unskip}     \fi
\ifx \showDOI      \undefined \def \showDOI       #1{#1}\fi
\ifx \showISBNx    \undefined \def \showISBNx     #1{\unskip}     \fi
\ifx \showISBNxiii \undefined \def \showISBNxiii  #1{\unskip}     \fi
\ifx \showISSN     \undefined \def \showISSN      #1{\unskip}     \fi
\ifx \showLCCN     \undefined \def \showLCCN      #1{\unskip}     \fi
\ifx \shownote     \undefined \def \shownote      #1{#1}          \fi
\ifx \showarticletitle \undefined \def \showarticletitle #1{#1}   \fi
\ifx \showURL      \undefined \def \showURL       {\relax}        \fi
\providecommand\bibfield[2]{#2}
\providecommand\bibinfo[2]{#2}
\providecommand\natexlab[1]{#1}
\providecommand\showeprint[2][]{arXiv:#2}

\bibitem[\protect\citeauthoryear{Andoni and Indyk}{Andoni and Indyk}{2006}]%
        {andoni2006near}
\bibfield{author}{\bibinfo{person}{Alexandr Andoni} {and}
  \bibinfo{person}{Piotr Indyk}.} \bibinfo{year}{2006}\natexlab{}.
\newblock \showarticletitle{Near-optimal hashing algorithms for approximate
  nearest neighbor in high dimensions}. In \bibinfo{booktitle}{\emph{2006 47th
  annual IEEE symposium on foundations of computer science (FOCS'06)}}. IEEE,
  \bibinfo{pages}{459--468}.
\newblock


\bibitem[\protect\citeauthoryear{Cao, Long, Wang, Zhu, and Wen}{Cao
  et~al\mbox{.}}{2016}]%
        {cao2016deep}
\bibfield{author}{\bibinfo{person}{Yue Cao}, \bibinfo{person}{Mingsheng Long},
  \bibinfo{person}{Jianmin Wang}, \bibinfo{person}{Han Zhu}, {and}
  \bibinfo{person}{Qingfu Wen}.} \bibinfo{year}{2016}\natexlab{}.
\newblock \showarticletitle{Deep quantization network for efficient image
  retrieval}. In \bibinfo{booktitle}{\emph{Proceedings of the AAAI Conference
  on Artificial Intelligence}}, Vol.~\bibinfo{volume}{30}.
\newblock


\bibitem[\protect\citeauthoryear{Cao, Long, Wang, and Yu}{Cao
  et~al\mbox{.}}{2017}]%
        {cao2017hashnet}
\bibfield{author}{\bibinfo{person}{Zhangjie Cao}, \bibinfo{person}{Mingsheng
  Long}, \bibinfo{person}{Jianmin Wang}, {and} \bibinfo{person}{Philip~S Yu}.}
  \bibinfo{year}{2017}\natexlab{}.
\newblock \showarticletitle{Hashnet: Deep learning to hash by continuation}. In
  \bibinfo{booktitle}{\emph{Proceedings of the IEEE international conference on
  computer vision}}. \bibinfo{pages}{5608--5617}.
\newblock


\bibitem[\protect\citeauthoryear{Chen, Lai, Ding, Lin, and Wong}{Chen
  et~al\mbox{.}}{2019}]%
        {chen2019deep}
\bibfield{author}{\bibinfo{person}{Yudong Chen}, \bibinfo{person}{Zhihui Lai},
  \bibinfo{person}{Yujuan Ding}, \bibinfo{person}{Kaiyi Lin}, {and}
  \bibinfo{person}{Wai~Keung Wong}.} \bibinfo{year}{2019}\natexlab{}.
\newblock \showarticletitle{Deep supervised hashing with anchor graph}. In
  \bibinfo{booktitle}{\emph{Proceedings of the IEEE International Conference on
  Computer Vision}}. \bibinfo{pages}{9796--9804}.
\newblock


\bibitem[\protect\citeauthoryear{Chua, Tang, Hong, Li, Luo, and Zheng}{Chua
  et~al\mbox{.}}{2009}]%
        {chua2009nus}
\bibfield{author}{\bibinfo{person}{Tat-Seng Chua}, \bibinfo{person}{Jinhui
  Tang}, \bibinfo{person}{Richang Hong}, \bibinfo{person}{Haojie Li},
  \bibinfo{person}{Zhiping Luo}, {and} \bibinfo{person}{Yantao Zheng}.}
  \bibinfo{year}{2009}\natexlab{}.
\newblock \showarticletitle{NUS-WIDE: a real-world web image database from
  National University of Singapore}. In \bibinfo{booktitle}{\emph{Proceedings
  of the ACM international conference on image and video retrieval}}.
  \bibinfo{pages}{1--9}.
\newblock


\bibitem[\protect\citeauthoryear{Dai, Guo, Kumar, He, and Song}{Dai
  et~al\mbox{.}}{2017}]%
        {dai2017stochastic}
\bibfield{author}{\bibinfo{person}{Bo Dai}, \bibinfo{person}{Ruiqi Guo},
  \bibinfo{person}{Sanjiv Kumar}, \bibinfo{person}{Niao He}, {and}
  \bibinfo{person}{Le Song}.} \bibinfo{year}{2017}\natexlab{}.
\newblock \showarticletitle{Stochastic generative hashing}. In
  \bibinfo{booktitle}{\emph{Proceedings of the 34th International Conference on
  Machine Learning-Volume 70}}. \bibinfo{pages}{913--922}.
\newblock


\bibitem[\protect\citeauthoryear{Deng, Yang, Liu, Li, Liu, and Tao}{Deng
  et~al\mbox{.}}{2019}]%
        {deng2019unsupervised}
\bibfield{author}{\bibinfo{person}{Cheng Deng}, \bibinfo{person}{Erkun Yang},
  \bibinfo{person}{Tongliang Liu}, \bibinfo{person}{Jie Li},
  \bibinfo{person}{Wei Liu}, {and} \bibinfo{person}{Dacheng Tao}.}
  \bibinfo{year}{2019}\natexlab{}.
\newblock \showarticletitle{Unsupervised semantic-preserving adversarial
  hashing for image search}.
\newblock \bibinfo{journal}{\emph{IEEE Transactions on Image Processing}}
  \bibinfo{volume}{28}, \bibinfo{number}{8} (\bibinfo{year}{2019}),
  \bibinfo{pages}{4032--4044}.
\newblock


\bibitem[\protect\citeauthoryear{Erin~Liong, Lu, Wang, Moulin, and
  Zhou}{Erin~Liong et~al\mbox{.}}{2015}]%
        {erin2015deep}
\bibfield{author}{\bibinfo{person}{Venice Erin~Liong}, \bibinfo{person}{Jiwen
  Lu}, \bibinfo{person}{Gang Wang}, \bibinfo{person}{Pierre Moulin}, {and}
  \bibinfo{person}{Jie Zhou}.} \bibinfo{year}{2015}\natexlab{}.
\newblock \showarticletitle{Deep hashing for compact binary codes learning}. In
  \bibinfo{booktitle}{\emph{Proceedings of the IEEE conference on computer
  vision and pattern recognition}}. \bibinfo{pages}{2475--2483}.
\newblock


\bibitem[\protect\citeauthoryear{Gionis, Indyk, Motwani, et~al\mbox{.}}{Gionis
  et~al\mbox{.}}{1999}]%
        {gionis1999similarity}
\bibfield{author}{\bibinfo{person}{Aristides Gionis}, \bibinfo{person}{Piotr
  Indyk}, \bibinfo{person}{Rajeev Motwani}, {et~al\mbox{.}}}
  \bibinfo{year}{1999}\natexlab{}.
\newblock \showarticletitle{Similarity search in high dimensions via hashing}.
  In \bibinfo{booktitle}{\emph{Vldb}}, Vol.~\bibinfo{volume}{99}.
  \bibinfo{pages}{518--529}.
\newblock


\bibitem[\protect\citeauthoryear{Gong, Lazebnik, Gordo, and Perronnin}{Gong
  et~al\mbox{.}}{2012}]%
        {gong2012iterative}
\bibfield{author}{\bibinfo{person}{Yunchao Gong}, \bibinfo{person}{Svetlana
  Lazebnik}, \bibinfo{person}{Albert Gordo}, {and} \bibinfo{person}{Florent
  Perronnin}.} \bibinfo{year}{2012}\natexlab{}.
\newblock \showarticletitle{Iterative quantization: A procrustean approach to
  learning binary codes for large-scale image retrieval}.
\newblock \bibinfo{journal}{\emph{IEEE transactions on pattern analysis and
  machine intelligence}} \bibinfo{volume}{35}, \bibinfo{number}{12}
  (\bibinfo{year}{2012}), \bibinfo{pages}{2916--2929}.
\newblock


\bibitem[\protect\citeauthoryear{Gui, Liu, Sun, Tao, and Tan}{Gui
  et~al\mbox{.}}{2017}]%
        {gui2017fast}
\bibfield{author}{\bibinfo{person}{Jie Gui}, \bibinfo{person}{Tongliang Liu},
  \bibinfo{person}{Zhenan Sun}, \bibinfo{person}{Dacheng Tao}, {and}
  \bibinfo{person}{Tieniu Tan}.} \bibinfo{year}{2017}\natexlab{}.
\newblock \showarticletitle{Fast supervised discrete hashing}.
\newblock \bibinfo{journal}{\emph{IEEE transactions on pattern analysis and
  machine intelligence}} \bibinfo{volume}{40}, \bibinfo{number}{2}
  (\bibinfo{year}{2017}), \bibinfo{pages}{490--496}.
\newblock


\bibitem[\protect\citeauthoryear{He, Wen, and Sun}{He et~al\mbox{.}}{2013}]%
        {he2013k}
\bibfield{author}{\bibinfo{person}{Kaiming He}, \bibinfo{person}{Fang Wen},
  {and} \bibinfo{person}{Jian Sun}.} \bibinfo{year}{2013}\natexlab{}.
\newblock \showarticletitle{K-means hashing: An affinity-preserving
  quantization method for learning binary compact codes}. In
  \bibinfo{booktitle}{\emph{Proceedings of the IEEE conference on computer
  vision and pattern recognition}}. \bibinfo{pages}{2938--2945}.
\newblock


\bibitem[\protect\citeauthoryear{He, Wang, and Cheng}{He et~al\mbox{.}}{2019}]%
        {he2019k}
\bibfield{author}{\bibinfo{person}{Xiangyu He}, \bibinfo{person}{Peisong Wang},
  {and} \bibinfo{person}{Jian Cheng}.} \bibinfo{year}{2019}\natexlab{}.
\newblock \showarticletitle{K-nearest neighbors hashing}. In
  \bibinfo{booktitle}{\emph{Proceedings of the IEEE Conference on Computer
  Vision and Pattern Recognition}}. \bibinfo{pages}{2839--2848}.
\newblock


\bibitem[\protect\citeauthoryear{Huiskes and Lew}{Huiskes and Lew}{2008}]%
        {huiskes2008mir}
\bibfield{author}{\bibinfo{person}{Mark~J Huiskes} {and}
  \bibinfo{person}{Michael~S Lew}.} \bibinfo{year}{2008}\natexlab{}.
\newblock \showarticletitle{The MIR flickr retrieval evaluation}. In
  \bibinfo{booktitle}{\emph{Proceedings of the 1st ACM international conference
  on Multimedia information retrieval}}. \bibinfo{pages}{39--43}.
\newblock


\bibitem[\protect\citeauthoryear{Indyk and Motwani}{Indyk and Motwani}{1998}]%
        {indyk1998approximate}
\bibfield{author}{\bibinfo{person}{Piotr Indyk} {and} \bibinfo{person}{Rajeev
  Motwani}.} \bibinfo{year}{1998}\natexlab{}.
\newblock \showarticletitle{Approximate nearest neighbors: towards removing the
  curse of dimensionality}. In \bibinfo{booktitle}{\emph{Proceedings of the
  thirtieth annual ACM symposium on Theory of computing}}.
  \bibinfo{pages}{604--613}.
\newblock


\bibitem[\protect\citeauthoryear{Jiang and Li}{Jiang and Li}{2017}]%
        {jiang2017asymmetric}
\bibfield{author}{\bibinfo{person}{Qing-Yuan Jiang} {and}
  \bibinfo{person}{Wu-Jun Li}.} \bibinfo{year}{2017}\natexlab{}.
\newblock \showarticletitle{Asymmetric deep supervised hashing}.
\newblock \bibinfo{journal}{\emph{arXiv preprint arXiv:1707.08325}}
  (\bibinfo{year}{2017}).
\newblock


\bibitem[\protect\citeauthoryear{Kang, Li, and Zhou}{Kang
  et~al\mbox{.}}{2016}]%
        {kang2016column}
\bibfield{author}{\bibinfo{person}{Wang-Cheng Kang}, \bibinfo{person}{Wu-Jun
  Li}, {and} \bibinfo{person}{Zhi-Hua Zhou}.} \bibinfo{year}{2016}\natexlab{}.
\newblock \showarticletitle{Column sampling based discrete supervised
  hashing.}. In \bibinfo{booktitle}{\emph{AAAI}}. \bibinfo{pages}{1230--1236}.
\newblock


\bibitem[\protect\citeauthoryear{Kong and Li}{Kong and Li}{2012}]%
        {kong2012isotropic}
\bibfield{author}{\bibinfo{person}{Weihao Kong} {and} \bibinfo{person}{Wu-Jun
  Li}.} \bibinfo{year}{2012}\natexlab{}.
\newblock \showarticletitle{Isotropic hashing}. In
  \bibinfo{booktitle}{\emph{Advances in neural information processing
  systems}}. \bibinfo{pages}{1646--1654}.
\newblock


\bibitem[\protect\citeauthoryear{Krizhevsky, Hinton, et~al\mbox{.}}{Krizhevsky
  et~al\mbox{.}}{2009}]%
        {krizhevsky2009learning}
\bibfield{author}{\bibinfo{person}{Alex Krizhevsky}, \bibinfo{person}{Geoffrey
  Hinton}, {et~al\mbox{.}}} \bibinfo{year}{2009}\natexlab{}.
\newblock \showarticletitle{Learning multiple layers of features from tiny
  images}.
\newblock  (\bibinfo{year}{2009}).
\newblock


\bibitem[\protect\citeauthoryear{Kulis and Grauman}{Kulis and Grauman}{2009}]%
        {kulis2009kernelized}
\bibfield{author}{\bibinfo{person}{Brian Kulis} {and} \bibinfo{person}{Kristen
  Grauman}.} \bibinfo{year}{2009}\natexlab{}.
\newblock \showarticletitle{Kernelized locality-sensitive hashing for scalable
  image search}. In \bibinfo{booktitle}{\emph{2009 IEEE 12th international
  conference on computer vision}}. IEEE, \bibinfo{pages}{2130--2137}.
\newblock


\bibitem[\protect\citeauthoryear{Li, Sun, He, and Tan}{Li
  et~al\mbox{.}}{2017b}]%
        {li2017deep}
\bibfield{author}{\bibinfo{person}{Qi Li}, \bibinfo{person}{Zhenan Sun},
  \bibinfo{person}{Ran He}, {and} \bibinfo{person}{Tieniu Tan}.}
  \bibinfo{year}{2017}\natexlab{b}.
\newblock \showarticletitle{Deep supervised discrete hashing}. In
  \bibinfo{booktitle}{\emph{Advances in neural information processing
  systems}}. \bibinfo{pages}{2482--2491}.
\newblock


\bibitem[\protect\citeauthoryear{Li, Gao, and Xu}{Li et~al\mbox{.}}{2017a}]%
        {li2017deeps}
\bibfield{author}{\bibinfo{person}{Tong Li}, \bibinfo{person}{Sheng Gao}, {and}
  \bibinfo{person}{Yajing Xu}.} \bibinfo{year}{2017}\natexlab{a}.
\newblock \showarticletitle{Deep multi-similarity hashing for multi-label image
  retrieval}. In \bibinfo{booktitle}{\emph{Proceedings of the 2017 ACM on
  Conference on Information and Knowledge Management}}.
  \bibinfo{pages}{2159--2162}.
\newblock


\bibitem[\protect\citeauthoryear{Li, Wang, and Kang}{Li et~al\mbox{.}}{2015}]%
        {li2015feature}
\bibfield{author}{\bibinfo{person}{Wu-Jun Li}, \bibinfo{person}{Sheng Wang},
  {and} \bibinfo{person}{Wang-Cheng Kang}.} \bibinfo{year}{2015}\natexlab{}.
\newblock \showarticletitle{Feature learning based deep supervised hashing with
  pairwise labels}.
\newblock \bibinfo{journal}{\emph{arXiv preprint arXiv:1511.03855}}
  (\bibinfo{year}{2015}).
\newblock


\bibitem[\protect\citeauthoryear{Lin, Lu, Chen, and Zhou}{Lin
  et~al\mbox{.}}{2016}]%
        {lin2016learning}
\bibfield{author}{\bibinfo{person}{Kevin Lin}, \bibinfo{person}{Jiwen Lu},
  \bibinfo{person}{Chu-Song Chen}, {and} \bibinfo{person}{Jie Zhou}.}
  \bibinfo{year}{2016}\natexlab{}.
\newblock \showarticletitle{Learning compact binary descriptors with
  unsupervised deep neural networks}. In \bibinfo{booktitle}{\emph{Proceedings
  of the IEEE Conference on Computer Vision and Pattern Recognition}}.
  \bibinfo{pages}{1183--1192}.
\newblock


\bibitem[\protect\citeauthoryear{Liu, Wang, Kumar, and Chang}{Liu
  et~al\mbox{.}}{2011}]%
        {liu2011hashing}
\bibfield{author}{\bibinfo{person}{Wei Liu}, \bibinfo{person}{Jun Wang},
  \bibinfo{person}{Sanjiv Kumar}, {and} \bibinfo{person}{Shih-Fu Chang}.}
  \bibinfo{year}{2011}\natexlab{}.
\newblock \showarticletitle{Hashing with graphs}. In
  \bibinfo{booktitle}{\emph{ICML}}.
\newblock


\bibitem[\protect\citeauthoryear{Liu, Li, Wang, Yu, Domenicon, and Zhang}{Liu
  et~al\mbox{.}}{2019}]%
        {liu2019cross}
\bibfield{author}{\bibinfo{person}{Xuanwu Liu}, \bibinfo{person}{Zhao Li},
  \bibinfo{person}{Jun Wang}, \bibinfo{person}{Guoxian Yu},
  \bibinfo{person}{Carlotta Domenicon}, {and} \bibinfo{person}{Xiangliang
  Zhang}.} \bibinfo{year}{2019}\natexlab{}.
\newblock \showarticletitle{Cross-modal zero-shot hashing}. In
  \bibinfo{booktitle}{\emph{2019 IEEE International Conference on Data Mining
  (ICDM)}}. IEEE, \bibinfo{pages}{449--458}.
\newblock


\bibitem[\protect\citeauthoryear{Lowe}{Lowe}{1999}]%
        {lowe1999object}
\bibfield{author}{\bibinfo{person}{David~G Lowe}.}
  \bibinfo{year}{1999}\natexlab{}.
\newblock \showarticletitle{Object recognition from local scale-invariant
  features}. In \bibinfo{booktitle}{\emph{Proceedings of the seventh IEEE
  international conference on computer vision}}, Vol.~\bibinfo{volume}{2}.
  Ieee, \bibinfo{pages}{1150--1157}.
\newblock


\bibitem[\protect\citeauthoryear{Luo, Nie, He, Wu, Chen, and Xu}{Luo
  et~al\mbox{.}}{2018}]%
        {luo2018fast}
\bibfield{author}{\bibinfo{person}{Xin Luo}, \bibinfo{person}{Liqiang Nie},
  \bibinfo{person}{Xiangnan He}, \bibinfo{person}{Ye Wu},
  \bibinfo{person}{Zhen-Duo Chen}, {and} \bibinfo{person}{Xin-Shun Xu}.}
  \bibinfo{year}{2018}\natexlab{}.
\newblock \showarticletitle{Fast scalable supervised hashing}. In
  \bibinfo{booktitle}{\emph{The 41st International ACM SIGIR Conference on
  Research \& Development in Information Retrieval}}.
  \bibinfo{pages}{735--744}.
\newblock


\bibitem[\protect\citeauthoryear{Maaten and Hinton}{Maaten and Hinton}{2008}]%
        {maaten2008visualizing}
\bibfield{author}{\bibinfo{person}{Laurens van~der Maaten} {and}
  \bibinfo{person}{Geoffrey Hinton}.} \bibinfo{year}{2008}\natexlab{}.
\newblock \showarticletitle{Visualizing data using t-SNE}.
\newblock \bibinfo{journal}{\emph{Journal of machine learning research}}
  \bibinfo{volume}{9}, \bibinfo{number}{Nov} (\bibinfo{year}{2008}),
  \bibinfo{pages}{2579--2605}.
\newblock


\bibitem[\protect\citeauthoryear{Qin, Huang, Wei, Xie, and Zhang}{Qin
  et~al\mbox{.}}{2020}]%
        {qin2020unsupervised}
\bibfield{author}{\bibinfo{person}{Qibing Qin}, \bibinfo{person}{Lei Huang},
  \bibinfo{person}{Zhiqiang Wei}, \bibinfo{person}{Kezhen Xie}, {and}
  \bibinfo{person}{Wenfeng Zhang}.} \bibinfo{year}{2020}\natexlab{}.
\newblock \showarticletitle{Unsupervised Deep Multi-Similarity Hashing with
  Semantic Structure for Image Retrieval}.
\newblock \bibinfo{journal}{\emph{IEEE Transactions on Circuits and Systems for
  Video Technology}} (\bibinfo{year}{2020}).
\newblock


\bibitem[\protect\citeauthoryear{Qiu, Pan, Yao, and Mei}{Qiu
  et~al\mbox{.}}{2017}]%
        {qiu2017deep1}
\bibfield{author}{\bibinfo{person}{Zhaofan Qiu}, \bibinfo{person}{Yingwei Pan},
  \bibinfo{person}{Ting Yao}, {and} \bibinfo{person}{Tao Mei}.}
  \bibinfo{year}{2017}\natexlab{}.
\newblock \showarticletitle{Deep semantic hashing with generative adversarial
  networks}. In \bibinfo{booktitle}{\emph{Proceedings of the 40th International
  ACM SIGIR Conference on Research and Development in Information Retrieval}}.
  \bibinfo{pages}{225--234}.
\newblock


\bibitem[\protect\citeauthoryear{Salakhutdinov and Hinton}{Salakhutdinov and
  Hinton}{2009}]%
        {salakhutdinov2009semantic}
\bibfield{author}{\bibinfo{person}{Ruslan Salakhutdinov} {and}
  \bibinfo{person}{Geoffrey Hinton}.} \bibinfo{year}{2009}\natexlab{}.
\newblock \showarticletitle{Semantic hashing}.
\newblock \bibinfo{journal}{\emph{International Journal of Approximate
  Reasoning}} \bibinfo{volume}{50}, \bibinfo{number}{7} (\bibinfo{year}{2009}),
  \bibinfo{pages}{969--978}.
\newblock


\bibitem[\protect\citeauthoryear{Shannon}{Shannon}{1948}]%
        {shannon1948mathematical}
\bibfield{author}{\bibinfo{person}{Claude~E Shannon}.}
  \bibinfo{year}{1948}\natexlab{}.
\newblock \showarticletitle{A mathematical theory of communication}.
\newblock \bibinfo{journal}{\emph{The Bell system technical journal}}
  \bibinfo{volume}{27}, \bibinfo{number}{3} (\bibinfo{year}{1948}),
  \bibinfo{pages}{379--423}.
\newblock


\bibitem[\protect\citeauthoryear{Shen, Shen, Liu, and Tao~Shen}{Shen
  et~al\mbox{.}}{2015}]%
        {shen2015supervised}
\bibfield{author}{\bibinfo{person}{Fumin Shen}, \bibinfo{person}{Chunhua Shen},
  \bibinfo{person}{Wei Liu}, {and} \bibinfo{person}{Heng Tao~Shen}.}
  \bibinfo{year}{2015}\natexlab{}.
\newblock \showarticletitle{Supervised discrete hashing}. In
  \bibinfo{booktitle}{\emph{Proceedings of the IEEE conference on computer
  vision and pattern recognition}}. \bibinfo{pages}{37--45}.
\newblock


\bibitem[\protect\citeauthoryear{Shen, Xu, Liu, Yang, Huang, and Shen}{Shen
  et~al\mbox{.}}{2018}]%
        {shen2018unsupervised}
\bibfield{author}{\bibinfo{person}{Fumin Shen}, \bibinfo{person}{Yan Xu},
  \bibinfo{person}{Li Liu}, \bibinfo{person}{Yang Yang}, \bibinfo{person}{Zi
  Huang}, {and} \bibinfo{person}{Heng~Tao Shen}.}
  \bibinfo{year}{2018}\natexlab{}.
\newblock \showarticletitle{Unsupervised deep hashing with similarity-adaptive
  and discrete optimization}.
\newblock \bibinfo{journal}{\emph{IEEE transactions on pattern analysis and
  machine intelligence}} \bibinfo{volume}{40}, \bibinfo{number}{12}
  (\bibinfo{year}{2018}), \bibinfo{pages}{3034--3044}.
\newblock


\bibitem[\protect\citeauthoryear{Shen, Qin, Chen, Yu, Liu, Zhu, Shen, and
  Shao}{Shen et~al\mbox{.}}{2020}]%
        {shen2020auto}
\bibfield{author}{\bibinfo{person}{Yuming Shen}, \bibinfo{person}{Jie Qin},
  \bibinfo{person}{Jiaxin Chen}, \bibinfo{person}{Mengyang Yu},
  \bibinfo{person}{Li Liu}, \bibinfo{person}{Fan Zhu}, \bibinfo{person}{Fumin
  Shen}, {and} \bibinfo{person}{Ling Shao}.} \bibinfo{year}{2020}\natexlab{}.
\newblock \showarticletitle{Auto-Encoding Twin-Bottleneck Hashing}. In
  \bibinfo{booktitle}{\emph{Proceedings of the IEEE/CVF Conference on Computer
  Vision and Pattern Recognition}}. \bibinfo{pages}{2818--2827}.
\newblock


\bibitem[\protect\citeauthoryear{Simonyan and Zisserman}{Simonyan and
  Zisserman}{2014}]%
        {simonyan2014very}
\bibfield{author}{\bibinfo{person}{Karen Simonyan} {and}
  \bibinfo{person}{Andrew Zisserman}.} \bibinfo{year}{2014}\natexlab{}.
\newblock \showarticletitle{Very deep convolutional networks for large-scale
  image recognition}.
\newblock \bibinfo{journal}{\emph{arXiv preprint arXiv:1409.1556}}
  (\bibinfo{year}{2014}).
\newblock


\bibitem[\protect\citeauthoryear{Song, He, Gao, Xu, Hanjalic, and Shen}{Song
  et~al\mbox{.}}{2018}]%
        {song2018binary}
\bibfield{author}{\bibinfo{person}{Jingkuan Song}, \bibinfo{person}{Tao He},
  \bibinfo{person}{Lianli Gao}, \bibinfo{person}{Xing Xu},
  \bibinfo{person}{Alan Hanjalic}, {and} \bibinfo{person}{Heng~Tao Shen}.}
  \bibinfo{year}{2018}\natexlab{}.
\newblock \showarticletitle{Binary generative adversarial networks for image
  retrieval}. In \bibinfo{booktitle}{\emph{Proceedings of the AAAI Conference
  on Artificial Intelligence}}, Vol.~\bibinfo{volume}{32}.
\newblock


\bibitem[\protect\citeauthoryear{Su, Zhang, Han, and Tian}{Su
  et~al\mbox{.}}{2018}]%
        {su2018greedy}
\bibfield{author}{\bibinfo{person}{Shupeng Su}, \bibinfo{person}{Chao Zhang},
  \bibinfo{person}{Kai Han}, {and} \bibinfo{person}{Yonghong Tian}.}
  \bibinfo{year}{2018}\natexlab{}.
\newblock \showarticletitle{Greedy hash: Towards fast optimization for accurate
  hash coding in cnn}. In \bibinfo{booktitle}{\emph{Advances in neural
  information processing systems}}. \bibinfo{pages}{798--807}.
\newblock


\bibitem[\protect\citeauthoryear{Tu, Mao, Feng, and Yu}{Tu
  et~al\mbox{.}}{2019}]%
        {tu2018object}
\bibfield{author}{\bibinfo{person}{Rong-Cheng Tu}, \bibinfo{person}{Xian-Ling
  Mao}, \bibinfo{person}{Bo-Si Feng}, {and} \bibinfo{person}{Shu-Ying Yu}.}
  \bibinfo{year}{2019}\natexlab{}.
\newblock \showarticletitle{Object detection based deep unsupervised hashing}.
\newblock  (\bibinfo{year}{2019}).
\newblock


\bibitem[\protect\citeauthoryear{Tu, Mao, Guo, Wei, and Huang}{Tu
  et~al\mbox{.}}{2021}]%
        {tu2021partial}
\bibfield{author}{\bibinfo{person}{Rong-Cheng Tu}, \bibinfo{person}{Xian-Ling
  Mao}, \bibinfo{person}{Jia-Nan Guo}, \bibinfo{person}{Wei Wei}, {and}
  \bibinfo{person}{Heyan Huang}.} \bibinfo{year}{2021}\natexlab{}.
\newblock \showarticletitle{Partial-Softmax Loss based Deep Hashing}. In
  \bibinfo{booktitle}{\emph{Proceedings of the Web Conference 2021}}.
  \bibinfo{pages}{2869--2878}.
\newblock


\bibitem[\protect\citeauthoryear{Tu, Mao, Ma, Hu, Yan, Wei, and Huang}{Tu
  et~al\mbox{.}}{2020b}]%
        {tu2020deep}
\bibfield{author}{\bibinfo{person}{Rong-Cheng Tu}, \bibinfo{person}{Xian-Ling
  Mao}, \bibinfo{person}{Bing Ma}, \bibinfo{person}{Yong Hu},
  \bibinfo{person}{Tan Yan}, \bibinfo{person}{Wei Wei}, {and}
  \bibinfo{person}{Heyan Huang}.} \bibinfo{year}{2020}\natexlab{b}.
\newblock \showarticletitle{Deep cross-modal hashing with hashing functions and
  unified hash codes jointly learning}.
\newblock \bibinfo{journal}{\emph{IEEE Transactions on Knowledge and Data
  Engineering}} (\bibinfo{year}{2020}).
\newblock


\bibitem[\protect\citeauthoryear{Tu, Mao, and Wei}{Tu et~al\mbox{.}}{2020a}]%
        {tu2020mls3rduh}
\bibfield{author}{\bibinfo{person}{Rong-Cheng Tu}, \bibinfo{person}{Xian-Ling
  Mao}, {and} \bibinfo{person}{Wei Wei}.} \bibinfo{year}{2020}\natexlab{a}.
\newblock \showarticletitle{MLS3RDUH: Deep Unsupervised Hashing via Manifold
  based Local Semantic Similarity Structure Reconstructing}. In
  \bibinfo{booktitle}{\emph{Proceedings of the Twenty-Ninth International Joint
  Conference on Artificial Intelligence}}. \bibinfo{pages}{3466--3472}.
\newblock


\bibitem[\protect\citeauthoryear{Wang, Liu, and Zhao}{Wang
  et~al\mbox{.}}{2018}]%
        {wang2018deep}
\bibfield{author}{\bibinfo{person}{Bingning Wang}, \bibinfo{person}{Kang Liu},
  {and} \bibinfo{person}{Jun Zhao}.} \bibinfo{year}{2018}\natexlab{}.
\newblock \showarticletitle{Deep Semantic Hashing with Multi-Adversarial
  Training}. In \bibinfo{booktitle}{\emph{Proceedings of the 27th ACM
  International Conference on Information and Knowledge Management}}.
  \bibinfo{pages}{1453--1462}.
\newblock


\bibitem[\protect\citeauthoryear{Wang, Zhang, Sebe, Shen, et~al\mbox{.}}{Wang
  et~al\mbox{.}}{2017}]%
        {wang2017survey}
\bibfield{author}{\bibinfo{person}{Jingdong Wang}, \bibinfo{person}{Ting
  Zhang}, \bibinfo{person}{Nicu Sebe}, \bibinfo{person}{Heng~Tao Shen},
  {et~al\mbox{.}}} \bibinfo{year}{2017}\natexlab{}.
\newblock \showarticletitle{A survey on learning to hash}.
\newblock \bibinfo{journal}{\emph{IEEE transactions on pattern analysis and
  machine intelligence}} \bibinfo{volume}{40}, \bibinfo{number}{4}
  (\bibinfo{year}{2017}), \bibinfo{pages}{769--790}.
\newblock


\bibitem[\protect\citeauthoryear{Weiss, Torralba, and Fergus}{Weiss
  et~al\mbox{.}}{2009}]%
        {weiss2009spectral}
\bibfield{author}{\bibinfo{person}{Yair Weiss}, \bibinfo{person}{Antonio
  Torralba}, {and} \bibinfo{person}{Rob Fergus}.}
  \bibinfo{year}{2009}\natexlab{}.
\newblock \showarticletitle{Spectral hashing}. In
  \bibinfo{booktitle}{\emph{Advances in neural information processing
  systems}}. \bibinfo{pages}{1753--1760}.
\newblock


\bibitem[\protect\citeauthoryear{Xia, Pan, Lai, Liu, and Yan}{Xia
  et~al\mbox{.}}{2014}]%
        {xia2014supervised}
\bibfield{author}{\bibinfo{person}{Rongkai Xia}, \bibinfo{person}{Yan Pan},
  \bibinfo{person}{Hanjiang Lai}, \bibinfo{person}{Cong Liu}, {and}
  \bibinfo{person}{Shuicheng Yan}.} \bibinfo{year}{2014}\natexlab{}.
\newblock \showarticletitle{Supervised hashing for image retrieval via image
  representation learning.}. In \bibinfo{booktitle}{\emph{AAAI}},
  Vol.~\bibinfo{volume}{1}. \bibinfo{pages}{2}.
\newblock


\bibitem[\protect\citeauthoryear{Xia, He, Kohli, and Sun}{Xia
  et~al\mbox{.}}{2015}]%
        {Xia_2015_CVPR}
\bibfield{author}{\bibinfo{person}{Yan Xia}, \bibinfo{person}{Kaiming He},
  \bibinfo{person}{Pushmeet Kohli}, {and} \bibinfo{person}{Jian Sun}.}
  \bibinfo{year}{2015}\natexlab{}.
\newblock \showarticletitle{Sparse Projections for High-Dimensional Binary
  Codes}. In \bibinfo{booktitle}{\emph{Proceedings of the IEEE Conference on
  Computer Vision and Pattern Recognition (CVPR)}}.
\newblock


\bibitem[\protect\citeauthoryear{Yang, Deng, Liu, Liu, and Tao}{Yang
  et~al\mbox{.}}{2018}]%
        {yang2018semantic}
\bibfield{author}{\bibinfo{person}{Erkun Yang}, \bibinfo{person}{Cheng Deng},
  \bibinfo{person}{Tongliang Liu}, \bibinfo{person}{Wei Liu}, {and}
  \bibinfo{person}{Dacheng Tao}.} \bibinfo{year}{2018}\natexlab{}.
\newblock \showarticletitle{Semantic structure-based unsupervised deep
  hashing}. In \bibinfo{booktitle}{\emph{Proceedings of the 27th International
  Joint Conference on Artificial Intelligence}}. \bibinfo{pages}{1064--1070}.
\newblock


\bibitem[\protect\citeauthoryear{Yang, Liu, Deng, Liu, and Tao}{Yang
  et~al\mbox{.}}{2019}]%
        {yang2019distillhash}
\bibfield{author}{\bibinfo{person}{Erkun Yang}, \bibinfo{person}{Tongliang
  Liu}, \bibinfo{person}{Cheng Deng}, \bibinfo{person}{Wei Liu}, {and}
  \bibinfo{person}{Dacheng Tao}.} \bibinfo{year}{2019}\natexlab{}.
\newblock \showarticletitle{Distillhash: Unsupervised deep hashing by
  distilling data pairs}. In \bibinfo{booktitle}{\emph{Proceedings of the IEEE
  Conference on Computer Vision and Pattern Recognition}}.
  \bibinfo{pages}{2946--2955}.
\newblock


\bibitem[\protect\citeauthoryear{Zhang, Gu, Yao, Zhang, Liu, Zhang, and
  Shao}{Zhang et~al\mbox{.}}{2020a}]%
        {zhangz2020deep}
\bibfield{author}{\bibinfo{person}{Haofeng Zhang}, \bibinfo{person}{Yifan Gu},
  \bibinfo{person}{Yazhou Yao}, \bibinfo{person}{Zheng Zhang},
  \bibinfo{person}{Li Liu}, \bibinfo{person}{Jian Zhang}, {and}
  \bibinfo{person}{Ling Shao}.} \bibinfo{year}{2020}\natexlab{a}.
\newblock \showarticletitle{Deep Unsupervised Self-evolutionary Hashing for
  Image Retrieval}.
\newblock \bibinfo{journal}{\emph{IEEE Transactions on Multimedia}}
  (\bibinfo{year}{2020}).
\newblock


\bibitem[\protect\citeauthoryear{Zhang, Wu, Zhou, Li, Wang, and Meng}{Zhang
  et~al\mbox{.}}{2020b}]%
        {zhang2020deep}
\bibfield{author}{\bibinfo{person}{Wanqian Zhang}, \bibinfo{person}{Dayan Wu},
  \bibinfo{person}{Yu Zhou}, \bibinfo{person}{Bo Li}, \bibinfo{person}{Weiping
  Wang}, {and} \bibinfo{person}{Dan Meng}.} \bibinfo{year}{2020}\natexlab{b}.
\newblock \showarticletitle{Deep Unsupervised Hybrid-similarity Hadamard
  Hashing}. In \bibinfo{booktitle}{\emph{Proceedings of the 28th ACM
  International Conference on Multimedia}}. \bibinfo{pages}{3274--3282}.
\newblock


\bibitem[\protect\citeauthoryear{Zhu, Long, Wang, and Cao}{Zhu
  et~al\mbox{.}}{2016}]%
        {zhu2016deep}
\bibfield{author}{\bibinfo{person}{Han Zhu}, \bibinfo{person}{Mingsheng Long},
  \bibinfo{person}{Jianmin Wang}, {and} \bibinfo{person}{Yue Cao}.}
  \bibinfo{year}{2016}\natexlab{}.
\newblock \showarticletitle{Deep hashing network for efficient similarity
  retrieval}. In \bibinfo{booktitle}{\emph{Thirtieth AAAI Conference on
  Artificial Intelligence}}.
\newblock


\end{thebibliography}
	
\end{document}